\documentclass[journal]{IEEEtran}

\usepackage{times}
\usepackage[pdftex]{graphicx}
\usepackage{subfigure}
\usepackage{amsmath,amssymb,amsopn,amstext,amsfonts}
\usepackage{cancel}
\usepackage[space]{cite}
\usepackage{pdfsync}
\usepackage{balance}
\usepackage{color}
\usepackage{mathtools}
\usepackage{bm}

\usepackage{diagbox}
\usepackage{float}
\usepackage{epstopdf}
\usepackage{pifont}
\usepackage{fixltx2e}
\usepackage{amsmath}
\usepackage{multirow}
\usepackage{url}
\usepackage{verbatim}
\usepackage[linkcolor=black,citecolor=black,urlcolor=black,colorlinks=true]{hyperref}
\usepackage{soul,xcolor}
\usepackage[linesnumbered,ruled,vlined]{algorithm2e}
\usepackage{subfiles}
\usepackage{array}

\newcommand{\tnb}[1]{\textnormal{\textbf{#1}}}
\newcommand{\tn}[1]{\textnormal{#1}}

\usepackage{ulem}
\let\oldnl\nl
\newcommand{\nonl}{\renewcommand{\nl}{\let\nl\oldnl}}


\graphicspath{{img/}{../img/}{../img2/}}
\DeclareGraphicsExtensions{,.jpg,.eps,.pdf,.jpeg,.png,.svg}

\SetCommentSty{mycommfont}
\newcolumntype{P}[1]{>{\centering\arraybackslash}p{#1}}

\begin{document}

\title{RACER: Rapid Collaborative Exploration with a Decentralized Multi-UAV System}
\author{Boyu Zhou,
        Hao Xu,
        and Shaojie Shen
\thanks{Boyu Zhou, Hao Xu and Shaojie Shen are with the Department of Electronic and Computer Engineering, Hong Kong University of Science and Technology, Hong Kong, China. {\tt\footnotesize $\{$bzhouai, hao.xu, eeshaojie$\}$@connect.ust.hk}}%
}

\maketitle

\begin{abstract}

Although the use of multiple Unmanned Aerial Vehicles (UAVs) has great potential for fast autonomous exploration, it has received far too little attention.
In this paper, we present RACER, a RApid Collaborative ExploRation approach using a fleet of decentralized UAVs.
To effectively dispatch the UAVs, a pairwise interaction based on an online hgrid space decomposition is used. 
It ensures that all UAVs simultaneously explore distinct regions, using only asynchronous and limited communication.
Further, we optimize the coverage paths of unknown space and balance the workloads partitioned to each UAV with a Capacitated Vehicle Routing Problem(CVRP) formulation.
Given the task allocation, each UAV constantly updates the coverage path and incrementally extracts crucial information to support the exploration planning. 
A hierarchical planner finds exploration paths, refines local viewpoints and generates minimum-time trajectories in sequence to explore the unknown space agilely and safely.
The proposed approach is evaluated extensively, showing high exploration efficiency, scalability and robustness to limited communication.
Furthermore, for the first time, we achieve fully decentralized collaborative exploration with multiple UAVs in real world.
We will release our implementation as an open-source package\footnote{To be released at \url{https://github.com/HKUST-Aerial-Robotics/FUEL}}.
\end{abstract}


\IEEEpeerreviewmaketitle


\section{Introduction}
\label{sec:intro}

\IEEEPARstart{A}{utonomous} exploration, which utilizes autonomous vehicles to map unknown environments, is a fundamental problem for various robotic applications like inspection, search-and-rescue, etc.
Recently, a considerable amount of literature studies autonomous exploration with UAVs, especially quadrotors.
It is demonstrated that UAVs are particularly suited to exploring complex environments efficiently, thanks to their agility and flexibility.

\begin{figure}[t!]
	\centering
	\vspace{-2cm}
\end{figure} 

Despite the significant progress, most works solely focus on exploration with a single UAV, while little attention has been paid to multi-UAV systems.
However, using a fleet of UAVs has incredible potential, since it not only enables faster accomplishment of exploration, but also is more fault-tolerant than a single UAV.
In this work, we bridge this gap by introducing \textbf{RACER}, a \textbf{RA}pid \textbf{C}ollaborative \tnb{E}xplo\tnb{R}ation approach using a team of decentralized UAVs.

Up to now, developing a robust, flexible and efficient multi-UAV exploration system has been fraught with difficulties.
First of all, approaches that coordinate multiple robots typically rely on a central controller or require reliable communication among the robots.
Unfortunately, unreliable and range-limited communication is common in practice, making the coordination vulnerable and less effective. 
To improve the flexibility and robustness of the system, a decentralized coordination approach with fewer communication requirements is desired. 
Secondly, many multi-robot exploration approaches solely consider the allocation of frontiers or viewpoints. 
Because the actual regions explored by each UAV are not accounted for, the strategies often result in interference among robots and inequitable workload allocation. 
Also, the global coverage routes of the unknown space are not taken into account, so the UAVs may redundantly revisit the same regions, significantly reducing the overall efficiency.
To fully realize the system's potential, a more sophisticated collaboration strategy is required.
Lastly, each member of the system should be able to fully utilize its capability to fulfill the task.
For this purpose, it is crucial that each UAV plans motions quickly in response to environmental changes, allowing itself to navigate and collect information agilely, while avoiding collision with previously unknown obstacles and other agents in the system.

\begin{figure}[t!]
	\centering
	\vspace{-1.2cm}
\end{figure} 

\begin{figure}[t!]
	\begin{center}          
		\subfigure
    {\includegraphics[width=0.8\columnwidth]{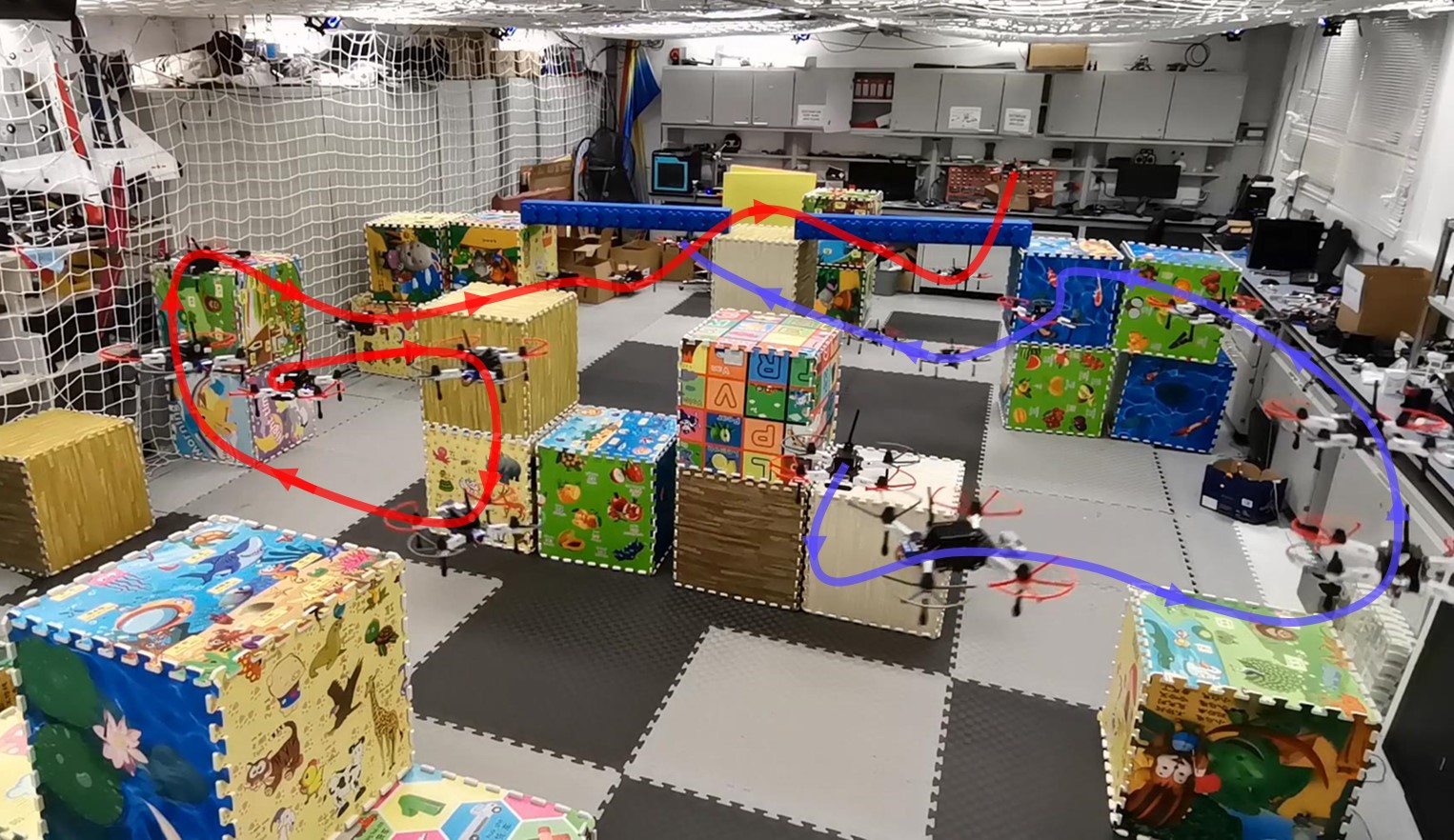}}     
		\subfigure
		{\includegraphics[width=0.625\columnwidth]{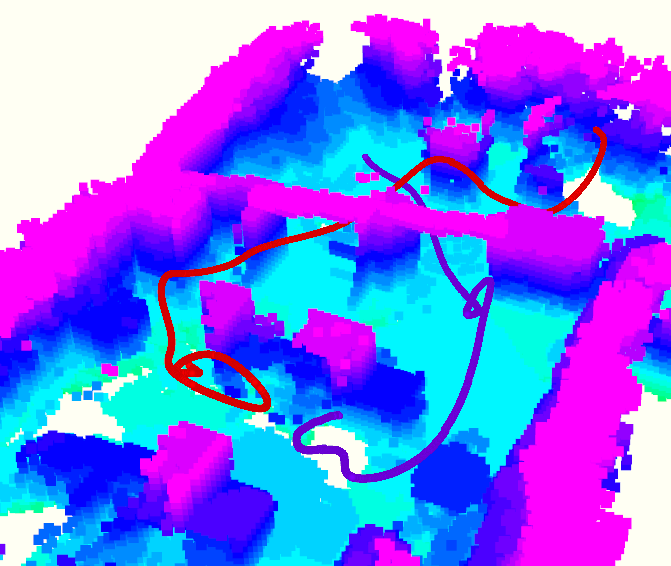}}    
		\vspace{-0.4cm}   
	\end{center}
   \caption{\label{fig:res-exp1} Two quadrotors simultaneously explore a complex unknown environment.
	 (a) The composite image of the experiment. (b) The visualization of the online built map and flight trajectories.
	 All quadrotors collaborate in a fully decentralized manner, i.e., no external computer is used to dispatch them,
	 each quadrotor runs the state estimation, mapping, coordination and planning algorithms independently on its onboard computer.
	 Video of the experiments is available at \url{https://www.dropbox.com/s/i5pcriool4su56r/racer.mp4?dl=0}. 
   }
	 \vspace{-1cm}
\end{figure}

In this paper, we present a systematic decentralized coordination and planning approach that enables rapid collaborative exploration, which takes the above-mentioned difficulties into consideration.
To effectively coordinate the quadrotors, the entire unknown space is constantly subdivided into hgrid online and distributed among the quadrotors by a pairwise interaction.
It only requires asynchronous and unreliable communication between pairs of nearby quadrotors, and ensures simultaneous exploration of distinct regions without causing interference among the quadrotors.
Moreover, a Capacitated Vehicle Routing Problem (CVRP) formulation is proposed for more efficient cooperation. 
It minimizes the lengths of multiple global coverage paths (CPs) and balances the workload partitioned to each quadrotor, effectively preventing repeated exploration and inequitable workload allocation.
Given the allocation of unexplored regions, each quadrotor updates its CP and extracts frontier information structures (FISs) incrementally to facilitate the exploration planning. 
A hierarchical planner finds local paths and refines viewpoints to cover the frontiers under the guidance of the CP. 
Minimum-time trajectories are generated to visit the viewpoints while avoiding collision with obstacles and other quadrotors.
The overall approach is computationally cheap and scalable, which enables the team to react quickly to environmental changes, leading to consistently fast exploration.

We evaluate the proposed approach comprehensively by simulations and challenging real-world experiments.
Our approach is shown to complete exploration significantly faster than existing centralized and decentralized approaches, has more consistent performance under restricted communication, and is computationally efficient and scalable for a large team.
What's more, we integrate our approach with a decentralized state estimation module and conduct fully autonomous exploration in complex environments. 
To the best of our knowledge, we are the first to achieve such exploration capability in real world, where the quadrotors cooperate in a fully decentralized fashion and {build a dense map of the environment completely and quickly.} 
We will release the source code to benefit the community.
In summary, the contributions are:

1) A pairwise interaction based on an online hgrid decomposition of the unexplored space, which {ensures the quadrotors explore} distinct regions jointly using only asynchronous and restricted communication. 

2) A CVRP formulation that minimizes the lengths of global CPs and balances the workload assigned to each quadrotor. It further enhances the cooperation of the quadrotor team.

3) A hierarchical exploration planner, ensuring a group of quadrotors to explore efficiently and safely. It is significantly extended from our previous work\cite{zhou2021fuel} by incorporating the guidance of CPs and enabling multi-robot collision avoidance. 


4) We integrate the proposed approach with multi-robot state estimation and conduct fully autonomous exploration experiments in challenging real-world environments.
The source code of our system will be released.


In what follows, we review related works in Sect.\ref{sec:related} and overview our system in Sect.\ref{sec:overview}. 
The hgrid-based pairwise interaction and the CVRP formulation for balanced workload allocation are detailed in Sect.\ref{sec:hgrid} and Sect.\ref{sec:cvrp} respectively. 
The hierarchical exploration planning is presented in Sect.\ref{sec:expl_plan}. 
Benchmark and experimental results are given in Sect.\ref{sec:results}. 
Sect.\ref{sec:conclude} concludes this article.


\section{Related Work}
\label{sec:related}

\begin{figure}
  \centering
  \vspace{-1.56cm}
\end{figure}

\subsection{Autonomous Exploration}
\label{subs:related_exploration}

Robotic exploration, which maps unknown environments with mobile robots, has been investigated over the years. Some works concentrate on quick coverage\cite{cieslewski2017rapid, dharmadhikari2020motion}, while others place more emphasis on precise reconstruction\cite{schmid2020efficient, song2017online}. One type of classic approaches is frontier-based approaches, which are presented earliest in \cite{yamauchi1997frontier} and assessed more comprehensively later in \cite{julia2012comparison}. {Shen \textit{et al.}\cite{shen2012stochastic} adapted a stochastic differential equation-based strategy, seeking frontiers in 3D environments.} Different from the original method\cite{yamauchi1997frontier} that chooses the closest frontier as the next target, {Cieslewski \textit{et al.}\cite{cieslewski2017rapid} chooses} the frontier inside the FOV that minimizes the change of velocity. This strategy is beneficial to maintain a high flight speed and demonstrates higher efficiency than \cite{yamauchi1997frontier}. {Deng\textit{et al.}\cite{deng2020robotic} presented} a differentiable formulation, which is useful for gradient-based path optimization. {Frontier-based methods have the advantage of simplicity. However, the information gain of each candidate target is not directly available, in comparison to the sampling-based approaches.} 

{Sampling-based approaches, on the other hand, randomly produce candidate viewpoints and evaluate their information gain to explore the space. These approaches can explicitly quantify the information gain of each candidate viewpoint, at the cost of higher computational burden.} Sampling-based approaches are closely related to next best view (NBV) \cite{connolly1985determination}, where viewpoints are computed repeatedly to model a scene completely. {Bircher \textit{et al.}\cite{bircher2016receding} first introduced} the concept of NBV in 3D exploration. It grows RRTs within free space and executes the most informative edge in a receding horizon fashion. Later, uncertainty of localization \cite{papachristos2017uncertainty}, visual importance \cite{dang2018visual} and inspection \cite{bircher2018receding} are considered under the framework in \cite{bircher2016receding}. To reuse past information, roadmaps capturing navigation history are built in \cite{witting2018history, wang2019efficient}, while a single tree is persistently refined by employing a rewiring scheme motivated by RRT*\cite{schmid2020efficient}. To realize faster flight, {Dharmadhikari \textit{et al.}\cite{dharmadhikari2020motion} proposed to} sample dynamically feasible motion primitives to enable higher flight speed.  

Approaches combining the frontier-based and sampling-based approaches are presented in \cite{charrow2015information,selin2019efficient,meng2017two,song2017online,caoexploring,yang2021graph, zhou2021fuel}. {{Charrow \textit{et al.}\cite{charrow2015information} and Selin \textit{et al.}\cite{selin2019efficient}} proposed to} generate global paths based on frontiers and sample paths to explore local regions. {Meng \textit{et al.}\cite{meng2017two} discussed a method that samples viewpoints around frontiers and finds the global shortest tour passing through them. Song \textit{et al.}\cite{song2017online}} finds inspection paths completely covering frontiers by a sampling-based algorithm. {A two-level framework \cite{caoexploring} samples viewpoints for incomplete surfaces around the robots and computes the shortest visiting path, which is connected to a coarse global path. Comparing to sampling-based methods, these methods are computationally cheaper as they require much fewer viewpoint sampling, thanks to the targeted sampling around frontiers. Our previous work \cite{zhou2021fuel} presented a more computationally efficient method that detects frontier and samples viewpoints around frontiers in an incremental fashion.}

{Given the candidate targets (frontiers or sampled viewpoints), many existing methods adopt a greedy strategy to make decisions. For example, the closest frontier\cite{yamauchi1997frontier,shen2012stochastic} is selected or the immediate information gain is maximized\cite{bircher2016receding,witting2018history,schmid2020efficient,dharmadhikari2020motion}. The greedy strategy ignores an efficient global route, resulting in unnecessary revisiting and reducing the overall efficiency. This issue is partially addressed by \cite{meng2017two} and our previous work \cite{zhou2021fuel}, which formulate a Traveling Salesman Problem to compute the shortest path visiting candidate viewpoints. However, they only consider paths for sampled viewpoints and still lack a global route that cover the entire unexplored region. In this work, we extend \cite{zhou2021fuel} to incorporate an efficient coverage path of the entire unknown space, guiding each quadrotor to explore different unknown regions in a more reasonable order. }


\begin{figure*}[t!]
  \centering
  {\includegraphics[width=1.6\columnwidth]{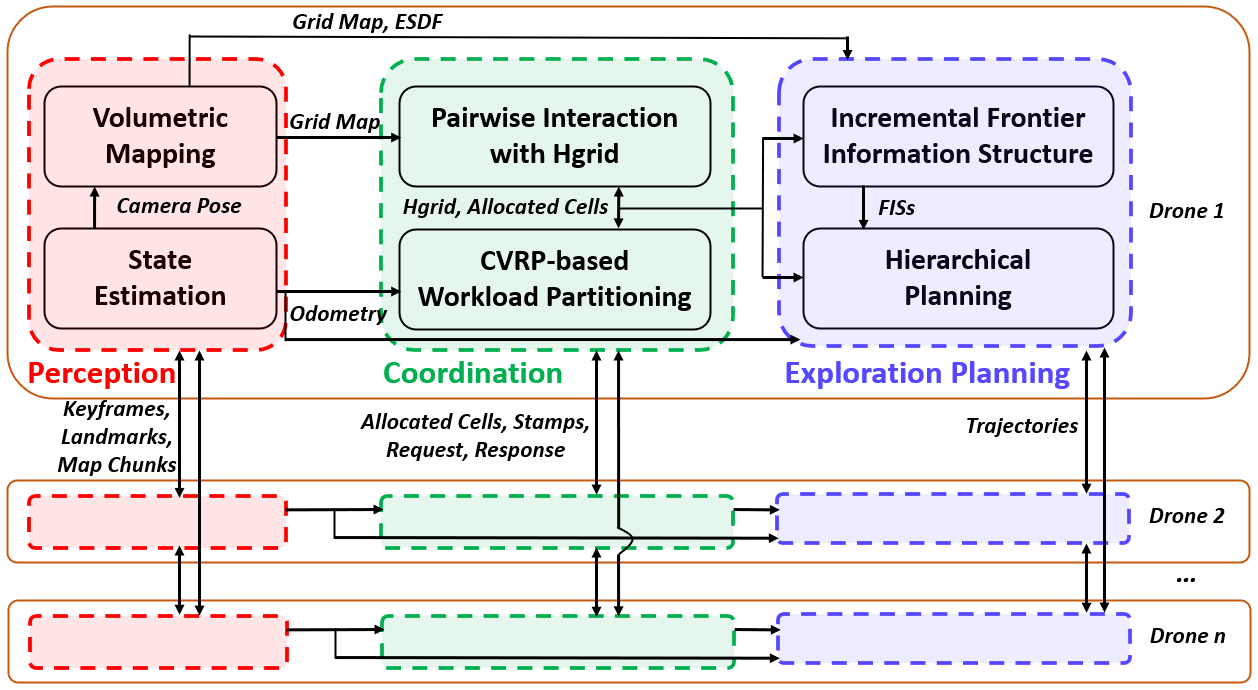}}       
  \caption{\label{fig:overview} An overview of our multi-quadrotor autonomous exploration system, including the perception, coordination and exploration planning modules.
  }
\end{figure*} 

\subsection{Coordinating Multiple Robots}

Multi-robot system has obvious advantages over a single robot, like covering environments faster and providing robustness to single point failures. Central to multi-robot exploration is the appropriate allocation of tasks, so that conflicting targets can be avoided and each robot explores different regions simultaneously. One common method is communicating with all robots and allocating tasks with a centralized server\cite{burgard2005coordinated, butzke2011planning, wurm2008coordinated, faigl2012goal, dong2019multi}. Among the approaches, the most straightforward one is iterative assignment \cite{burgard2005coordinated, butzke2011planning}, in which the algorithm greedily selects pairs of robot and target based on distance and information gain. However, such a strategy only allocates a single target to each robot, which may not be effective when multiple robots are assigned to nearby targets. To overcome this limitation, approaches based on segmentation\cite{wurm2008coordinated,karapetyan2017efficient} and multiple Travelling Salesman Problem (mTSP)\cite{faigl2012goal, dong2019multi, hardouin2020next} are proposed. {Wurm \textit{et al.}\cite{wurm2008coordinated} divided} the already explored area into segments using a Voronoi graph. By assigning robots to different segments instead of single targets, the robots are distributed more evenly over the environments. {Similarly, a breath-first-search-like clustering algorithm groups the decomposed areas and assign them to robots\cite{karapetyan2017efficient}.} {Faigl \textit{et al.}\cite{faigl2012goal} and Dong \textit{et al.}\cite{dong2019multi} formulated} the coordination as mTSP. Since the problem is NP-hard, divide-and-conquer schemes are proposed to find the approximate solutions. {Faigl \textit{et al.}\cite{faigl2012goal} grouped} all targets into clusters by a variant of the K-means algorithm and assigns the clusters to robots. {Dong \textit{et al.}\cite{dong2019multi} solved} a discrete optimal transport problem to find a promising assignment of all targets to robots, while the optimal path of each robot is determined by TSP. {Hardouin \textit{et al.}\cite{hardouin2020next} adopted a TSP-greedy allocation algorithm to greedily assign a cluster of viewpoint to each robot. However, it also considers a centralized architecture where communication is assumed to be perfect.}   

One major issue of centralized coordination is requiring all robots to maintain communication with the server. However, the requirement is usually impractical, for example, robots may get far from the server while exploring large-scale environments, or signals may be frequently occluded in cluttered environments. Moreover, robots are unable to continue exploration if the server fails. In contrast, decentralized coordination \cite{yamauchi1999decentralized, zlot2002multi, smith2018distributed, berhault2003robot, corah2019distributed, yusmmr, klodt2015equitable} is more robust to communication loss and failure, since each agent in the system is able to operate independently. The first decentralized multi-robot exploration is developed in \cite{yamauchi1999decentralized}, in which all robots share map information and move to the closest frontiers. Although being simple and completely distributed, the coordination is not effective enough since more than one robot may move to the same place. A different means of decentralized coordination utilizes auction-based mechanisms\cite{zlot2002multi, smith2018distributed, berhault2003robot}. {Zlot \textit{et al.}\cite{zlot2002multi} and Smith \textit{et al.}\cite{smith2018distributed} settled} conflicts among robots using a distributed single-bid local auction. To jointly consider the allocation of multiple targets, {Berhault \textit{et al.}\cite{berhault2003robot} formulated} a combinatorial auction. Apart from auction-based approaches, {Corah \textit{et al.}\cite{corah2019distributed} extended} greedy assignment method to a distributed version, however, the procedure typically requires several rounds of communication among robots. {Corah \textit{et al.}\cite{corah2021volumetric} also discussed different objective functions and their applicability for the distributed assignment algorithm in\cite{corah2019distributed}. Yu \textit{et al.}\cite{yusmmr} utilized} a multi-robot multi-target potential field to dispatch robots to different frontiers. 
{A common drawback of these methods is that they require multiple robots to connect stably at the same time, which can be easily affected by unstable networks and become less effective. To resolve this issue, it is important to simplify the requirement for communication structure, reducing the number of negotiating robots.}
The most relevant approach to us is the pairwise optimization scheme presented by {Klodt \textit{et al.} \cite{klodt2015equitable}}, where targets are repeatedly reallocated among robots by pairwise interactions. {The method evolved from \cite{durham2011discrete}, which was originally designed to solve a robot coverage control problem.}

Our pairwise interaction is inspired by \cite{klodt2015equitable}. 
{However, we adopt hgrid as the elementary task unit to allow more effective task allocation.
Meanwhile, instead of approximating each pair's traveling distance and using a local hill descent to improve the allocation locally\cite{klodt2015equitable}, we find the optimal allocation between two robots by minimizing the exact traveling distance. We also design a request-response scheme to address the issues of inequitable interaction chance and conflicting interactions. }



\subsection{{Multi-robot Exploration Systems}}

{Several work studied multi-robot system for autonomous exploration, focusing on different aspects of the system. Motivated by the DARPA Subterranean Challenge, where teams of robots search for objects of interest in underground environments, researchers presented systems towards subterranean exploration using legged, wheeled and flying robots\cite{kulkarni2021autonomous,agha2021nebula,hudson2021heterogeneous,rouvcek2021system,ohradzansky2021multi,petravcek2021large}. Those robots are typically equipped with various sensors and high-gain communications antennas, as well as localization, mapping, and path planning capabilities, allowing them to search for objects in complex, large-scale subterranean environments. These work specialized in the integration of hardware and algorithmic components, as well as subterranean engineering adaptations for system robustness. However, the systems rely more or less on a central computer to accomplish tasks. Meanwhile, coordination among robots is paid with less attention and simple strategies like greedy assignment\cite{kulkarni2021autonomous} or auction\cite{hudson2021heterogeneous,ohradzansky2021multi} are adopted. Petr{\'a}{\v{c}}ek \textit{et al.}\cite{petravcek2021large} also discussed a homing strategy such that a group of robots is able to build up a communication tree with the base station. Apart from them, some researchers investigated the mapping\cite{corah2019communication,tian2020search} and collaborative communication\cite{cesare2015multi} of the system. Corah \textit{et al.}\cite{corah2019communication} employed a Gaussian mixture model for global mapping, which maintains a small memory footprint and enables a lower volume of communication. Tian \textit{et al.}\cite{tian2020search} presented a cycle consistency-based method for robust data association, improving the precision of collaborative localization and mapping. Cesare \textit{et al.}\cite{cesare2015multi} proposed a collaboration scheme that allows robots with low battery to take on the role of a relay to improve communication between team members. In \cite{mcguire2019minimal}, a swarm gradient bug algorithm for autonomous navigation of tiny flying robots is proposed. Each robot uses wall following and received signal strength to avoid collision. It allows light-weight robots to explore unknown environments at the cost of lower efficiency and unavailability of accurate online mapping. Among these work, none of them studied decentralized algorithms that allow consistent effective collaboration even under intermittent communication. Also, none of the work minimizes the overall coverage path length and balances the workload to exploit the system's full capability. } 


\begin{figure}
  \centering
  \vspace{-1.24cm}
\end{figure}

\subsection{Quadrotor Motion Planning}
\label{subs:related_trajectory}

Relevant to exploration is local trajectory replanning, where gradient-based methods gain wide popularity\cite{oleynikova2016continuous, fei2017iros, usenko2017real, zhou2019robust, zhou2020robust, zhou2020raptor, zhou2020ego}. {The methods were revived by Ratliff \textit{et al.}\cite{ratliff2009chomp}} and extended to polynomial trajectories\cite{oleynikova2016continuous,fei2017iros} and B-spline trajectories\cite{usenko2017real}. More recently, {Zhou \textit{et al.}\cite{zhou2019robust, zhou2020robust,zhou2020raptor} further exploited the properties of B-splines, and used} topological guiding paths and active perception to achieve aggressive flight in complex scenes. {The computation time of trajectory optimization is further reduced} by eliminating the need of distance field\cite{zhou2020ego}.

To avoid collision between robots, decentralized approaches are studied in \cite{van2011reciprocal, van2011reciprocal2, arul2020dcad, liu2020mapper,park2020efficient,zhou2020ego2}. Velocity obstacles are leveraged to avoid collision of robots with different constraints\cite{van2011reciprocal, van2011reciprocal2,arul2020dcad}. {Liu \textit{et al.}\cite{liu2020mapper} proposed} an asynchronous strategy to avoid obstacles and other vehicles. {Park \textit{et al.}\cite{park2020efficient} employed} safe flight corridor and Bernstein polynomial to generate feasible trajectories. {Zhou \textit{et al.}\cite{zhou2020ego2} extended} the gradient-based replanning method \cite{zhou2020ego} to a team of quadrotors. Our trajectory planning in this work is based on the method in \cite{zhou2019robust, zhou2021fuel}, which generates safe and minimum-time trajectories for fast exploration. We extend it to avoid inter-drone collision using asynchronous communication among quadrotors.


\section{System Overview}
\label{sec:overview}

The overall multi-quadrotor exploration system is illustrated in Fig. \ref{fig:overview}. As most recent works\cite{bircher2016receding,cieslewski2017rapid,schmid2020efficient} do, we aim to build a complete volumetric map of a designated space. Each quadrotor in the system performs decentralized state estimation to localize itself and other quadrotors. In experiments, the state estimation approach in \cite{xu2021omni} is integrated into our system. Given the estimated relative pose, each quadrotor exchanges map information frequently with nearby ones when communication is available, to allow more informed decision making. The implementation details of state estimation and mapping are presented in Sect.\ref{sec:implement}. 

The coordination of the quadrotor team relies on the online hgrid decomposition and pairwise interaction. As each quadrotor explores and collects information, {the entire unknown space, which is bounded by a predefined boundary as most methods,} is constantly decomposed into hgrid containing cells of varying sizes, which serve as elementary task units to be allocated among the quadrotors. Scheduled by a request-response scheme, pairs of quadrotors communicate and update the ownership of the cells by pairwise interaction (Sect.\ref{sec:hgrid}). The cells are partitioned by a CVRP formulation that minimizes the lengths of the CPs and balances the volume of unknown space to be explored, as is presented in Sect.\ref{sec:cvrp}.

Given the allocated cells, each quadrotor independently plans paths and trajectories to explore the designated space while avoiding other quadrotors and obstacles. In the assigned regions, it computes the shortest CPs and updates the FISs incrementally as the map changes. The CPs serve as high-level guidance of the exploration, along which optimal local paths passing through a local set of frontier clusters are found. Finally, minimum-time trajectories are generated to cover the viewpoints and avoid collision (Sect.\ref{sec:expl_plan}). The process continues until no frontier is discovered inside the allocated cells and no other cells are assigned to the quadrotor.


\section{Hgrid-based Pairwise Interaction}
\label{sec:hgrid}

We utilize a pairwise interaction to distribute tasks, which only requires intermittent communication with nearby quadrotors. Unlike most approaches that allocate frontiers and viewpoints, our task representation is based on an online hgrid decomposition of the unknown space, which ensures joint exploration of distinct regions without causing interference among quadrotors. 



\subsection{Hgrid Decomposition}
\label{subsec:hgrid0}

\begin{figure}[t!]
  \centering
  {\includegraphics[width=0.75\columnwidth]{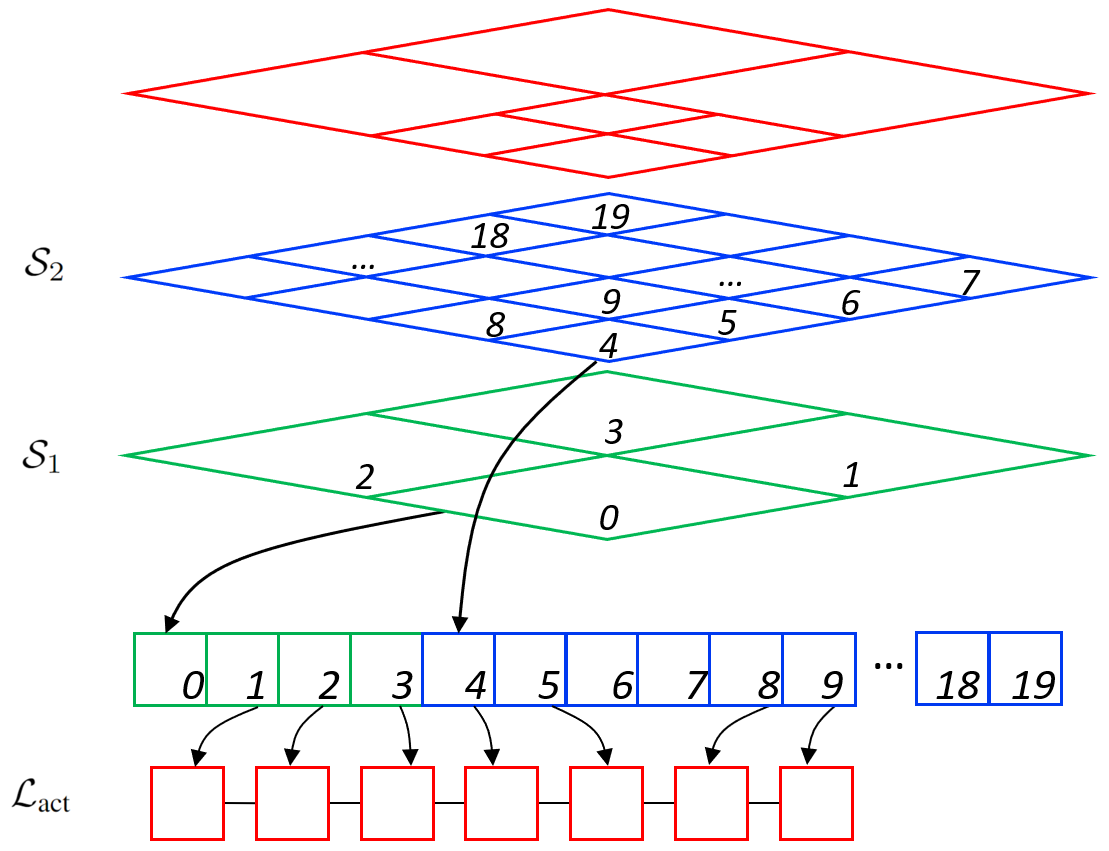}}       
  \caption{\label{fig:hgrid} {An example of a 2D hgrid decomposition with 2 levels, and its implementation using a single array. For convenience we illustrate 2D hgrid here, but in implementation 3D hgrid is used. The current decomposition of the environment (red) is recorded by list $\mathcal{L}_{act}$.}
  }
  \vspace{-1.0cm}
\end{figure}

We constantly decompose the entire unknown space online into a collection of disjoint regions, representing elementary task units to be allocated among quadrotors. The considered decomposition here is hgrid\cite{ericson2004real}, which is a hierarchical decomposition consisting of cells from coarse to fine resolutions. Any cell in level $l$ is uniformly subdivided into $8$ sub-cells ($4$ in 2D) in level $l+1$. The varying cell sizes of hgrid provides sufficient flexibility to represent the unknown regions. Examples of hgrid are shown in Fig.\ref{fig:hgrid} and \ref{fig:update_hgrid}.

Given the online built map, each quadrotor maintains a hgrid decomposition located in the same world coordinate frame. {The establishment of a common coordinate frame is detailed in Sect.\ref{sec:implement}}. The hgrid contains a total number of $L$ levels $ \left\{ \mathcal{S}_1, \mathcal{S}_2, \cdots, \mathcal{S}_L \right\} $, where $ \mathcal{S}_1 $ has the coarsest cells and $ \mathcal{S}_L $ has the finest ones. For each cell, it records the number of unknown voxels contained by it and the centroid of the unknown voxels. {To store the entire hgrid cells with a single array, we use a mapping similar to that of a standard 3D grid. Let the space is discretized into $ h_l = h_{l,x} \times h_{l,y} \times h_{l,z} $ cells in $ \mathcal{S}_l $. For a cell whose integer coordinate is $ \left( i_l,j_l,k_l \right) $, its index is determined by:
\begin{equation}
  \label{equ:index2}
  \gamma(i_l,j_l,k_l) = i_l h_{l,y} h_{l,z} + j_l h_{l,z} + k_l + \sum_{m = 1}^{l} h_{m-1}, \ \ h_0 = 0 
\end{equation}}
Geometrically, hgrid is similar to an {octree-based decomposition\cite{meagher1982geometric2,wurm2010octomap}}, but it is implemented using arrays, allowing $ O(1) $ retrieval of any cell. {Moreover, the linearized data structure adopted in our implementation increases the level of data locality, which improves the CPU data cache hit rate and thus overall performance\cite{ericson2004real}.}



\subsection{Online Hgrid Updating}
\label{subsec:hgrid}

{To represent the decomposition of the unknown regions that will be covered by the quadrotors, a list $ \mathcal{L}_{\text{act}} $ is utilized, as is illustrated in Fig.\ref{fig:hgrid}. During the exploration, it stores a subset of all hgrid cells that have not been completely explored, have not yet been subdivided into finer ones, and are reachable by the sensor.} Meanwhile, to facilitate the update of hgrid, we record $ \mathcal{B}_{m} = \left\{ B_1, B_2, \cdots, B_{M} \right\} $, which are the axis-aligned bounding boxes (AABBs) of updated map regions, consisting of the region explored by a quadrotor itself and the ones shared by nearby quadrotors.

\begin{figure}[t!]
  \centering
  {\includegraphics[width=0.85\columnwidth]{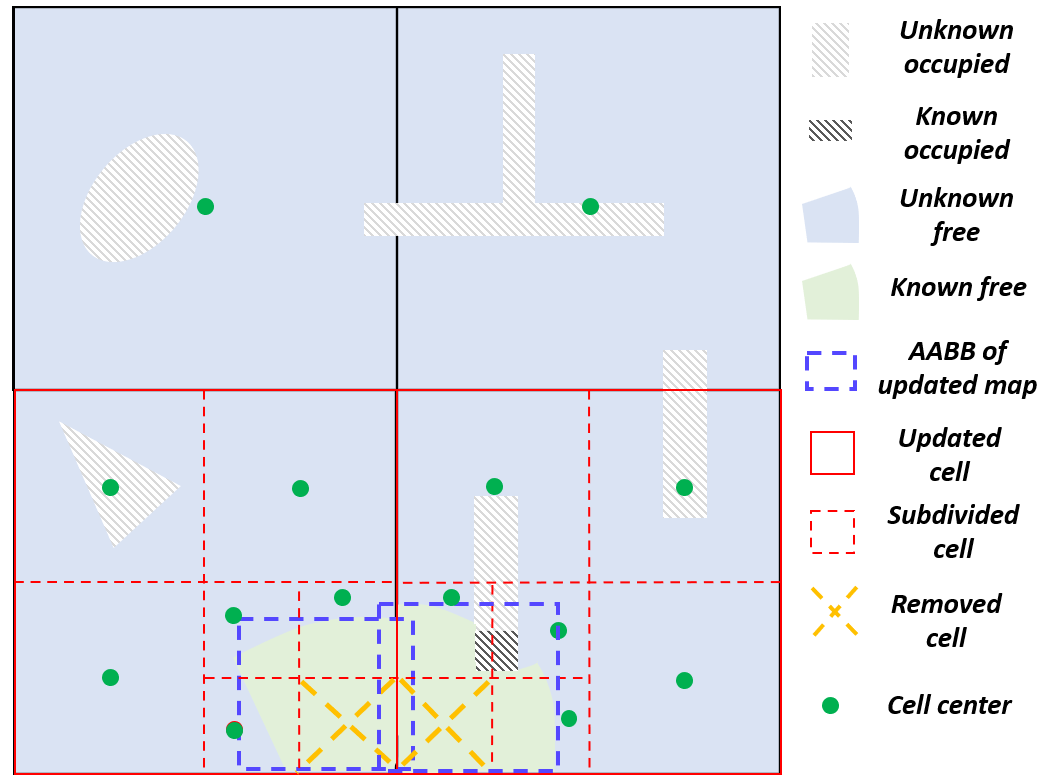}}       
  \caption{\label{fig:update_hgrid} An illustration of the procedure to update a 3-level hgrid.
    Cells that overlap with any AABBs in $ \mathcal{B}_m $ are updated and subdivided into finer cells if necessary. Cells that have been covered completely are removed from $\mathcal{L}_{\text{act}}$.
  }
\end{figure}

The online hgrid decomposition of unknown space is illustrated in Fig.\ref{fig:update_hgrid} and Alg.\ref{alg:update_hgrid}. Initially, the environment is completely unknown and is decomposed into the coarsest cells, i.e., $ \mathcal{L}_{\text{act}} $ only contains cells in $\mathcal{S}_1$. As the quadrotors build and share map, some regions become partially unknown and the associated cells are further subdivided to represent the yet unknown regions with finer resolutions. The update of hgrid proceeds by iterating each cell $ s_k $ in $ \mathcal{L}_{\text{act}} $ and screening the ones that have overlap with any $ B_m \in \mathcal{B}_m $ (Line 2). The centroid and the unknown voxel number of them are recomputed (Line 4). Then, if a certain ratio $ \alpha_{u} $ of voxels contained by $ s_k $ are already known, and $s_k$ is not at the finest level, it is subdivided uniformly (Line 5-9). Correspondingly, it means removing $ s_k $ from $ \mathcal{L}_{\text{act}} $ and appending the subdivided cells in the next level to $ \mathcal{L}_{\text{act}} $. Meanwhile, if $ s_k $ is at the finest level and contains only a tiny number of unknown voxels less than $\delta_{u}$, it is removed from $ \mathcal{L}_{\text{act}} $ (Line 11-12). Finally, cells whose centroids are unreachable from the current position of the quadrotor are also removed (Line 13-14), which eliminates regions that can not be covered by sensors, like hollow spaces in thick walls.

\begin{algorithm}[t!]
  \DontPrintSemicolon
  \tcp{$\mathcal{L}_{\text{act}} = \mathcal{S}_1 $ at the beginning of exploration}
  \For{$ s_k \in \mathcal{L}_{\text{act}} $}{
    \If{$ \lnot \ \tnb{HaveOverlap}(s_k, \mathcal{B}_m) $}{
      $ \tnb{continue} $\;
    }
    $ \tnb{UpdateCellInfo}(s_k) $\;
    \If{$ \tnb{KnownRatio}(s_k) \ge \alpha_{u} \land \tnb{Level}(s_k) \neq L $}{
      $ S_{\tn{div}} \gets \tnb{FindSubdivided}(s_k) $\;
      $ \mathcal{L}_{\text{act}}.\tnb{erase}(s_k) $\;
      \For{$ s_j \in S_{\tn{div}} $}{
        $ \mathcal{L}_{\text{act}}.\tnb{append}(s_{j}) $\;
      }
    }
    \If{$ \tnb{UnknownNum}(s_k) < \delta_{u} \land \tnb{Level}(s_k) = L $}{
      $ \mathcal{L}_{\text{act}}.\tnb{erase}(s_k) $\;
      }
    \If{$ \lnot \ \tnb{Reachable}(q_i) $}{
        $ \mathcal{L}_{\text{act}}.\tnb{erase}(s_k) $\;
    }
  }
\caption{Online Hgrid Decomposition. \label{alg:update_hgrid}}
\end{algorithm}


\subsection{Pairwise Interaction with Hgrid}
\label{subsec:pairwise}

The pairwise interaction is inspired by \cite{klodt2015equitable}, in which the problem of task allocation among all quadrotors is decomposed into subproblems that can be solved by pairs of quadrotors. {When two robots interact, a local hill descent based on their previous allocation of frontiers is performed by moving one frontier between them, which reduces an approximated path length locally.} In this way, tasks are allocated in a distributed fashion, while communication requirements are reduced significantly as only two quadrotors have to communicate at a time.

{Our pairwise interaction takes some steps further. Considering the hgrid cells for any interacting quadrotors $q_i$ and $q_j$, it finds the optimal allocation between them by minimizing the exact path length required to cover the cells. Specifically,} let $ \mathcal{T}_i $ and $ \mathcal{T}_j $ be the two sets of cells before interaction, the target is to find two new sets $ \mathcal{T}_i^{\prime} $ and $ \mathcal{T}_j^{\prime} $, such that $ \mathcal{T}_i \cup \mathcal{T}_j = \mathcal{T}_i^{\prime} \cup \mathcal{T}_i^{\prime} $ and $ \mathcal{T}_i^{\prime} $, $ \mathcal{T}_j^{\prime} $ {minimize the overall length of coverage paths.} 

{Without the schedule of a central computer, the quadrotors may interact with each other in an uneven manner, which is detrimental to effectively distributing the task. Furthermore, a quadrotor may happen to interact with multiple other quadrotors at the same time, producing several conflicting allocation results. To address the two issues,} a request-response scheme is designed, as shown in Alg.\ref{alg:pairwise}. As a preliminary, several types of information are {broadcast} periodically so that nearby quadrotors maintain a common knowledge about them. Take quadrotor $q_i$ as an example, {it broadcasts the hgrid cells that belong to it and the timestamp of the latest attempt to perform pairwise interaction, which are denoted as $ \mathcal{T}_i $ and $ q_i.T_{\text{att}}$ respectively.} Also, $ q_i $ records $ \left\{ T_{\text{succ},j} | j \in [1,N_q]\setminus i  \right\} $, the timestamps of the latest successful interaction with the other $ N_q-1 $ quadrotors.

The interaction starts by finding nearby quadrotors which have communication with $q_i$ (Line 4-5). {Those that have already attempted another interaction within a short period of time $ \varepsilon_{\tn{att}} $ will not be considered} (Line 6-7), preventing a quadrotor to interact with several ones simultaneously, which typically leads to conflicting interaction results. Among them, we select $ q_{\eta} $, the one that has not experienced successful interaction with $ q_i $ for the longest time (Line 8-9), which ensures that other quadrotors have similar opportunities to interact with $q_i$.Then an appropriate partitioning is found and a request message containing the result is sent to $ q_{\eta} $ (Line 10-11). The specific partitioning approach is detailed in Sect.\ref{sec:cvrp}. Finally, it updates $ T_{\text{att}} $ and waits for a response in a duration of $ \varepsilon_{\tn{att}} $ (Line 12-13). Only if a \textit{succ} response is received, the partitioning result is updated by both $q_i$ and $q_{\eta}$ (Line 14-24), otherwise, the interaction is ignored and new ones will be tried later. Note that a double check of recent interaction attempt is necessary (Line 19). 
{Sometimes $q_{\eta}$ may just request an interaction with another quadrotor $q_j$ and broadcast the updated timestamp. Due to communication latency, this timestamp may be unavailable to $q_i$ when it attempts to interact with $q_{\eta}$. In this case, the double check ensures that $q_{\eta}$ does not interact with $q_i$ and $q_j$ at the same time, avoiding conflicting task allocation.}



\begin{algorithm}[t!]
  \DontPrintSemicolon
  \SetKwProg{Fn}{\normalfont{Function}}{:}{}
  \SetKwFunction{FRequest}{\tnb{RequestInteraction}}
  \Fn{\FRequest{}}{
      $ T_n \gets \tnb{TimeNow}() $\;
      $ T_{\tn{max}} \gets 0 $\;
      $ \mathcal{Q}_{\tn{n}} \gets \tnb{NearbyQuadrotors}() $\;
      \For{ $ q_k \in \mathcal{Q}_{\tn{n}} $ }{
        \If{$ T_n - q_k.T_{\tn{att}} \le \varepsilon_{\tn{att}} $}{
          $\tnb{continue}$\;
        }
        \If{$T_n - T_{\tn{suc},k} > T_{\tn{max}}$}{
          $ T_{\tn{max}} \gets T_n - T_{\tn{succ},k}, \ \eta \gets k $\;
        }
      }
      $ \mathcal{T}_i^{\prime}, \mathcal{T}_{\eta}^{\prime} \gets \tnb{FindPartitioning}(\mathcal{T}_i, \mathcal{T}_{\eta }) $\;
      $ \tnb{Request}(\mathcal{T}_i^{\prime}, \mathcal{T}_{\eta}^{\prime}, T_n, i, \eta) $\;
      $ T_{\text{att}} \gets T_n $\;
      $ \textit{res} \gets \tnb{WaitResponse}(\varepsilon_{\tn{att}}) $\;
      \If{$ \textit{res} == \textit{succ} $}{
        $ \ T_{\tn{succ},i} \gets T_n $\;
        $ \tnb{UpdatePartitioning}(\mathcal{T}_i^{\prime}, \mathcal{T}_{\eta}^{\prime}, i, \eta) $\;
      }
  }
  \;
  \SetKwFunction{FRespond}{\tnb{RespondInteraction}}
  \Fn{\FRespond{$\mathcal{T}_i^{\prime}, \mathcal{T}_{\eta}^{\prime}, T_n, i, \eta$}}{
      \If{$ T_n - T_{\tn{att}} \le \varepsilon_{\tn{att}} $}{
          $ \tnb{Respond}(\textit{fail}) $\;
          \KwRet\;
      }
      $ \tnb{Respond}(\textit{succ}) $\;
      $ T_{\tn{att}} \gets T_n, \ T_{\tn{succ},i} \gets T_n $\;
      $ \tnb{UpdatePartitioning}(\mathcal{T}_i^{\prime}, \mathcal{T}_{\eta}^{\prime}, i, \eta) $\;
  }
\caption{Pairwise Interaction with Hgrid. \label{alg:pairwise}}
\end{algorithm}

\begin{figure}[t!]
  \centering
\vspace{-1.8cm}
\end{figure}

{Theoretically, the pairwise interaction may find suboptimal allocation compared with a centralized counterpart. However, with a single round of interaction among the robots, it quickly reaches a promising allocation, as is shown in our quantitative study in Sect.\ref{subsec:suboptimal}.}
{Also, note that the pairwise interaction does not necessarily reduce the communication events. Instead, it simplifies the communication structure, requiring only two robots to negotiate at the same time, which is more resilient to unstable communication.}




\section{CVRP-based Workload Partitioning}
\label{sec:cvrp}

To appropriately partition the hgrid cells belonging to a pair of interacting quadrotors (Alg.\ref{alg:pairwise}), a CVRP formulation is devised.
The key idea is minimizing the overall lengths of the quadrotors' coverage paths(CPs) and balancing the amount of unknown space allocated to them, making them collaborate more effectively. 

\begin{figure}[t!]
  \centering
  \subfigure[]{\includegraphics[width=0.49\columnwidth]{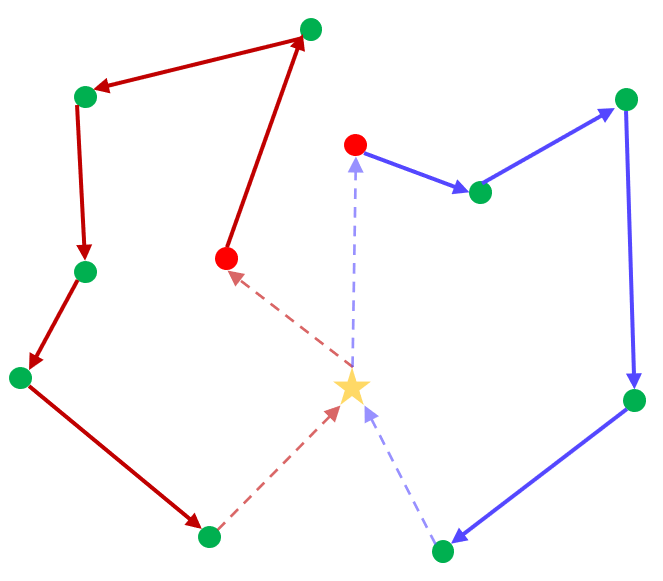}}       
  \subfigure[]{\includegraphics[width=0.49\columnwidth]{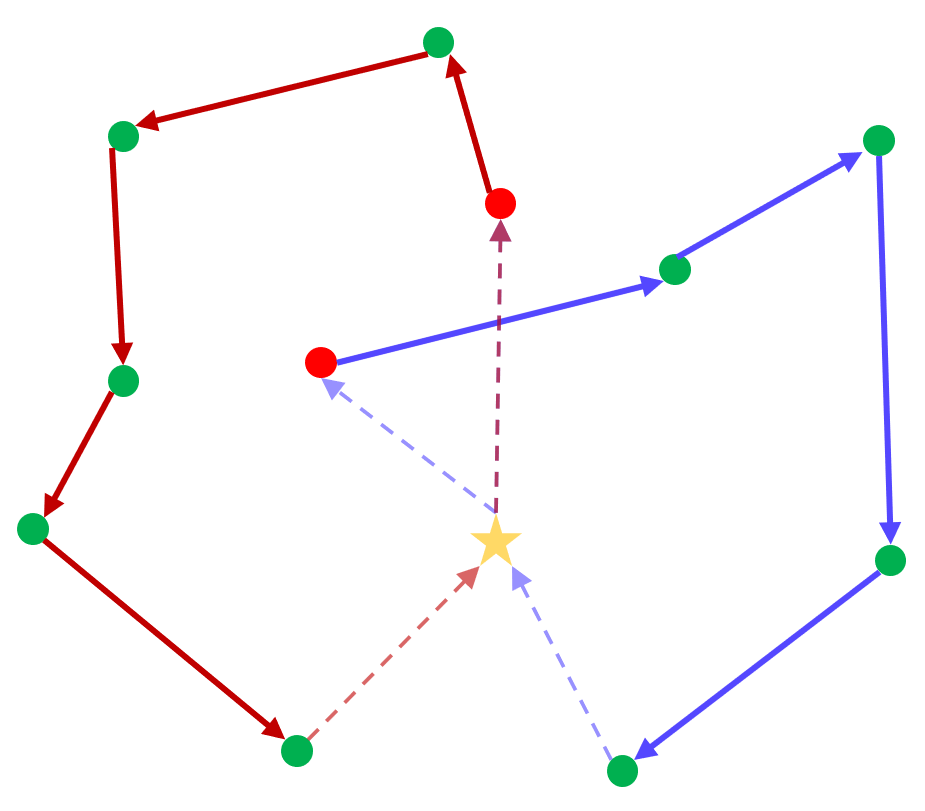}}       
  \caption{\label{fig:cvrp} Illustration of the Asymmetric VRP.
  Nodes associated with the virtual depot, quadrotors and hgrid cells are shown in yellow star, red circles and green circles respectively.
  The sequentially connected edges shown in red and purple represent the two routes that minimize the overall cost.
  (a) and (b) are two alternative solutions with identical cost.
  The desired coverage paths are obtained by removing the edges connecting with the virtual depot (displayed in dash).
  }
\end{figure}

\subsection{CVRP Formulation}
\label{subsec:cvrp}

The optimal CPs of the hgrid cells can be found by Vehicle Routing Problem (VRP)\cite{kusnur2021planning}, which generalizes the Traveling Salesman Problem (TSP) to multiple vehicles.
In our setting, there are two vehicles corresponding to the pair of interacting quadrotors $q_1, q_2$.
Different from the standard VRP, where there is a central depot and the vehicles' routes form closed loops, we require open paths starting from the quadrotors' positions and passing the hgrid cells. 
We reduce this variant into an Asymmetric VRP by introducing a virtual depot and properly {designing} the connection costs.

\subsubsection{Cost Matrix}
Suppose there are $ N_h $ hgrid cells, the Asymmetric VRP involves $ N_h +3 $ nodes, in which there are $ N_h $ nodes for the hgrid cells, $ 2 $ nodes for $q_1$, $q_2$ and one for the virtual depot.  
The associated cost matrix $ \mathbf{C}_{\text{avrp}} \in \mathbb{R}^{(N_h +3)\times(N_h +3)} $ has the form:
\begin{align}
  \mathbf{C}_{\text{avrp}} = \left[ \begin{array}{ccc}
    0 & -\mathbf{M}_{\text{inf}} & \mathbf{0} \\
    \mathbf{0}  & \mathbf{0} & \mathbf{C}_{q,h} \\
    \mathbf{0} & \mathbf{C}_{q,h}^{\text{T}} & \mathbf{C}_{h} \end{array} \right]
\end{align}

\begin{figure}[t!]
  \centering
  \subfigure{\includegraphics[width=0.85\columnwidth]{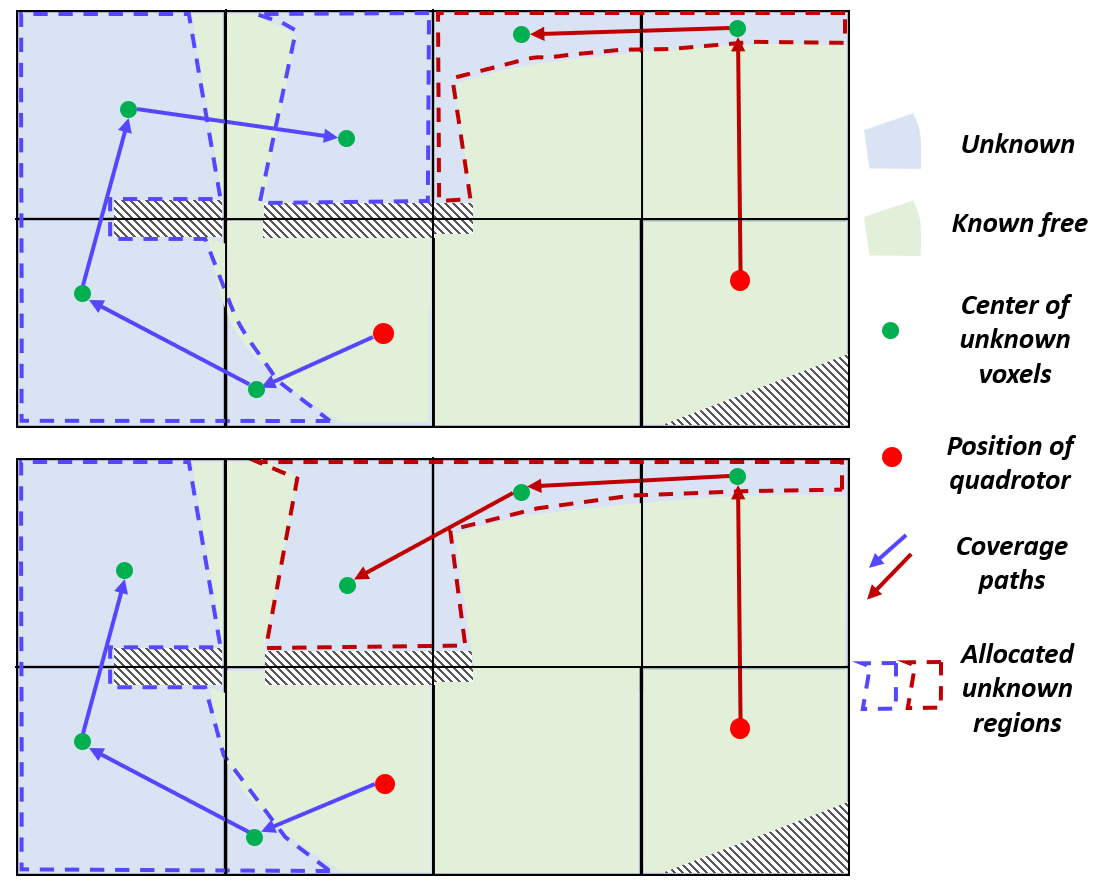}}       
  \caption{\label{fig:capacity} 
  The coverage paths of two quadrotors and the corresponding allocation of unknown space.
  Top: only the total path length is considered. The allocation of unknown space is unbalanced, in which the red one could be completely explored much faster.
  Bottom: the workloads of two quadrotors are more equitable by introducing the capacity constraints.
  }
\vspace{-0.5cm}
\end{figure}
$ \mathbf{C}_h $ is a $ N_h \times N_h $ block that corresponds to the connection costs among all hgrid cells.
Since we aim to find the shortest CPs, path lengths between pairs of cells are used as the connection costs:
\begin{align}
\label{equ:cell_cell}
  \mathbf{C}_h&(h_1, h_2) = \mathbf{C}_h(h_2, h_1) \\ \nonumber
  & = \tnb{Len}\left[ P(\mathbf{c}_{h_1},\mathbf{c}_{h_2}) \right], \ h_1, h_2 \in \left[ 1, N_h \right]
\end{align} 
Here $ \mathbf{c}_{h_1}, \mathbf{c}_{h_2} $ represents the centroid of unknown voxels in the $h_1$-th and $h_2$-th cells, while $ P(\mathbf{c}_{h_1},\mathbf{c}_{h_2}) $ is the collision-free path, which can be found by standard path searching algorithms.

The costs between quadrotors and hgrid cells are accounted for by $ \mathbf{C}_{q,h} $, which is a $ 2 \times N_h $ block.
The connection cost is similar to Equ.\ref{equ:cell_cell} except for a consistency term:
\begin{align}
  & \mathbf{C}_{q,h}(i,h) = \tnb{len}\left[ P(\mathbf{p}_{q_i},\mathbf{c}_{h}) \right] +
  c_{\text{con}}(i,h), \\ \nonumber
  & h \in \left[ 1, N_h \right], \ i \in \left[1,2\right]
\end{align} 
\begin{equation}
\label{equ:consistency}
  c_{\text{con}}(i,h) =\begin{cases}
    \beta_{\text{con}}, & \parbox[t]{0.5\columnwidth}{$q_i$ is connected directly to the $h$-th cell in the last CP} \\
    0, & \text{Else} \\
    \end{cases}
\end{equation}
It has been observed that there are sometimes multiple solutions with comparable lengths but distinctive coverage patterns (Fig.\ref{fig:cvrp}).
Therefore, $ c_{\text{con}}(\cdot) $ is introduced to prevent the paths from changing frequently among different patterns, which could result in inconsistent movements and slow down the exploration.

To reduce our problem to an Asymmetric VRP, the connection costs from the virtual depot to the two quadrotors are assigned with the block $ -\mathbf{M}_{\text{inf}} = -\left[ M_{\text{inf}}, M_{\text{inf}} \right] $, where $ M_{\text{inf}} $ is a huge value.
The large negative costs make the virtual depot's node connect the two quadrotors' directly, since it immensely reduces the overall cost of the output routes. 
In this way, the output of the Asymmetric VRP is composed of the desired shortest paths and four extra edges, in which two edges link the depot to quadrotors while the other two ones link two cells to the depot.
The reduction of our problem to the Asymmetric VRP is better illustrated in Fig.\ref{fig:cvrp}. 

\subsubsection{Capacity Constraints}
Although the length of routes is optimized, the actual amount of unknown areas explored by each quadrotor is not considered, which can still lead to unbalanced partitioning of workloads sometimes, as the example in Fig.\ref{fig:capacity} shows.
For this reason, we introduce \textit{capacity} constraints of vehicles to further balance the workload allocated to the quadrotors.
For each node $ \nu_k $ in the VRP, if it is associated with a hgrid cell (suppose the $h_k$-th cell), then a \textit{demand} $ \rho_k $ equal to the number of unknown voxels $ u_{h_k} $ is assigned. Otherwise, a zero demand is set. 
We restrict the capacity of each vehicle to a percentage of the total number of unknown voxels.
Denoting the set of nodes in the route of quadrotor $q_i$ as $ \mathcal{R}_i $, the capacity constraints are:
\begin{equation}
  \sum_{\nu_{k} \in \mathcal{R}_i } \rho_{k} \le  \alpha_{\rho} \sum_{h = 1}^{N_h} u_h, \ \ i =1,2
\end{equation}
With the capacity constraints, the problem becomes a CVRP.
To solve it, we utilize an {Lin-Kernighan-Helsgaun solver\cite{helsgaun2017extension} extended by Helsgaun}. 
The algorithm transforms the VRP into an equivalent standard traveling salesman problem (TSP), and uses penalty functions for handling the capacity constraints.
Although the problem is known to be NP-hard, our problem scale is small, for which optimal solution can be obtained in most cases\cite{helsgaun2017extension}.

\subsection{Sparse Graph for Path Searching}
\label{subsec:topo_graph}

The major overhead involved in the CVRP is the computation of $ \mathbf{C}_{h} $ and $ \mathbf{C}_{q,h} $, which requires $ O(N_h^2) $ path searching.
This can be considerably expensive in large-scale environments, when a large number of global paths should be searched directly on the volumetric map.
We take inspiration from \cite{yang2021graph, oleynikova2018sparse} to relieve this computation burden.
Specifically, a sparse graph embedded in the hgrid is maintained, which is leveraged to do global path searching.  
For each hgrid cell, a graph node is attached and the node of every adjacent cell is connected to it by a weighted edge, if a collision-free path exists within the ranges of the two cells.
Whenever a hgrid cell is updated, the path to every adjacent cell is recomputed on the volumetric map.
If a solution is found, the edge weight is updated by the path length.
Otherwise, the edge is removed from the graph. 

Leveraging this sparse graph, connection costs among hgrid cells and quadrotors can be computed more easily.
For every pair of hgrid cells, a path is searched on the sparse graph to approximate the shortest path on the volumetric map.
For a hgrid cell distant from the quadrotor, the cost is approximated by the path length from the quadrotor to the closest hgrid cell, plus the path length between the two cells on the sparse graph.
In practice, the approximated costs lead to comparable solutions of the CVRP comparing with using the exact costs, but substantially reduce the computation time.


\section{Exploration Planning}
\label{sec:expl_plan}

\begin{figure}[t!]
  \centering
  \subfigure{\includegraphics[width=0.7\columnwidth]{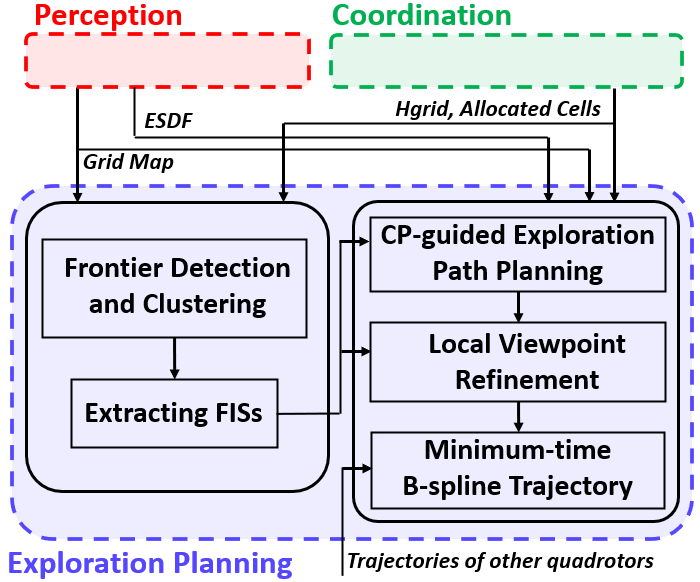}}       
  \caption{\label{fig:planning_procedure} The detailed components of the exploration planning module.
  }
  \vspace{-0.5cm}
\end{figure}

The exploration planning approach is based on \cite{zhou2021fuel}, which extract frontier information structures (FISs) incrementally and plan motions in several hierarchies. {In this work, we extend the method to a multi-UAV system, allowing effective and scalable team exploration. We also extend the exploration planning by incorporating the coverage paths (CPs) as high-level guidance, significantly improving the exploration rate. Collision avoidance under intermittent communication is also considered.} The key components of the exploration planning is shown in Fig.\ref{fig:planning_procedure}.


\subsection{Incremental Frontier Information Structure}
\label{subsec:frontier}

To facilitate the exploration planning, {we adopt the method in \cite{zhou2021fuel} to incrementally} detect frontiers\cite{yamauchi1997frontier}, which are known-free voxels adjacent to unknown ones. Detected frontiers are group into clusters and rich information are extracted (Alg.\ref{alg:frontier}). 
{The procedure is summarized and more details can be found in \cite{zhou2021fuel}. It starts by removing outdated frontier clusters in the updated map, in which a fast prescreening using the the clusters' AABBs is performed before a detailed check (Line 1-4). Next, new clusters are searched. Clusters formed by sensor noises are filtered, and large clusters are split into smaller ones so that each can be covered by a viewpoint (Line 5-9). }

\begin{algorithm}[t!]
  \DontPrintSemicolon
  \For{$ F_i \in \mathcal{F}_{\tn{v}} $}{
      \If{$ \tnb{HaveOverlap}(F_i, \mathcal{B}_m) $}{
        \If{$ \tnb{IsClusterChanged}(F_i) $}{
          $ \mathcal{F}_{\tn{v}}.\tnb{erase}(F_i) $\;
        }
      }
  }
  $ \mathcal{F}_{\tn{new}} \gets \tnb{DetectNewFrontierCluster}(\mathcal{B}_m) $\;
  \For{$ F_i \in \mathcal{F}_{\tn{new}} $}{
      \If{$ \tnb{VoxelNum}(F_i) < \varepsilon_{F,1}  $}{
        $\mathcal{F}_{\tn{new}}.\tnb{erase}(F_i)  $\;
      }
      {  $ \tnb{SplitLargeClusters}(F_i) $}\;
  }
  {$ \tnb{Separate}(\mathcal{F}_{\tn{new}}, \mathcal{F}_{a}, \mathcal{F}_{s}  ) $}\;
  \For{$ F_i \in \mathcal{F}_{a} $}{
      $ \mathbf{p}_{F_i} \gets \tnb{Centroid}(F_i) $\;
      { 
      $ \mathcal{V}_{F_i} \gets \tnb{SampleViewpoints}(\mathbf{p}_{F_i}) $\;
      $ \tnb{SortAndPrune}(\mathcal{V}_{F_i}, N_{\mathcal{V}_F}) $\;
      }
  }
  { $ \tnb{UpdateConnectionCosts}(\mathcal{F}_{\tn{v}}, \mathcal{F}_{a}) $ }\;
  $ \mathcal{F}_{\tn{v}} \gets \mathcal{F}_{\tn{v}} \cup \mathcal{F}_{\tn{new}} $\;
\caption{Frontier Information Structure. Frontier clusters are represented by $ \mathcal{F}_{\tn{v}} $. \label{alg:frontier}}
\end{algorithm}

\begin{figure}[t!]
  \centering
\vspace{-1.7cm}
\end{figure}


{For each frontier cluster $F_i$, the centroid is computed (Line 12), while viewpoints are uniformly sampled (Line 13), where each viewpoint $ \mathbf{q}_{i,j} $ is represented by the position and yaw\footnote{The roll-pitch-yaw angles convention is used, which rotates around the world frame's x-axis, then the y-axis, and finally the z-axis.} angle $ \left\{ \mathbf{p}_{i,j}, \varphi_{i,j}  \right\} $. Viewpoints with sufficient coverage of the cluster are sorted in $ \mathcal{V}_{F_i} $ in the order of descending coverage. Only the first $ N_{\mathcal{V}_F} $ viewpoints are preserved (Line 14).}
{Lastly, the connection cost between each pair of clusters is computed to assist the exploration planning (Sect.\ref{subsec:planning}). For two viewpoints $ \mathbf{q}_{k_1,j_1}, \mathbf{q}_{k_2,j_2} $, the time lower bound when moving between them is estimated as:}
\begin{align}
  t_{\text{lb}}&(\mathbf{q}_{k_1,j_1}, \mathbf{q}_{k_2,j_1}) = \max \bigg\{ \frac{\tnb{Len}\left[P \left( \mathbf{p}_{k_1,j_1}, \mathbf{p}_{k_2,j_2} \right)\right]}{v_{\text{max}}},  \\ \nonumber
  & \frac{\min\left( |\varphi_{k_1,j_1}-\varphi_{k_2,j_2}|, 2\pi-|\varphi_{k_1,j_1}-\varphi_{k_2,j_2}| \right)}{\dot{\varphi}_{\text{max}}} \bigg\}
\end{align} 
{where $ \tnb{Len}[P(\cdot)] $ is the length of a collision-free path and while $ v_{\text{max}}, \dot{\varphi}_{\text{max}} $ are the maximal velocity and yaw angle rate. The connection cost for clusters $ F_{k_1}, F_{k_2} $ is then given by $ t_{\text{lb}}(\mathbf{q}_{k_1,1}, \mathbf{q}_{k_2,1}) $.}

{In contrast to \cite{zhou2021fuel}, some adaptations are made for a multi-UAV system. Instead of only computing FISs within the map region updated by a quadrotor, regions shared by nearby quadrotors should also be considered to constantly keep a complete list of frontiers. The shared regions are continuously tracked by their AABBs $ \mathcal{B}_m $, as mentioned in Sect.\ref{subsec:hgrid}, to allow incremental updates. Compared to a single quadrotor, a greater number of clusters are detected, imposing a significant computational burden when extracting information for all of them. We circumvent this burden by storing clusters inside and outside the quadrotor's allocated working area in two separate lists $ \mathcal{F}_{\text{a}} $ and $ \mathcal{F}_{\text{s}} $ (Line 10). Information are only extracted for clusters in $ \mathcal{F}_{\text{a}} $, with no effect on exploration planning. If the allocated area changes after interaction (Sect.\ref{subsec:pairwise}), associated clusters are reallocated in the two lists accordingly.}

\begin{figure}[t!]
  \centering
  \subfigure[]{\includegraphics[width=0.7\columnwidth]{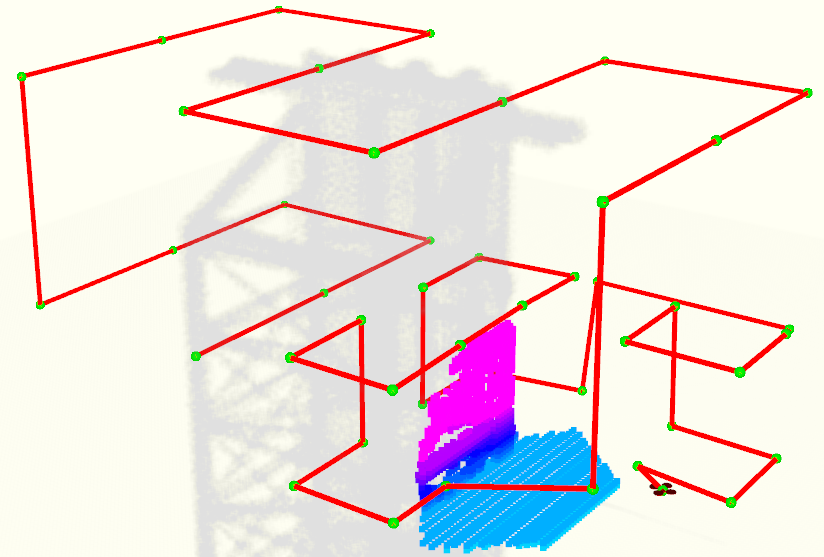}}       
  \subfigure[]{\includegraphics[width=0.75\columnwidth]{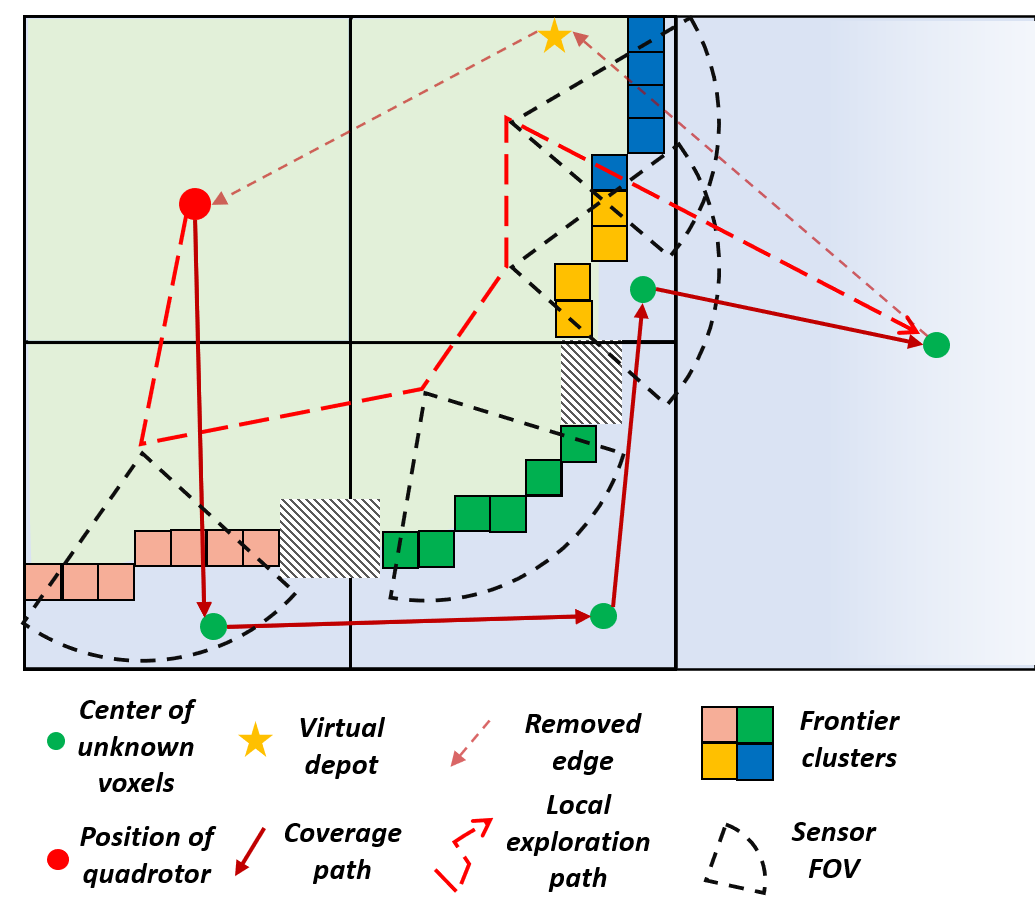}}       
  \caption{\label{fig:CP-guided} An illustration of CP-guided path planning.
  (a) A sample CP of the subdivided unknown space. 
  (b) A local path is computed to cover the frontier clusters along the CP ($ N_{\tn{CP}}=3 $).
  }
  \vspace{-0.6cm}
\end{figure}

\subsection{Hierarchical Planning}
\label{subsec:planning}

Our previous approach\cite{zhou2021fuel} employs a hierarchical planning pipeline, achieving significant improvement of exploration rate compared with recent methods\cite{cieslewski2017rapid, bircher2016receding}. However, it does not consider global coverage routes, so the quadrotor may revisit the same regions repeatedly, leading to decreased performance. To further improve efficiency, we exploit the global CPs to guide the exploration planning, so that the quadrotor visits different regions in a more sensible order. 

\subsubsection{CP-guided exploration path planning}
\label{subsubsec:CP-guided}

In Sect.\ref{sec:cvrp}, the CVRP outputs the CP of the hgrid cells assigned to each quadrotor. When the hgrid is updated as the map changes, we recompute the CP to guide the exploration planning. For the next $ N_{\tn{CP}} $ hgrid cells along the CP, we retrieve the frontier clusters whose centroids lie inside them. Then we find a path that starts at the quadrotor's current viewpoint, visits each of the clusters, and ends at the $ (N_{\tn{CP}}+1) $-th hgrid cell's centroid, as shown in Fig.\ref{fig:CP-guided}. Inspired by \cite{meng2017two}, the problem is formulated as a variant of TSP with fixed start and end points. Since TSP is a special case of VRP (the number of vehicle is 1), a procedure similar to that in Sect.\ref{subsec:cvrp} can be adopted to solve the problem, except that an extra end-point constraint introduced by the $ (N_{\tn{CP}}+1) $-th hgrid cell should be considered.

Assume there are $ N_{\tn{ftr}} $ clusters totally, the engaged TSP has $ N_{\tn{ftr}}+3 $ nodes, in which $ N_{\tn{ftr}} $ nodes are for the clusters and 3 ones are for the virtual depot, quadrotor and hgrid cell respectively. The cost matrix $ \mathbf{C}_{\text{tsp}} \in \mathbb{R}^{(N_{\tn{ftr}} +3)\times(N_{\tn{ftr}} +3)} $ is:
\begin{align}
\label{equ:mat_tsp}
  \mathbf{C}_{\text{tsp}} = \left[ \begin{array}{cccc}
    0 & -M_{\text{inf}} & \mathbf{0} & 0 \\
    0  & 0 & \mathbf{C}_{q,f} & 0 \\
    \mathbf{0} & \mathbf{C}_{q,f}^{\tn{T}} & \mathbf{C}_{f} & \mathbf{C}_{h,f}^{\text{T}} \\ 
    -M_{\text{inf}} & 0 & \mathbf{C}_{h,f} & 0
  \end{array} \right]
\end{align}
$ \mathbf{C}_{f} $ is the major symmetric block recording the connection costs between clusters, whose entries are computed by:
\begin{align}
\label{equ:cluster_cluster}
  \mathbf{C}_{f}(k_1,k_2) = t_{\text{lb}}(\mathbf{q}_{k_1,1}, \mathbf{q}_{k_2,1}), \ k_1, k_2 \in \left[1, N_{\text{ftr}} \right] 
\end{align}
Different from the costs of hgrid cells (Equ.\ref{equ:cell_cell}), Equ.\ref{equ:cluster_cluster} takes into account not only translational distance, but also the change of yaw angle. As these costs are already precomputed when new frontier clusters are extracted (Sect.\ref{subsec:frontier}), $ \mathbf{C}_{f} $ can be filled without extra overhead. 

$ \mathbf{C}_{q,f} \in \mathbb{R}^{1 \times N_{\tn{ftr}}} $ is the costs from the current viewpoint $\mathbf{q}_0 = \left( \mathbf{p}_0, \varphi_0 \right)$ to the $N_{\text{ftr}}$ clusters:
\begin{equation}
  \label{equ:drone_to_frontier}
  \mathbf{C}_{q,f}(k) = t_{\text{lb}}(\mathbf{q}_0,\mathbf{q}_{k,1}) + w_{\text{con}} \cdot t_{\text{con}}(\mathbf{q}_{k,1}), \ k \in \left[1, N_{\text{cls}} \right]
\end{equation}
{
\begin{equation}
  \label{equ:drone_to_frontier2}
  t_{\text{con}}(\mathbf{q}_{k,j}) = \left\{ \begin{array}{ll}
    \cos^{-1} \frac{(\mathbf{p}_{k,j}-\mathbf{p}_0)\cdot \mathbf{v}_0}{\left\| \mathbf{p}_{k,j}-\mathbf{p}_0 \right\| \left\|\mathbf{v}_0 \right\|} & , \mathbf{v}_0 \neq \mathbf{0}\\
  0 & , \text{else}\end{array} \right. 
\end{equation}
}
where $\mathbf{v}_0$ is the current velocity, while $ t_{\text{con}}(\cdot) $ penalizes large changes in flight direction, which is introduced to enable more consistent movements, similar to Equ.\ref{equ:consistency}.

The costs from the clusters to the hgrid cell are accounted for by $ \mathbf{C}_{h,f} \in \mathbb{R}^{1 \times N_{\text{ftr}}} $, which is evaluated by:
\begin{equation}
  \mathbf{C}_{h,f}(k) = \tnb{Len}[P(\mathbf{p}_{k,1}, \mathbf{c}_{N_{\tn{CP}}+1})] / v_{\tn{max}}
\end{equation}

We transform our TSP variant to a standard one by introducing a huge negative cost $ -M_{\text{inf}} $, which is assigned between the virtual depot and quadrotor, as well as between the hgrid cell and virtual depot. It ensures that in the output route, the nodes of the quadrotor and hgrid cell are adjacent to the depot's. As a result, we can obtain the desired path by removing the depot node and the two edges connected with it (Fig.\ref{fig:CP-guided}).

\subsubsection{Local Viewpoint Refinement}
\label{subsubses:planning_local}

\begin{figure}[t!]
  \centering
  \subfigure{\includegraphics[width=0.7\columnwidth]{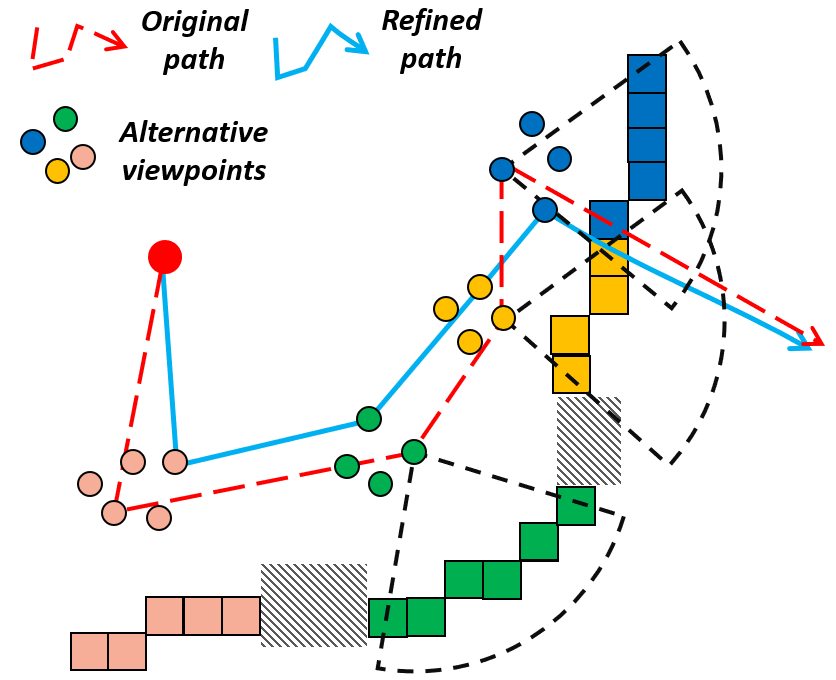}}       
  \caption{\label{fig:local_viewpoint} Local viewpoint refinement based on the graph search approach. Along the local exploration path, alternative viewpoints of each frontier cluster are considered, further improving the path quality.
  }
  \vspace{-0.5cm}
\end{figure}

{A promising initial path is found using the CP-guided path planning. However, the path only considers a single viewpoint for each cluster, despite the fact that multiple possible viewpoints exist. To further improve the path's quality, the graph-based viewpoint refinement method in \cite{zhou2021fuel} is used. It generates a directed acyclic graph by taking into account the quadrotor's current viewpoint, the viewpoints of each cluster, and the hgrid cell on the initial path. It captures all possible viewpoint combinations along the initial path. Note that to incorporate the guidance of the CP, extra graph nodes and edges are introduced for the next hgrid cell to be explored, which differs from the original method. Given this graph, the Dijkstra algorithm is employed to find the optimal path that minimizes the exploration cost. The process is depicted in Fig.\ref{fig:local_viewpoint}.}

%

\subsubsection{Minimum-time B-spline Trajectory Generation}
\label{subsubsec:planning_traj}

Given the path comprising discrete viewpoints, a continuous-time trajectory executable by the quadrotor is needed. {We base our trajectory generation on \cite{zhou2019robust,zhou2021fuel}, which generates smooth, safe and dynamically feasible B-spline trajectories in real-time. We further consider collision avoidance among the quadrotors.}

The differential flatness of quadrotor\cite{MelKum1105} allows us to generate trajectories simply for the flat output $ \mathbf{q} \in \left( x,y,z,\varphi \right) $. For each quadrotor $q_i$, we find the uniform B-spline trajectory $ \mathbf{\Psi}_{\tn{b},i}(t) = \left( \mathbf{p}_{\tn{b}}(t), \varphi_{\tn{b}}(t) \right)  $ that minimizes smoothness cost and total trajectory time under the constraints of safety, dynamic feasibility and boundary state. The B-spline has a degree $ p_{\tn{b}} $ and is defined by a set of $ N_{\tn{b}}+1 $ control points $ \mathcal{Q}_{\text{c,b}} = \left\{ \mathbf{q}_{\text{c},0}, \mathbf{q}_{\text{c},1}, \cdots, \mathbf{q}_{\text{c},N_{\tn{b}}} \right\} $ where $ \mathbf{q}_{\text{c},m} = \left( \mathbf{p}_{\text{c},m}, \varphi_{\text{c},m} \right) $, and a knot span $ \Delta t_{\tn{b}} $. An optimization problem is formulated to find the desired solution:
\begin{equation}
   \underset{ \mathcal{Q}_{\text{c,b}}, \Delta t_{\tn{b}} }{\arg \min } \ J_{\tn{s}}+ w_{\tn{t}} T + \lambda_{\tn{c}} \left( J_{\tn{c,o}} + J_{\tn{c,q}} \right)  + \lambda_{\tn{d}}\left( J_{\tn{v}} + J_{\tn{a}} \right) + \lambda_{\tn{bs}} J_{\tn{bs}} \nonumber
\end{equation}

{Here, $ J_{\tn{s}} $ is the elastic band smoothness cost, $ T $ is the total trajectory time, $ J_{\tn{c,o}}, J_{\tn{c,q}} $ are the penalties to avoid collisions with obstacles and other quadrotors respectively. $ J_v, J_a $ are the constraints of dynamics feasibility, while $ J_{\tn{bs}} $ is the boundary state constraints considering the instantaneous state $ ( \mathbf{q}_0, \dot{\mathbf{q}}_0, \ddot{\mathbf{q}}_0) $ and the target viewpoint obtained in Sect.\ref{subsubses:planning_local}. The method has been shown to find high-quality trajectories in real-time and can support high-speed navigation in complex environments for a single quadrotor. Detailed formulation can be found in \cite{zhou2021fuel} and only the multi-robot collision avoidance term $ J_{\tn{c,q}} $ is specified here.} 

For a quadrotor to avoid collisions with nearby ones $ \mathcal{Q}_{\tn{n}} $, their trajectories $ \mathbf{\Psi}_{\tn{b},j}(t), q_j \in \mathcal{Q}_{\tn{n}} $ are involved. The distance between $q_i$ and every $q_j$ is enforced, similar to \cite{zhou2020ego2}:
\begin{align}
  & J_{\tn{c,q}} = \sum_{q_j \in \mathcal{Q}_{\tn{n}} } \sum_{k = 1}^{ T/\delta t_{\tn{q}}} \mathcal{J}(d_{\tn{q},j}(T_k), d_{\tn{min,q}})  \\
  & d_{\tn{q},j}(t) = \left\lVert \mathbf{E}^{\frac{1}{2}} \left[ \mathbf{\Psi}_{\tn{b},i}(t) - \mathbf{\Psi}_{\tn{b},j}(t) \right]  \right\rVert
\end{align} 
here $ T_k = T_s + k \delta t_{\tn{q}} $ where $ T_s $ is the start time of $ \mathbf{\Psi}_{\tn{b},i}(t) $, $ \mathbf{E} = \tn{diag}(1,1,\beta_{\tn{q}}), \beta_{\tn{q}} < 1 $ transforms the Euclidean distance to account for the downwash effect of quadrotors. {The trajectory generation is also performed in a decentralized manner and it occurs at a higher frequency than in \cite{zhou2021fuel}. In addition to planning a new trajectory after a new exploration path is computed, each quadrotor shares its latest trajectory periodically via the broadcast network. If a quadrotor receives a trajectory from others and a collision is detected, it immediately generates a new trajectory that avoids the collision while still reaching the same target viewpoint. The constant exchange of trajectories and replanning ensures safety under communication latency or packet losses.  }


\vspace{0.2cm}

\section{Implementation Details}
\label{sec:implement}

\subsection{Mapping and Information Sharing}

To achieve fast exploration, an efficient mapping framework is essential. 
In this work, we utilize a volumetric mapping \cite{han2019fiesta}, which has shown promising performance in fast autonomous flights\cite{zhou2019robust, zhou2020raptor} and exploration in complex scenes. 
Similar to \cite{wurm2010octomap}, which is widely adopted in exploration, it builds an occupancy grid map by fusing sensor measurements like depth images.  
It allows efficient and probabilistic updates of occupied and free space, and models the unknown space for the exploration planning.
Meanwhile, it also maintains an ESDF using an incremental algorithm to facilitate the gradient-based trajectory planning (Sect.\ref{subsubsec:planning_traj}). 
More details about the mapping framework can be found in \cite{han2019fiesta}.

For a more informed exploration planning, it is important to exchange map data.
Unlike some multi-robot mapping approaches which share map information as submaps at a low frequency\cite{yusmmr,dubois2020dense}, we group the newly observed voxels into \textit{chunk}s and share them immediately with nearby quadrotors when unknown areas are explored.
To update the latest map information promptly, the chunks are {broadcast} by adopting a UDP protocol to alleviate the costly handshakes\cite{schwartz1986telecommunication}.
Meanwhile, to deal with packet drop and limited communication range, each quadrotor stores a bookkeeping recording the held chunks, which are either produced by itself or received from others. 
The bookkeeping is {broadcast} at interval, while every quadrotor receiving the bookkeeping finds the unrecorded chunks and shares the ones it holds.
This sharing scheme allows retrieving lost information and relaying messages from quadrotor to quadrotor, which makes the system less sensitive to unreliable and range-limited communication.

\subsection{Decentralized Multi-robot State Estimation}

In a decentralized exploration system, each quadrotor should be able to localize itself and others independently. In real-world experiments, we employ Omni-Swarm \cite{xu2021omni}, a decentralized state estimator that fuses measurements from camera, IMU, and landmarks and keyframes shared among quadrotors. By fusing the image and IMU data with a tightly-coupled optimization-based method, it estimates the ego-motion of a quadrotor at high-frequency. For relative pose estimation, a map-based module is introduced, which relies on the landmarks \& keyframes shared among quadrotors and a loop closure detection procedure. Based on it, relative localization and re-localization can be performed by identifying common locations visited by different quadrotors, which also enables fast initialization of the relative state estimation. A graph-based optimization and forward propagation backend fuses the above-mentioned measurements, generating accurate and globally consistent estimation in real-time. {More details about the performance of this estimator, as well as its computational overhead and requirement of bandwidth can be found in \cite{xu2021omni}}

{The relative state estimation allows the quadrotors to establish a common coordinate frame. Note that the state estimator should be initialized at the start of the mission. At this point, we position all quadrotors in areas with common visual features so that the map-based module can detect loops and determine relative poses. The estimator then works even in non-line-of-sight conditions, such as when the quadrotors are completely separated by walls, which has been verified experimentally in \cite{xu2021omni}. By incorporating bearing measurements\cite{nguyen2020vision} among the robots, it is possible to improve the initialization stage. However, this is an estimation problem that is beyond the scope of this paper.}

\subsection{System Setup}

In real-world experiments, customized quadrotor platforms are used. The sensor used by each quadrotor for dense mapping is an Intel RealSense depth camera D435, which has a FoV of $ [87 \times 58] \ \tn{deg} $. {Each quadrotor has a DJI manifold 2G onboard computer which contains a NVIDIA Jetson TX2 module\footnote{https://www.dji.com}. Each quadrotor weighs 1.32 kg in total. The propulsion system uses approximately 246 W of power. The onboard computer consumes up to 15 W, the depth camera consumes 3.5 W, and an UWB module consumes 1.3 W. With all hardware and software modules operational, the quadrotor can hover or fly at a low speed for 15 minutes.}

The CVRP and TSP problems are solved by the LKH3 package\footnote{http://akira.ruc.dk/~keld/research/LKH-3/}, while the trajectory optimization is implemented based on Nlopt\footnote{https://nlopt.readthedocs.io/en/latest/}. A geometric controller\cite{lee2010} is adopted for trajectory tracking control. All the state estimation, mapping, planning, and control algorithms run on the onboard computer.  The quadrotors exchange information through the wireless ad hoc network. The exchange of messages is implemented by the tools in LCM\footnote{https://lcm-proj.github.io/}. For simulation, we use a customized simulating package containing the quadrotor dynamics model, map generator and depth camera model. All simulations run on an Intel Core i7-8700K CPU and GeForce GTX 1080 Ti GPU.


\section{Results}
\label{sec:results}

The proposed approach is evaluated extensively through simulation and real-world experiments.
To justify the design of each component in our algorithm, we conduct ablation studies comparing the complete method against baseline variants (Sect.\ref{subsec:ablation}).
We compare different strategies to coordinate the multi-robot exploration in Sect.\ref{subsec:res-multi-robot}.
In Sect.\ref{subsec:number}, we study how the number of quadrotors influences the exploration. 
Finally, the real-world experiments including indoor and outdoor ones are presented in Sect.\ref{subsec:res-real-world}. 

\subsection{Ablation Studies}
\label{subsec:ablation}

\begin{figure}[t!]
  \centering
  \subfigure{
    \includegraphics[width=0.45\columnwidth]{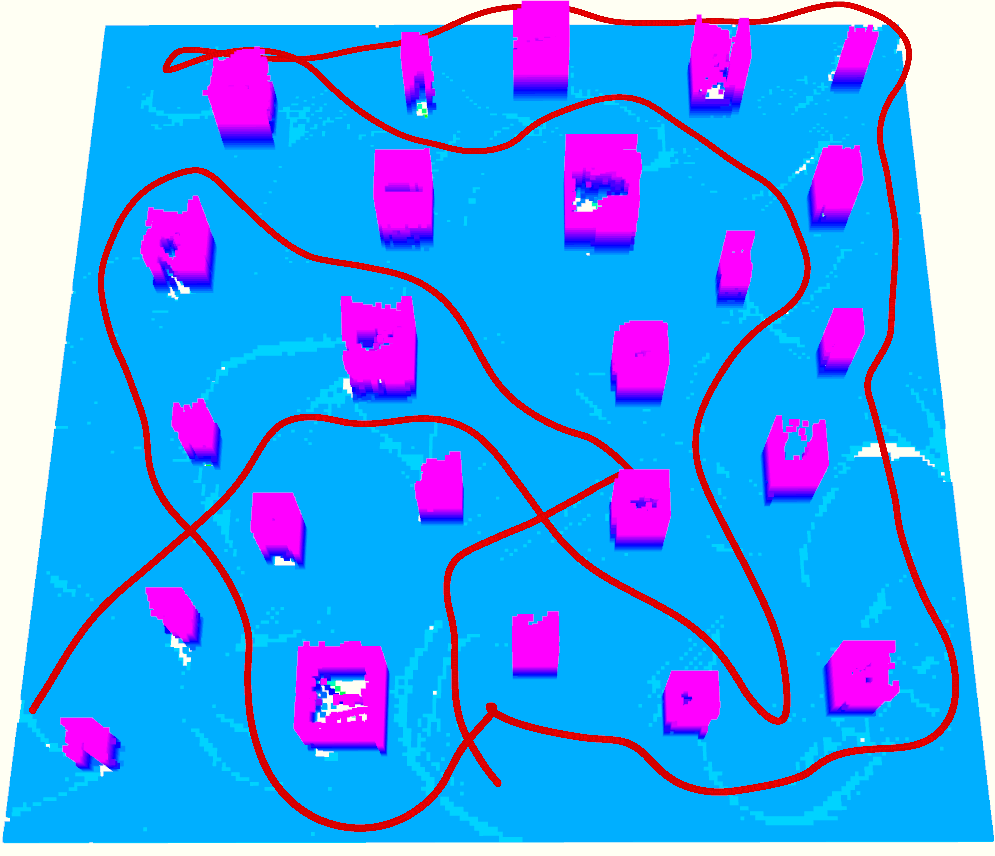}
    \includegraphics[width=0.44\columnwidth]{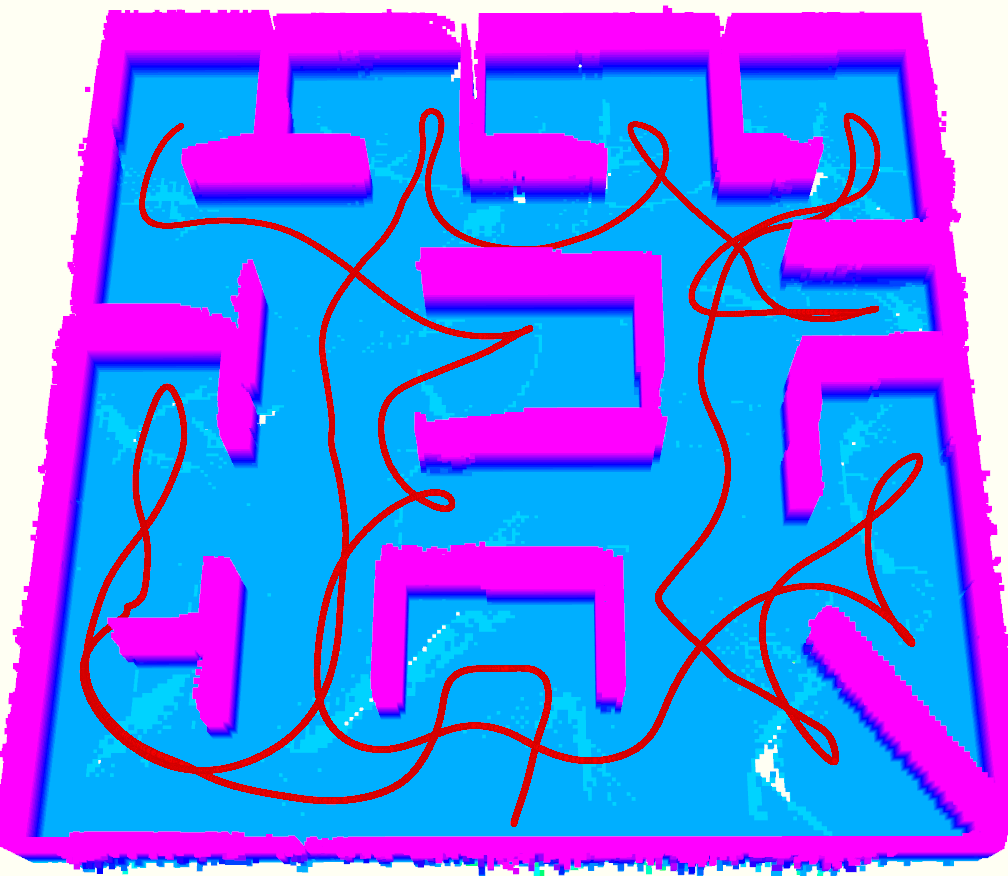}
  }       
  \subfigure{
    \includegraphics[width=0.45\columnwidth]{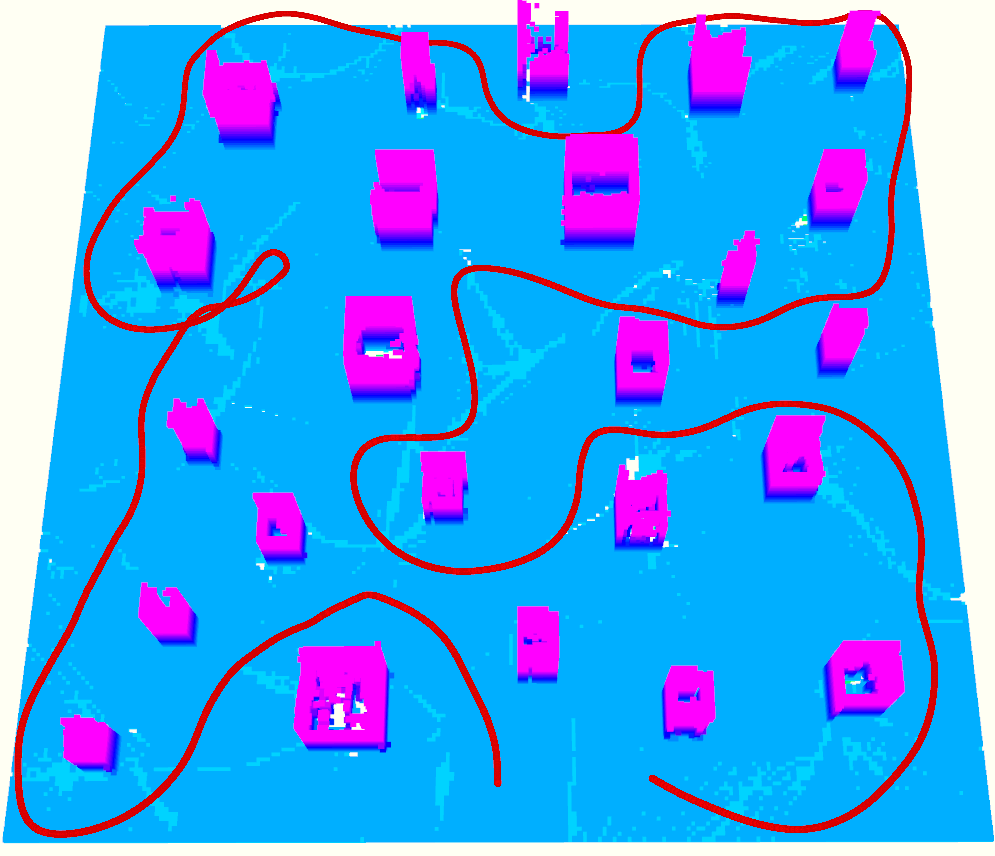}
    \includegraphics[width=0.44\columnwidth]{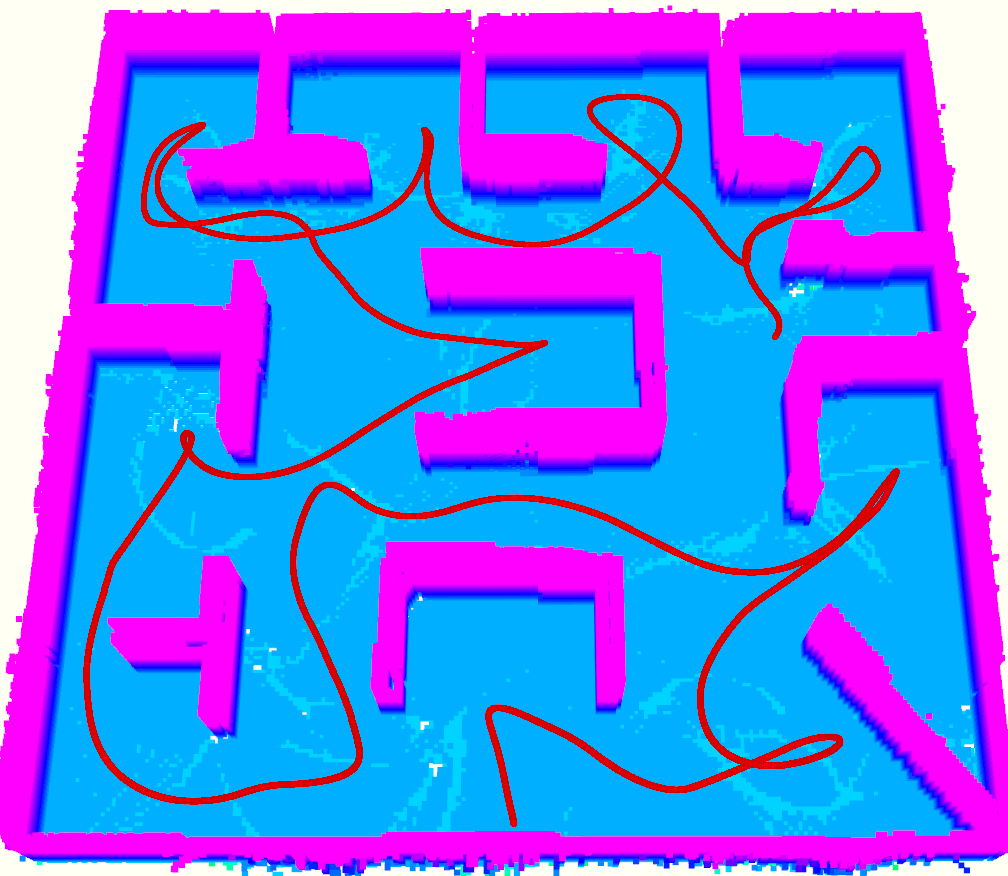}
  }       
  \caption{\label{fig:res-path} Ablation study of the CP-guided path planning.
  Top: without considering the CP, the quadrotor visits the same regions redundantly.
  Bottom: different regions are visited in a more sensible sequence. 
  }
  \vspace{-0.8cm}
\end{figure}

\subsubsection{Coordinating Multiple Robots}
\label{subsubsec:ablation-coord}

To validate the effectiveness of the hgrid-based pairwise interaction and the CVRP-based task allocation, we compare our complete coordination approach \textit{Full} to {three} variants. The first variant \textit{NoHgrid} does pairwise interaction (Sect.\ref{subsec:pairwise}) to allocate frontiers among quadrotors. The key difference is that it does not decompose the entire space. Instead, it just regards the frontier clusters as elementary task units, as most approaches do, and partitions the clusters between pairs of quadrotors by a mTSP formulation. {The second and third variants \textit{H+BFS} and \textit{H+mTSP} decompose the unknown space into hgrid cells, while different strategies are employed to allocate the cells. \textit{H+BFS} clusters the cells in a breath-first-search (BFS) manner\cite{karapetyan2017efficient}, whereas \textit{H+mTSP} uses a mTSP formulation to reduce the total length of coverage paths.} Except for the difference in coordination, the same path planning approach presented in \cite{zhou2021fuel} and trajectory generation (Sect.\ref{subsubsec:planning_traj}) are adopted for a fair comparison.

In the tests, the dynamic limits are set as $v_{\tn{max}} = 1.5 $ m/s and $ \dot{\varphi}_{\tn{max}} =0.9 $ rad/s.   The FoVs of the sensors are set as $ [80 \times 60] $ deg with a maximum range of $4.5$ m. The tests are conducted in two scenes, where one contains randomly generated pillars and the other is an office-like environment, as the two scenes shown in Fig.\ref{fig:benchmark}. Four quadrotors are initially placed near the boundaries of the explored space. Each approach runs 5 times in each scene.

As shown in Tab.\ref{tab:ablation1}, H+mTSP outperforms NoHgrid in both scenes, demonstrating that using subdivided regions as task units rather than frontier clusters is more effective. Although frontiers provide hints on how to navigate the space, they do not contain information on workloads. Specifically, a large unexplored area may lie behind a small frontier cluster, and vice versa, thus coordinating robots by frontier clusters usually leads to unbalanced partitioning of workloads. In contrast, the areas/volumes of subdivided unexplored areas directly indicate the amount of work, making it a more reasonable choice. Besides, since the hgrid cells represent disjoint regions, it prevents multiple quadrotors from visiting identical places and interfering each other. {On the other hand, H+mTSP has a higher exploration efficiency than H+BFS. H+BFS provides a simple heuristic to divide the cells quickly, however, it does not explicitly optimize the length of coverage path, which limits its performance in complex scenes.} Lastly, it is clear from the statistics that the CVRP-based partitioning further improves efficiency by a large margin. Comparing to the {BFS and} mTSP allocation, the CVRP accounts for not only the lengths of CPs, but also the actual amount of unexplored regions. Hence, workloads are more appropriately distributed.

\begin{table}[t]
  \centering
  \caption{\label{tab:ablation1} Ablation study of coordination approach.}
  \begin{tabular}{p{0.5cm}P{0.7cm}p{0.3cm}p{0.3cm}p{0.3cm}p{0.3cm}p{0.3cm}p{0.3cm}p{0.3cm}p{0.3cm}} 
  \hline\hline
  \textbf{Scene}                 &\textbf{Method}           & \multicolumn{4}{c}{\textbf{Exploration time (s)} }            & \multicolumn{4}{l}{\textbf{Path length (m)}}               \\ 
  \cline{3-10}               
  \multirow{4}{*}{Pillar}       &               & \textbf{Avg} & \textbf{Std} & \textbf{Max} & \textbf{Min} & \textbf{Avg} & \textbf{Std} & \textbf{Max} & \textbf{Min}  \\
                                & NoHgrid       & 48.6  &  \tnb{0.80}  &  49.7  &  47.6  &  204.9  &  8.15  &  212.1 &  193.5          \\
                                & {H+BFS}         & {46.9}  &  {1.31}  &  {48.3}  &  {45.1}  &  {201.3}  &  {8.97}  &  {209.7}  &  {192.6}               \\                 
                                & {H+mTSP}        & 44.0  &  0.85  &  45.2  &  43.2  &  194.6  &  9.05  &  204.1 &  182.4               \\
                                & Full          & \tnb{38.9} &  1.01  &  \tnb{40.6}  &  \tnb{37.6}  &  \tnb{171.8}  &  \tnb{3.21}  &  \tnb{177.6} &  \tnb{169.3}             \\
 \hline                      
 \multirow{3}{*}{Office}        & NoHgrid       &  47.5 &  1.12  &  49.2  &  46.2  &  210.1 &  \tnb{9.01}  &  219.9 &  197.5          \\
                                & {H+BFS}         &  {45.7} &  {1.28}  &  {47.9}  &  {44.9}  &  {205.4} &  {10.13} &  {212.9} &  {198.5}                 \\
                                & {H+mTSP}        &  41.7 &  \tnb{1.08}  &  43.2  &  40.6  &  197.5 &  18.13 &  221.9 &  178.5           \\
                                & Full          &  \tnb{35.4} &  2.06  &  \tnb{37.1}  &  \tnb{31.8}  &  \tnb{169.8} &  13.22 &  \tnb{179.4} &  \tnb{146.1}             \\
  \hline\hline
  \end{tabular}
\end{table}

\subsubsection{Exploration Path Planning}

\begin{table}[t]
  \centering
  \caption{\label{tab:ablation2} Ablation study of exploration path planning.}
  \begin{tabular}{p{0.5cm}P{0.7cm}p{0.3cm}p{0.3cm}p{0.3cm}p{0.3cm}p{0.3cm}p{0.3cm}p{0.3cm}p{0.3cm}} 
  \hline\hline
  \textbf{Scene}                 &\textbf{Method}           & \multicolumn{4}{c}{\textbf{Exploration time (s)} }            & \multicolumn{4}{l}{\textbf{Path length (m)}}               \\ 
  \cline{3-10}               
  \multirow{3}{*}{Pillar}       &               & \textbf{Avg} & \textbf{Std} & \textbf{Max} & \textbf{Min} & \textbf{Avg} & \textbf{Std} & \textbf{Max} & \textbf{Min}  \\
                                & NoCP          & 103.7 &  8.51 &  114.6 &  93.8 &  146.1 &  7.89 &  156.9 &  138.2          \\
                                & Full          &  \tnb{86.7} &  \tnb{4.33} &   \tnb{92.5} &  \tnb{82.1} &  \tnb{117.2} &  \tnb{3.20} &  \tnb{121.7} &  \tnb{114.6}             \\
 \hline                      
 \multirow{2}{*}{Office}        & NoCP          & 120.7 &  \tnb{1.87} &  123.1 &  118.6&  155.9 &  \tnb{1.46} &  157.9 &  154.4          \\
                                & Full          &  \tnb{96.9} &  2.56 &  \tnb{99.5}  &   \tnb{93.4} &  \tnb{125.1} &  4.47 &  \tnb{130.8} &  \tnb{119.9}             \\
  \hline\hline
  \end{tabular}
  \vspace{-0.6cm}
\end{table}

To examine the CP-guided exploration path planning, we compare our new approach (\textit{Full}) with our previous one\cite{zhou2021fuel} (\textit{NoCP}).
\cite{zhou2021fuel} is shown to substantially outperform recent exploration planning approaches including \cite{cieslewski2017rapid, bircher2016receding}, completing exploration 3-8 times faster. 
The improvement comes from the consideration of efficient frontier coverage tours, the promising viewpoint and trajectory optimization and its high computation efficiency. 
Despite its improvement, it does not consider the coverage route of the entire space, which sometimes leads to unnecessarily long paths.
In this work, we show that the efficiency of exploration can be further improved by incorporating the global CPs (Sect.\ref{subsubsec:CP-guided}). 
To focus on the evaluation of exploration path planning, we compare exploration with a single quadrotor, as is shown in Fig.\ref{fig:res-path}. 
The same parameters as those in Sect.\ref{subsubsec:ablation-coord} are set.

The advantage of introducing CPs is apparent from Fig.\ref{fig:res-path} and Tab.\ref{tab:ablation2}, which shows a more sensible exploration pattern, a significantly shorter exploration time and overall path length.  
Without taking the coverage route into account, it is observed that the quadrotor frequently moves to another region before thoroughly exploring one.
As a result, the quadrotor must revisit known regions later in order to explore previously missed areas, resulting in inefficiency. 
The incorporation of CPs, on the other hand, consistently provides a visitation sequence of the decomposed unexplored regions.
As a result, the quadrotor can explore the space in a more rational manner, rather than returning to the same locations repeatedly.


\subsection{Coordination of Multi-robot Exploration}
\label{subsec:res-multi-robot}

To further evaluate our coordination method, we benchmark it against four widely adopted approaches including centralized \cite{burgard2005coordinated, faigl2012goal} and decentralized \cite{smith2018distributed,klodt2015equitable} ones. 
\textit{Iter}\cite{burgard2005coordinated} uses a central controller that iteratively determines the appropriate target frontier cluster for each robot.
In each round, the best pair of robot and frontier cluster is computed, after which the utility of each frontier is updated according to the previous assignments. 
The process is repeated until each robot is assigned with a frontier cluster.
\textit{mTSP}\cite{faigl2012goal} is more sophisticated than \textit{Iter} because it considers the optimal allocation of all frontier clusters for each robot.
The allocation problem leads to an mTSP formulation.
Different from \textit{Iter} and \textit{mTSP}, \cite{smith2018distributed,klodt2015equitable} do not use a central controller, but have all robots to make decision independently.
\textit{Auc}\cite{smith2018distributed} exploits an auction-based architecture to achieve coordination among robots. 
In the framework, robots continuously negotiate with nearby ones, which allows each robot to explore their optimal target if there is no conflict, and resolves conflicting targets by comparing the expected travel costs and rewards.
Like \textit{mTSP}, \textit{Pair}\cite{klodt2015equitable} allocates all targets among robots, but it adopts a pairwise optimization in order to achieve decentralized allocation. 
The idea of pairwise interaction is similar to ours, but it entirely relies on frontier clusters rather than decomposed cells.
The same path planning, trajectory generation and parameters as those in Sect.\ref{subsubsec:ablation-coord} are used.

\begin{figure}[t!]
  \centering
  \subfigure[\label{fig:baseline1} Paths produced by approach \textit{Iter}\cite{burgard2005coordinated}.]
  {
    \includegraphics[width=0.50\columnwidth]{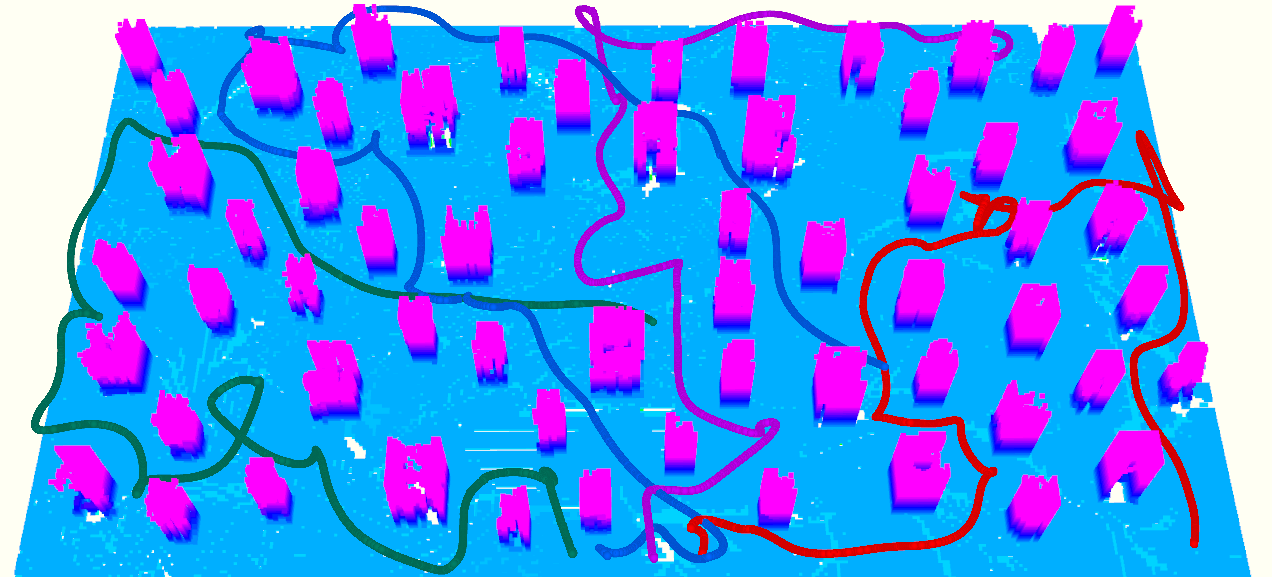}
    \includegraphics[width=0.49\columnwidth]{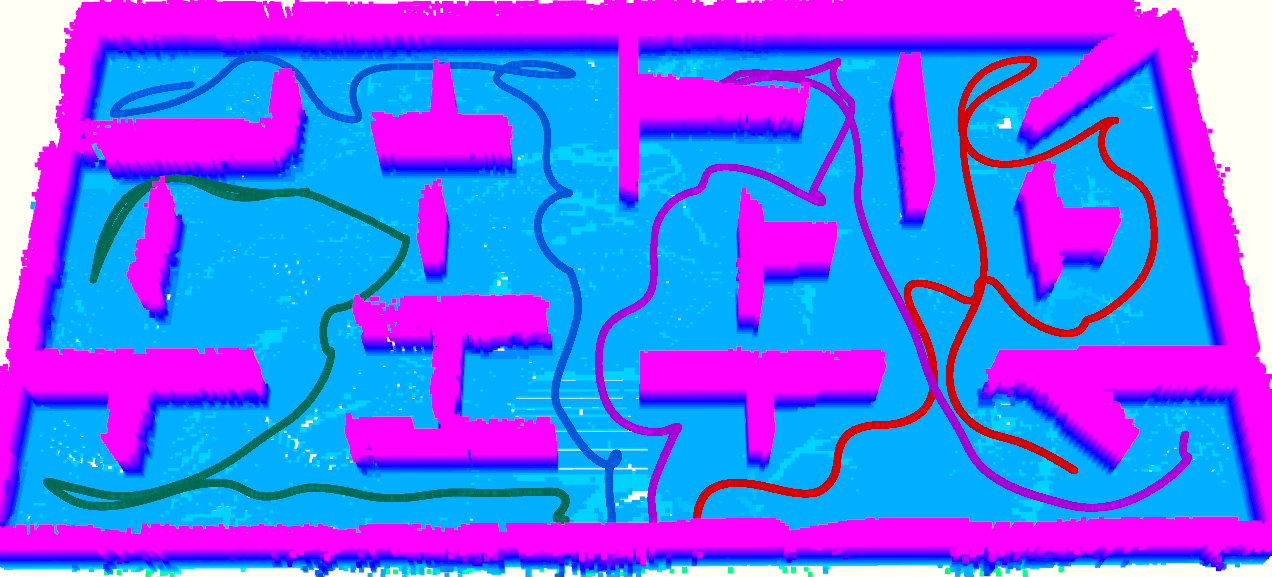}
  }
  \subfigure[\label{fig:baseline2} Paths produced by approach \textit{mTSP}\cite{faigl2012goal}.]{
    \includegraphics[width=0.50\columnwidth]{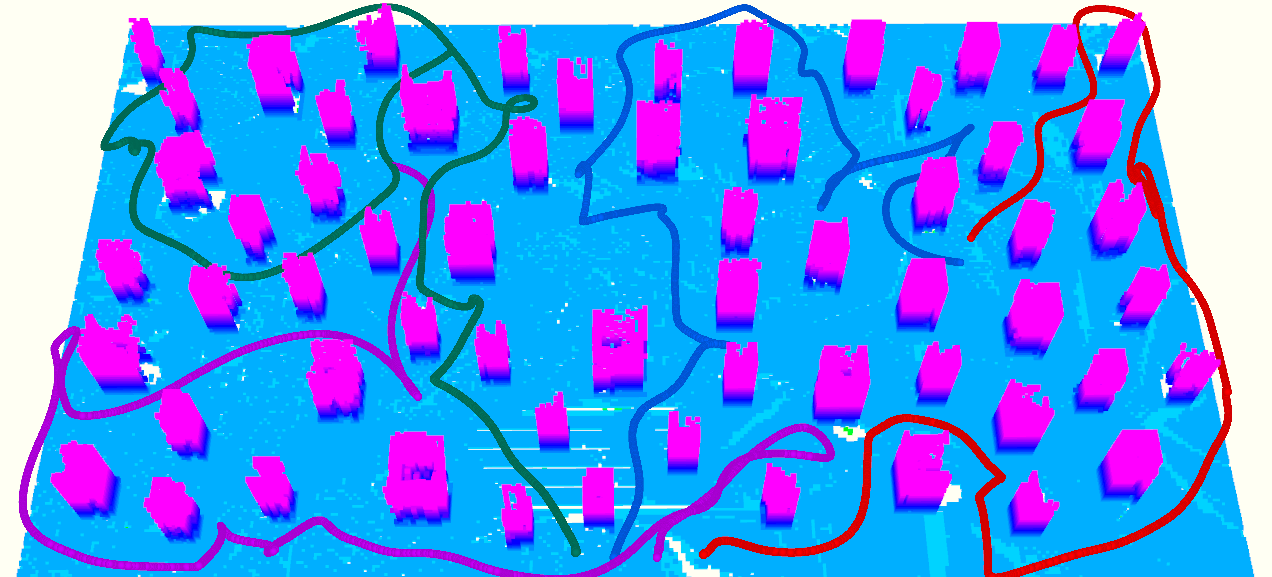}
    \includegraphics[width=0.49\columnwidth]{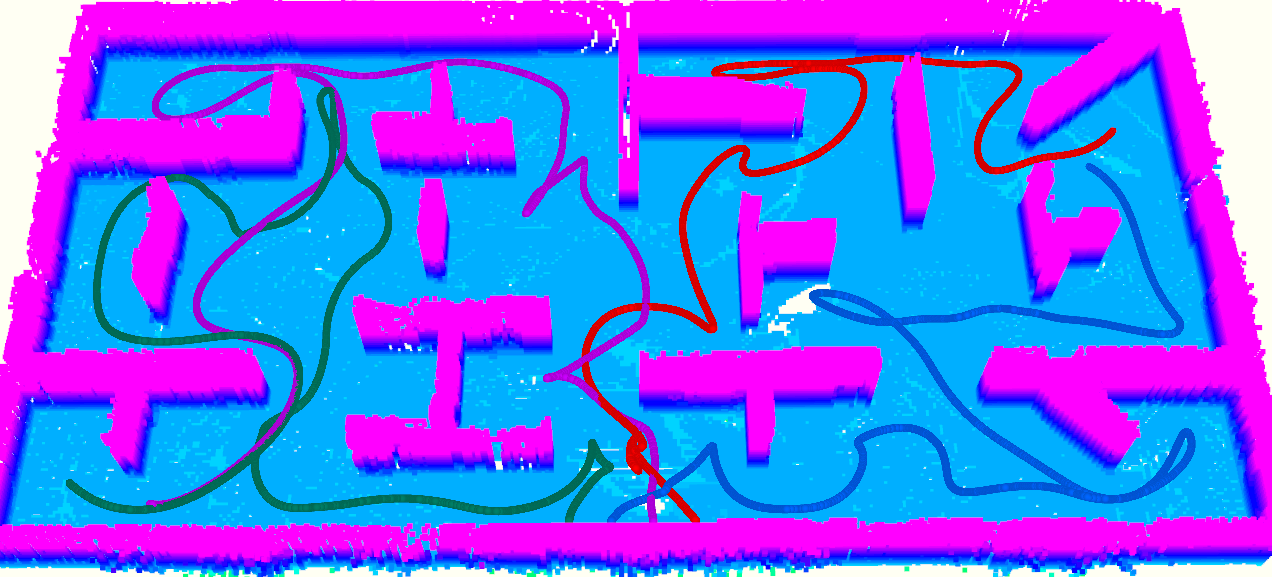}
    }
  \subfigure[\label{fig:baseline3} Paths produced by approach \textit{Pair}\cite{klodt2015equitable}.]{
    \includegraphics[width=0.50\columnwidth]{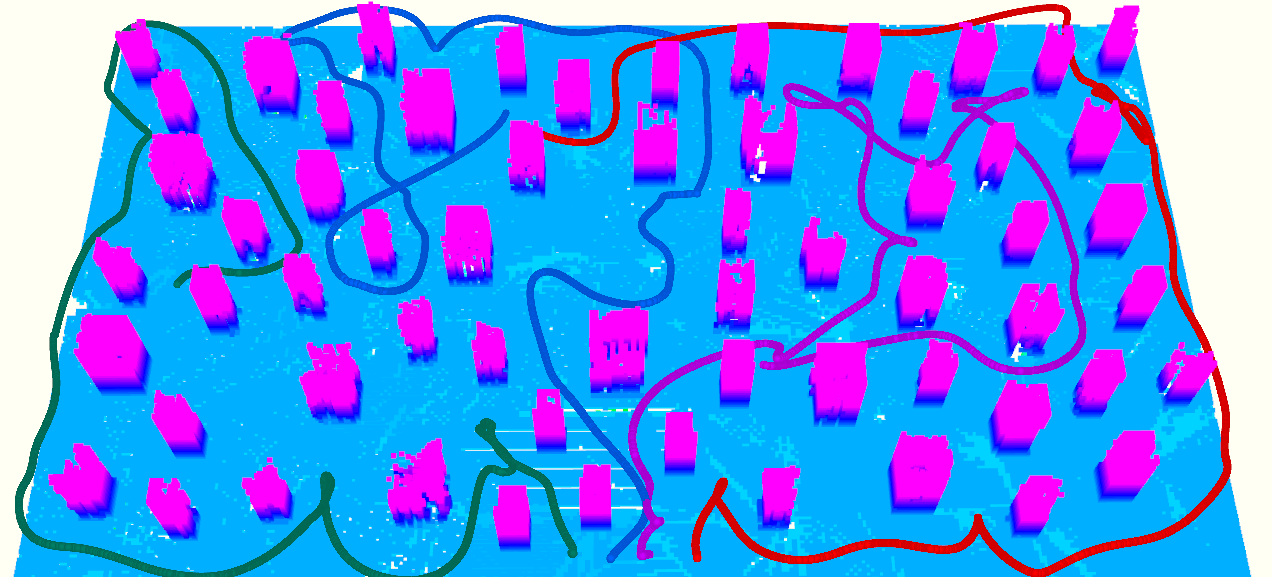}
    \includegraphics[width=0.49\columnwidth]{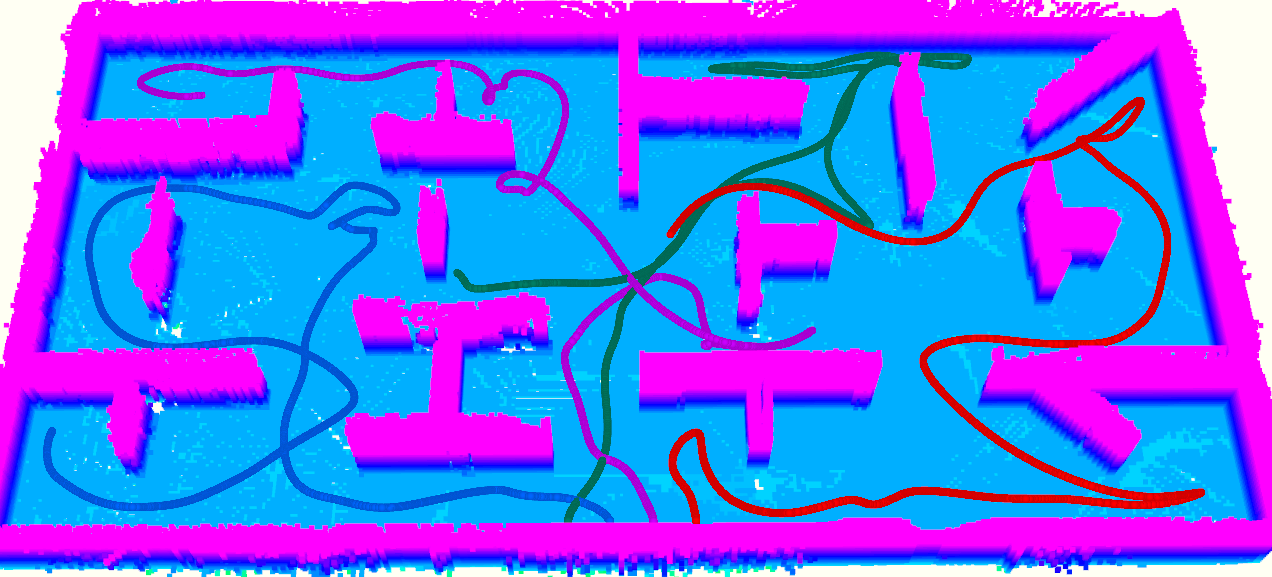}
  }
  \subfigure[\label{fig:baseline4} Paths produced by approach \textit{Auc}\cite{smith2018distributed}.]{
    \includegraphics[width=0.50\columnwidth]{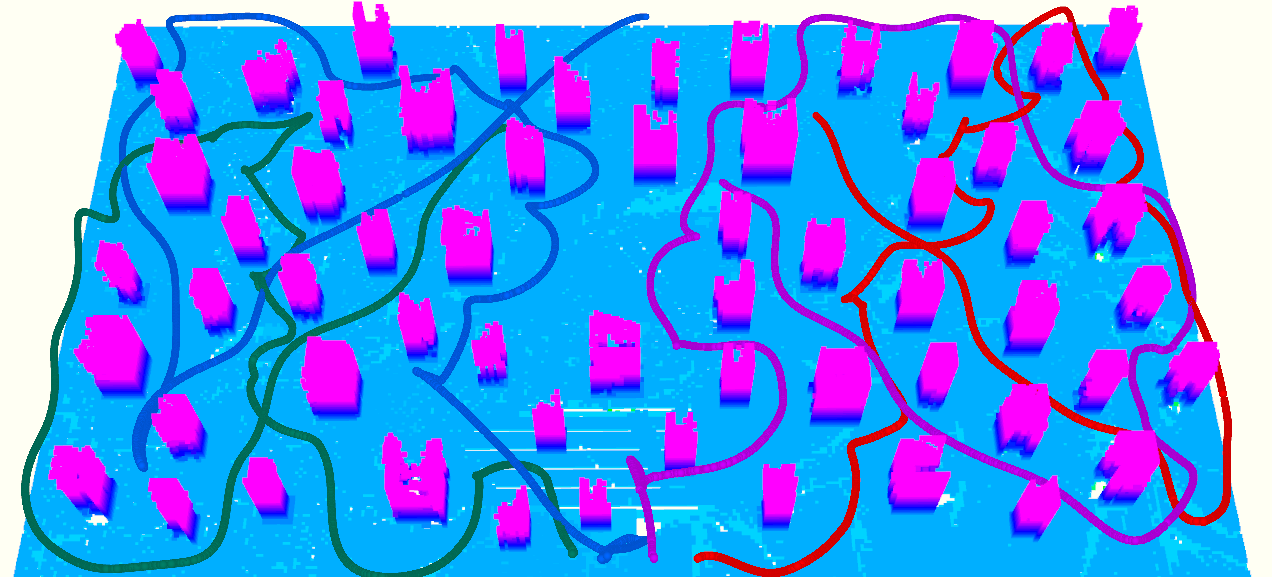}
    \includegraphics[width=0.49\columnwidth]{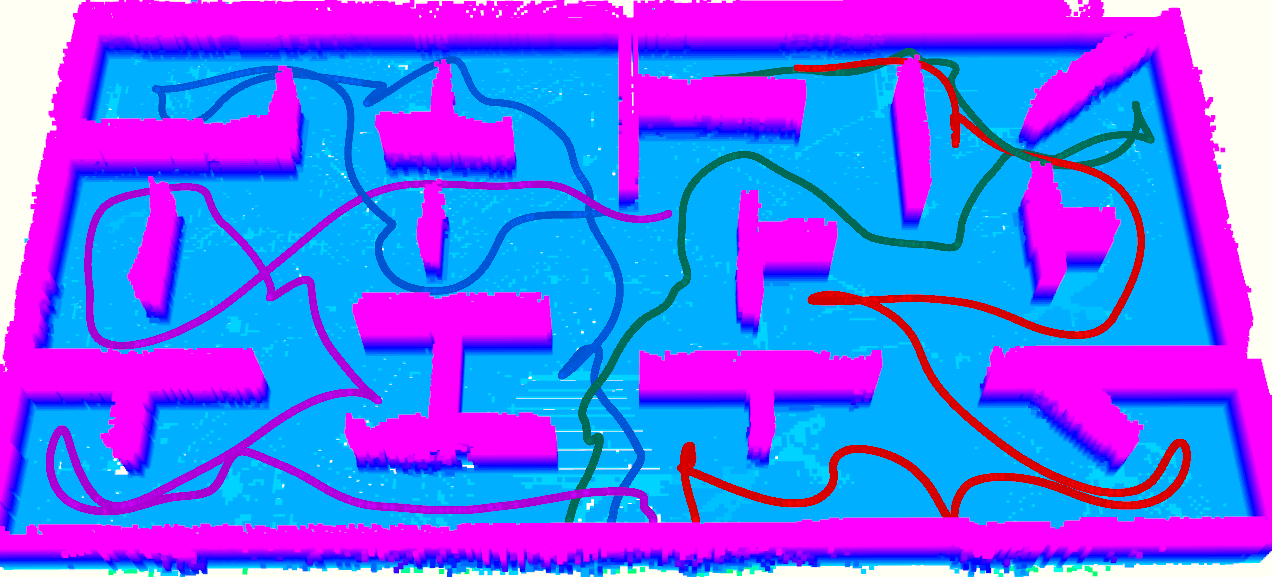}
  }
  \subfigure[\label{fig:benchmark_proposed} Path produced by our approach.]{
    \includegraphics[width=0.50\columnwidth]{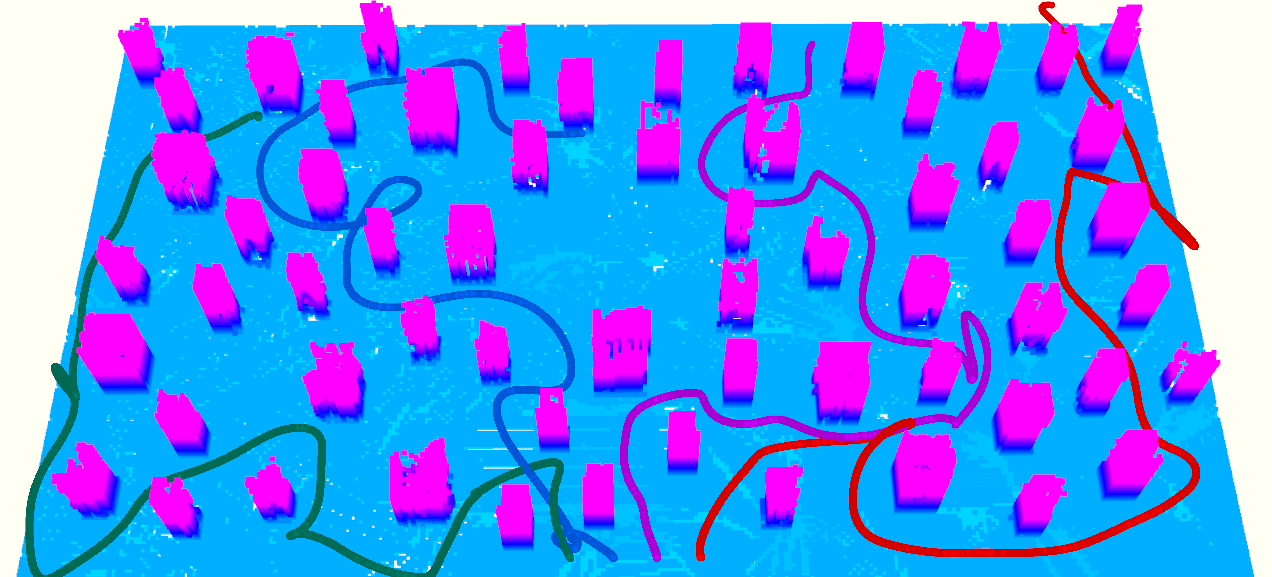}
    \includegraphics[width=0.49\columnwidth]{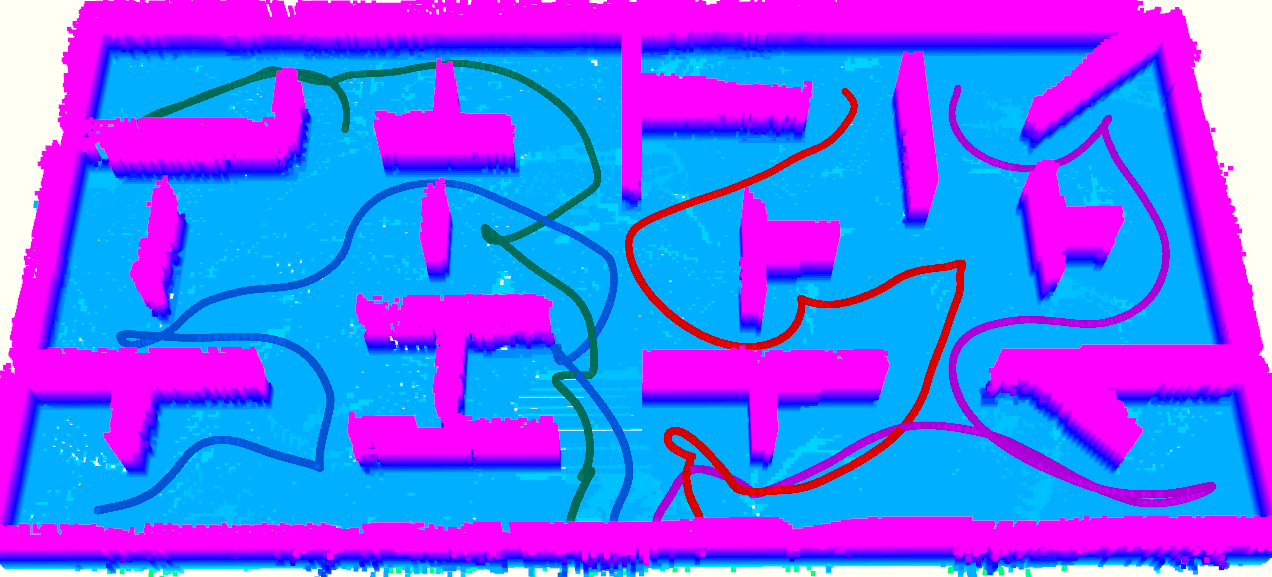}
  }
  \caption{\label{fig:benchmark} Comparisons of coordination approaches in unstructured (left) and structured (right) environments. Ours dispatch the quadrotors more effectively to explore distinct regions.
  }
  \vspace{-0.7cm}
\end{figure}

\begin{table}[t]
  \centering
  \caption{\label{tab:benchmark} Exploration statistic of benchmarked coordination approaches.}
  \begin{tabular}{p{0.5cm}P{0.5cm}P{0.7cm}p{0.3cm}p{0.3cm}p{0.3cm}p{0.3cm}p{0.3cm}p{0.3cm}p{0.3cm}p{0.3cm}} 
  \hline\hline
  \textbf{Scene}         & \textbf{CR(m)}    &\textbf{Method}           & \multicolumn{4}{c}{\textbf{Exploration time (s)} }            & \multicolumn{4}{l}{\textbf{Total path length (m)}}               \\ 
  \cline{4-11}
  \multirow{12}{*}{Pillar}     & \multirow{6}{*}{$+\infty $}  &         & \textbf{Avg} & \textbf{Std} & \textbf{Max} & \textbf{Min} & \textbf{Avg} & \textbf{Std} & \textbf{Max} & \textbf{Min}  \\
                               &                              & Iter    &  51.6 &  1.90 &  54.3 &  50.2 &  213.1 &  6.90 &  221.6 &  204.7         \\
                               &                              & mTSP    &  45.9 &  2.34 &  49.2 &  44.2 &  187.5 &  5.14 &  194.8 &  183.9               \\
                               &                              & Pair    &  48.2 &  \tnb{0.83} &  49.3 &  47.2 &  204.5 &  8.18 &  211.7 &  193.1           \\
                               &                              & Auc     &  51.6 &  1.72 &  53.3 &  49.2 &  212.9 &  7.35 &  222.4 &  204.4           \\
                               &                              & Ours    &  \tnb{38.5} &  1.24 &  \tnb{40.2} &  \tnb{37.2} &  \tnb{171.3} &  \tnb{1.14} &  \tnb{172.5} &  \tnb{169.8}             \\
  \cline{2-11}
                               & \multirow{3}{*}{10}          & Pair    &  51.9  &  1.70 &  54.3 &  50.3  &  209.2 &  15.78 &  229.0 &  \tnb{190.3}           \\
                               &                              & Auc     &  53.6  &  4.11 &  59.2 &  49.3  &  231.8 &  11.58 &  248.0 &  221.4    \\
                               &                              & Ours    &  \tnb{45.0}  &  \tnb{0.59} &  \tnb{45.6} &  \tnb{44.2}  &  \tnb{206.4} &   \tnb{8.92} &  \tnb{212.7} &  193.8           \\
  \cline{2-11}
                               & \multirow{3}{*}{5}           & Pair    &  60.2  &  6.70 &  71.4 &  53.7  &  271.0 &  42.13 &  343.2 &  238.1         \\
                               &                              & Auc     &  62.3  &  7.13 &  72.3 &  56.3  &  279.3 &  45.34 &  343.4 &  246.2           \\
                               &                              & Ours    &  \tnb{46.4}  &  \tnb{1.29} &  \tnb{47.4} &  \tnb{44.6}  &  \tnb{208.9} &  \tnb{14.66} &  \tnb{228.8} &  \tnb{194.0}             \\
  \hline
 \multirow{11}{*}{Office}      &  \multirow{5}{*}{$+\infty $} & Iter    & 49.4 &  0.48 &  50.0&  48.8 &  201.7 &   13.74&  215.9 &  183.1          \\
                               &                              & mTSP    & 45.0 &  2.30 &  48.3&  43.2 &  205.3  &   \tnb{8.18} &  216.7 &  197.8              \\
                               &                              & Pair    & 47.7 &  \tnb{0.42} &  48.2&  47.2 &  209.7  &   9.31&  219.7 &  197.3           \\
                               &                              & Auc     & 50.6 &  3.40 &  55.3&  47.2 &  229.3  &  13.90&  247.0 &  213.0           \\
                               &                              & Ours    & \tnb{35.8} &  2.14 &  \tnb{37.5}&  \tnb{32.2} &  \tnb{169.3}  &  13.63&  \tnb{178.6} &  \tnb{145.7}            \\
  \cline{2-11}
                               & \multirow{3}{*}{10}          & Pair    &  51.6 &  2.57 &  54.9 &  48.6 &  224.9 &   \tnb{8.55} &  235.1  &  214.2          \\
                               &                              & Auc     &  51.5 &  \tnb{1.69} &  53.1 &  49.1 &  245.1 &  11.82 &  260.4  &  231.6          \\
                               &                              & Ours    &  \tnb{40.2} &  3.06 &  \tnb{43.2} &  \tnb{36.0} &  \tnb{176.0} &  10.35 &  \tnb{185.2}  &  \tnb{161.6}            \\
  \cline{2-11}
                               & \multirow{3}{*}{5}           & Pair    &  55.9 &  1.86 &  58.5 &  54.1 &  253.2 &   \tnb{6.37} &  259.1  &  244.4     \\
                               &                              & Auc     &  57.7 &  \tnb{1.38} &  59.6 &  56.3 &  262.0 &   9.74 &  275.3  &  252.2         \\
                               &                              & Ours    &  \tnb{47.3} &  1.53 &  \tnb{48.6} &  \tnb{45.2} &  \tnb{202.6} &   8.05 &  \tnb{213.3}  &  \tnb{193.8}         \\
  \hline\hline
  \end{tabular}
  \vspace{-0.5cm}
\end{table}

First, we compare all five approaches under ideal communication, i.e., connections to the central server or among quadrotors are always available. 
Results are listed in Tab.\ref{tab:benchmark} and Fig.\ref{fig:benchmark}.
Among the baselines, approaches that consider the allocation of all targets jointly (\textit{mTSP}, \textit{Pair}) performs better than those allocating a single target at a time (\textit{Iter}, \textit{Auc}).
Comparing to \textit{Pair}, \textit{mTSP} is marginally more efficient.
However, \textit{Pair} has lower communication requirements as only two quadrotors have to communicate at a time, allowing coordination in communication-limited scenarios.
In terms of exploration time and total movement distance, the proposed method significantly outperforms all baselines.
From Fig.\ref{fig:baseline1}-\ref{fig:benchmark_proposed}, we can see that the paths produced by the baselines are longer and have more intersections than ours, indicating more waste of energy.
In comparison, the proposed coordination approach dispatches the quadrotors more reasonably to explore distinct regions, resulting in minor interference between the quadrotors.
It is remarkable that our approach outperforms the centralized coordination, despite being completely decentralized, owing to the more proper choice of task units and partitioning algorithm, as has been justified in Sect.\ref{subsubsec:ablation-coord}. 

To further assess the robustness of the approaches, different communication ranges are simulated.
The quadrotors cannot exchange information for coordination or share map data when they are out of communication range.
Since \cite{burgard2005coordinated, faigl2012goal} require communication at all time, we only compare against \cite{smith2018distributed,klodt2015equitable}. 
As the communication range decreases, the exploration time and path lengths of all approaches increase.
However, the time and lengths of \cite{smith2018distributed,klodt2015equitable} increase more particularly in the \textit{Pillar} scene.
Because of their limited communication range, quadrotors are less aware of which areas have already been explored by others.
As a result, multiple quadrotors may explore the same regions redundantly at different times, wasting a significant amount of time and energy.
This phenomenon is less severe in the \textit{Office} scene, as we find that in the structured environment, quadrotors have a better chance of meeting each other and exchanging more information.
In comparison to them, our approach has a more consistent performance in both scenes. 
The key point is that we subdivide the entire space into disjoint cells, which inherently ensures that a quadrotor does not visit regions assigned to others, even if communication is lost. 
By contrast, allocating frontier among quadrotors does not have this advantage, so its performance may suffer when communication is limited.

\begin{figure}[t!]
  \centering
  \subfigure[1 drone.]{\includegraphics[width=0.45\columnwidth]{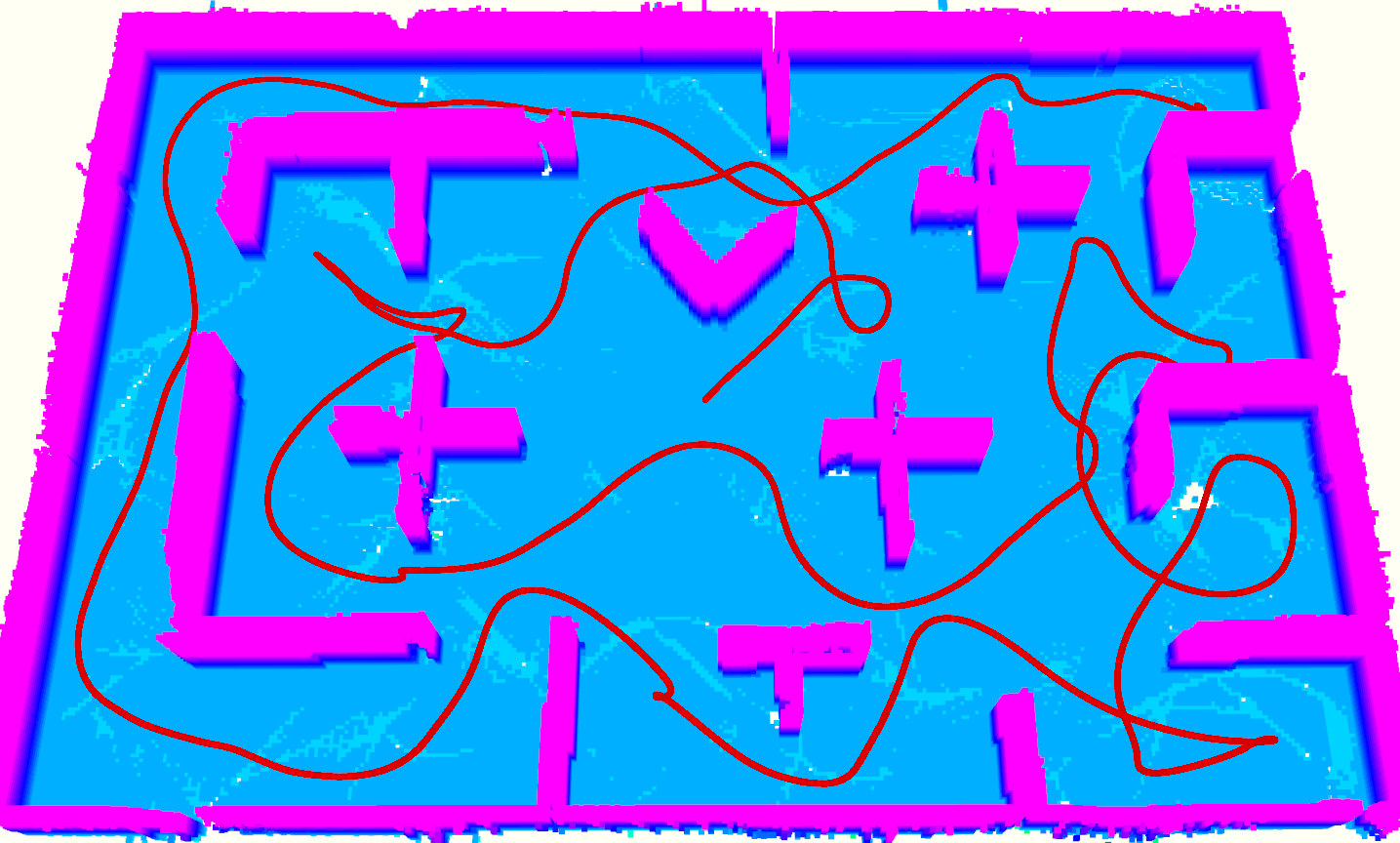}}       
  \subfigure[2 drones.]{\includegraphics[width=0.45\columnwidth]{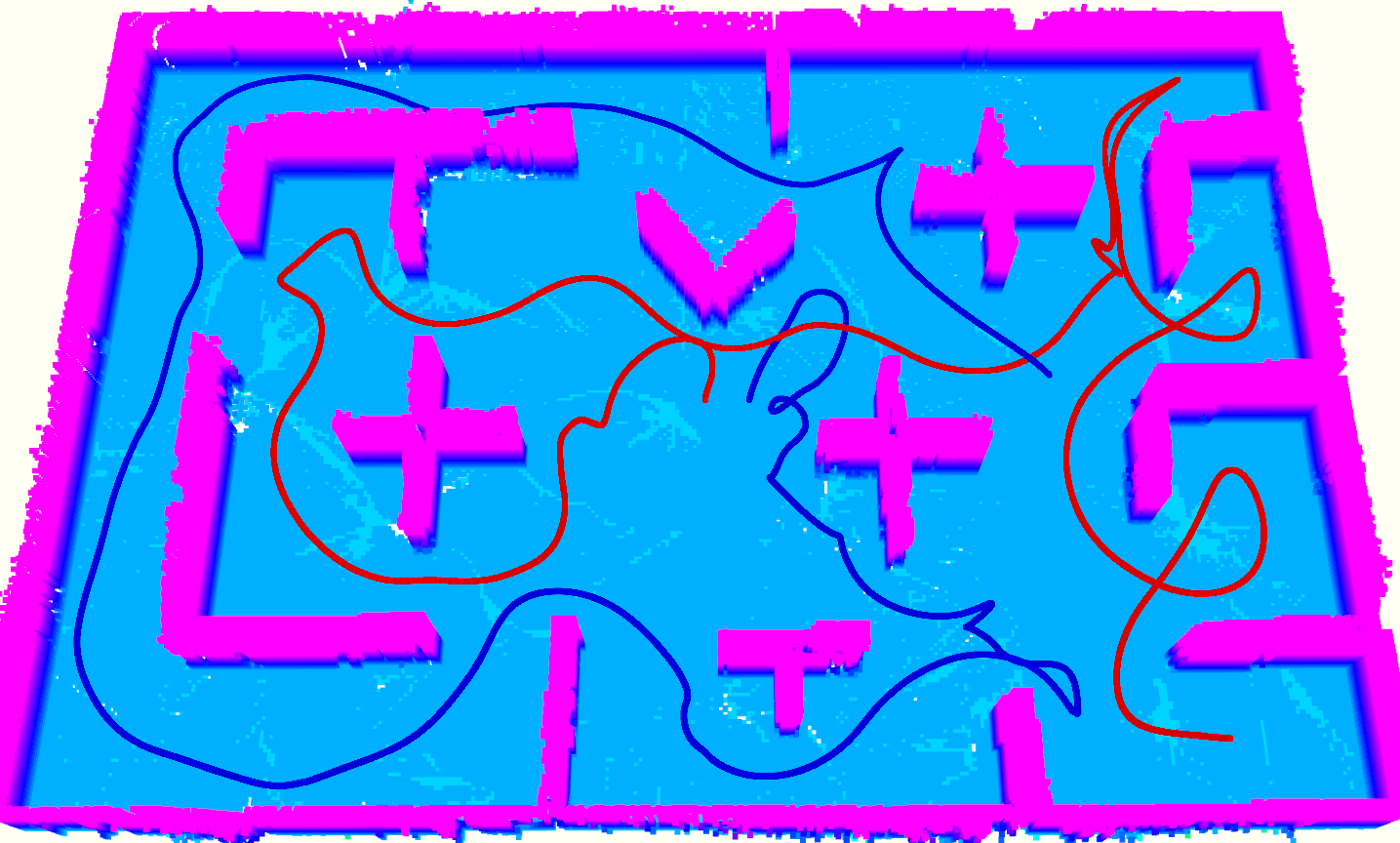}}       
  \subfigure[4 drones.]{\includegraphics[width=0.45\columnwidth]{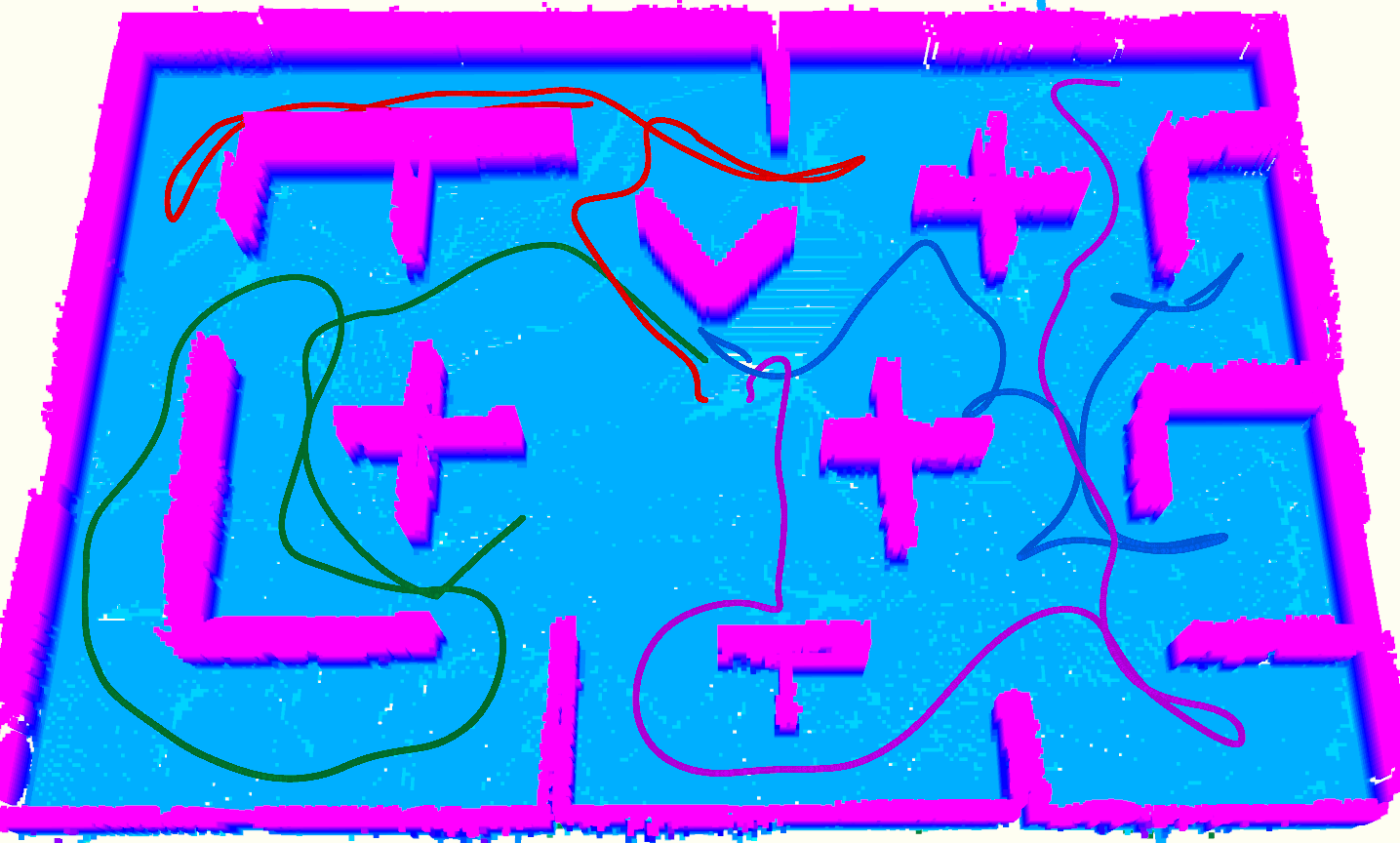}}       
  \subfigure[6 drones.]{\includegraphics[width=0.45\columnwidth]{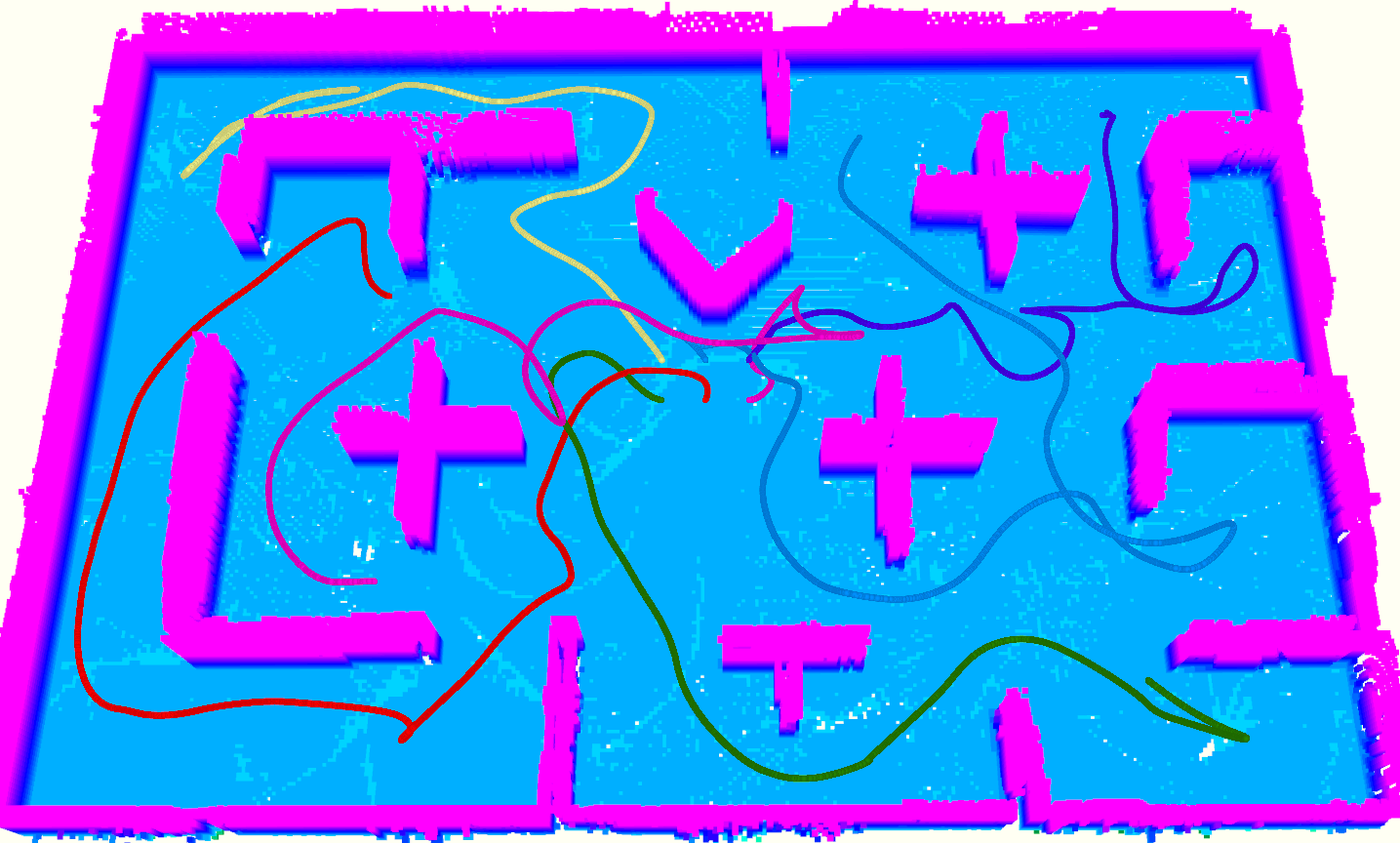}}       
  \subfigure[8 drones.]{\includegraphics[width=0.45\columnwidth]{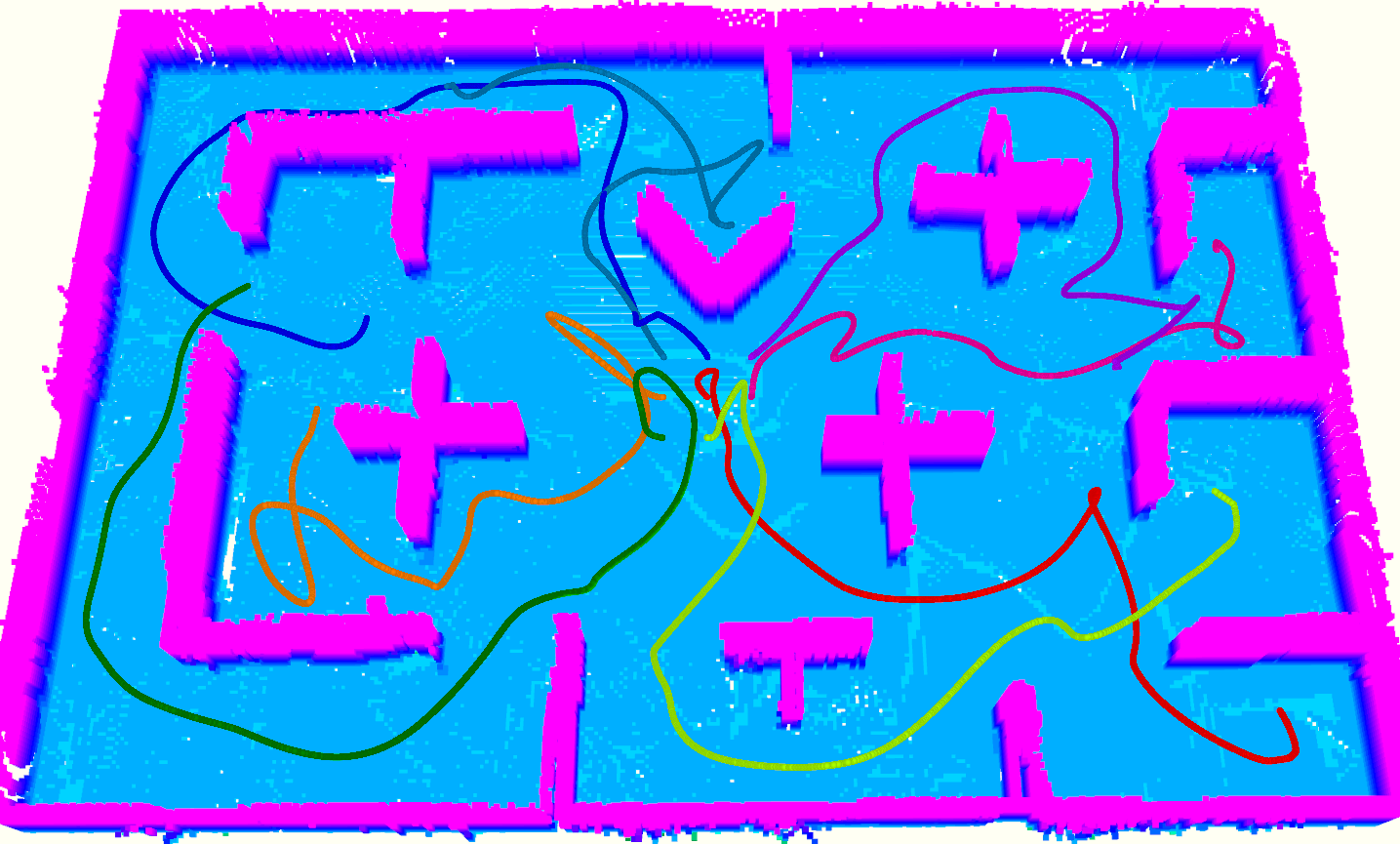}}       
  \subfigure[10 drones.]{\includegraphics[width=0.45\columnwidth]{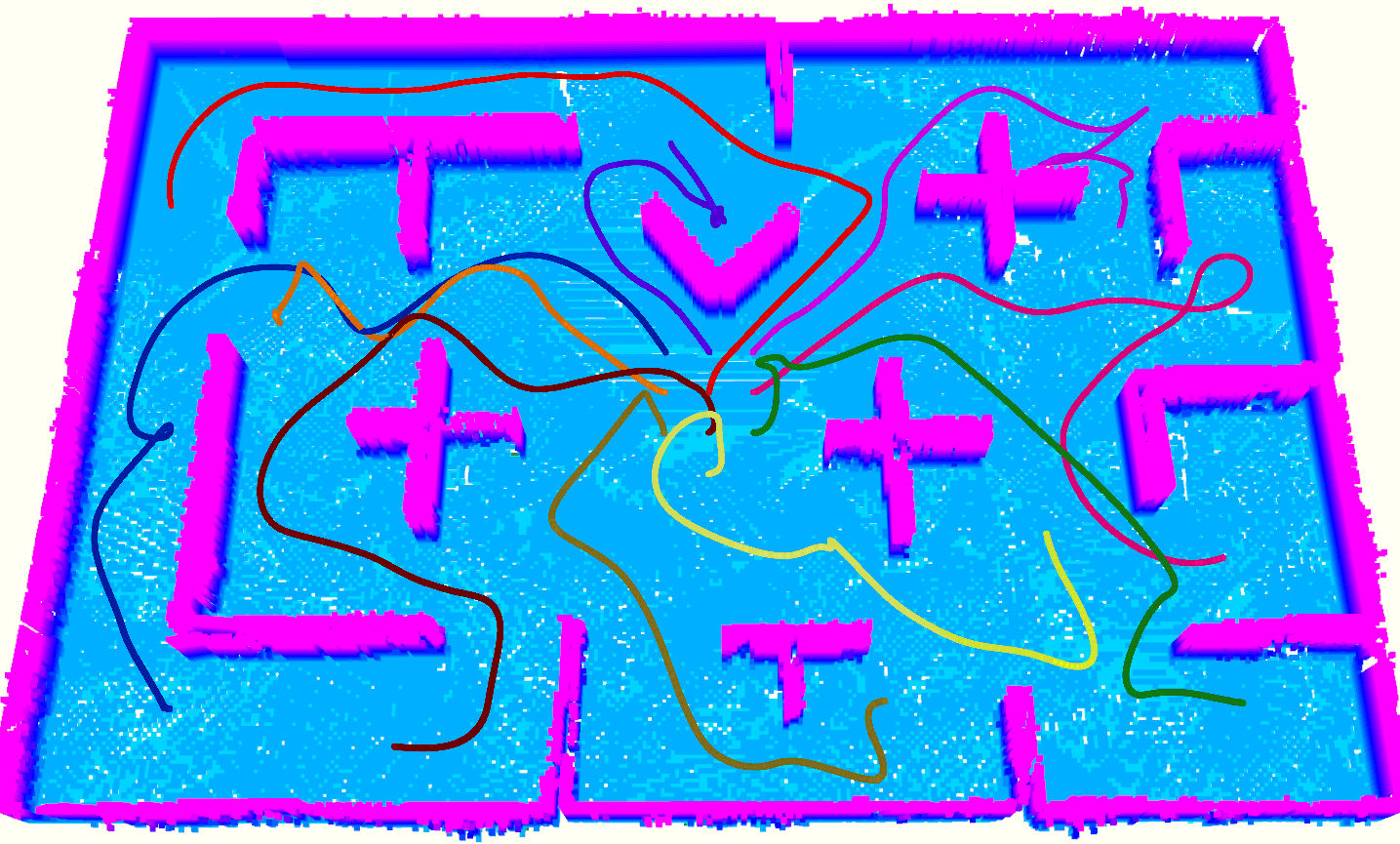}}       
  \caption{\label{fig:number2} Exploration paths with different numbers of quadrotors.
  A large team can be dispatched effectively without significant interference.
  }
  \vspace{-0.1cm}
\end{figure}

\begin{figure}[t!]
  \centering
  \subfigure{\includegraphics[width=0.75\columnwidth]{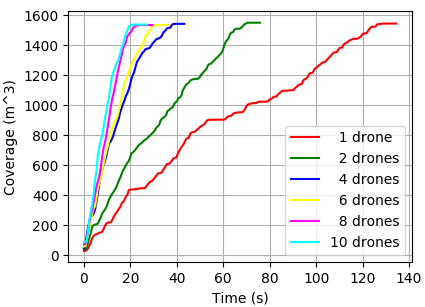}}       
  \vspace{-0.3cm}
  \caption{\label{fig:number} Progress curves of exploration with different numbers of quadrotors.
  The exploration rate increases steadily as the team size grows.
  }
  \vspace{-0.9cm}
\end{figure}

\subsection{Study on Number of Quadrotors}
\label{subsec:number}

\begin{figure*}[t!]
  \centering
  \subfigure[\label{fig:comp_time1}]{\includegraphics[width=0.67\columnwidth]{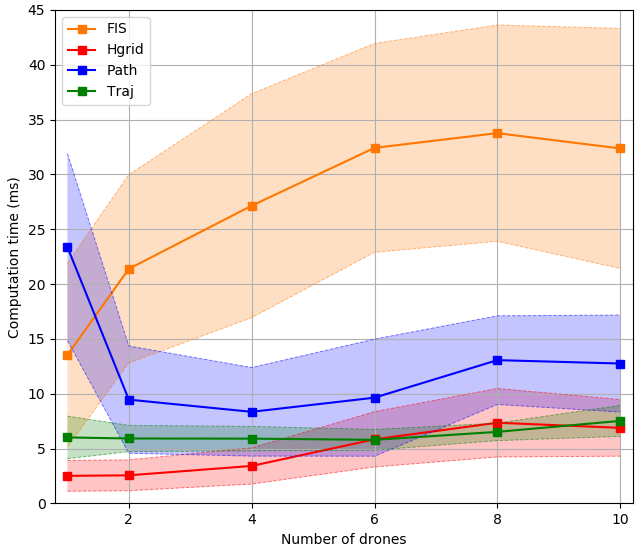}}       
  \subfigure[\label{fig:comp_time2}]{\includegraphics[width=0.67\columnwidth]{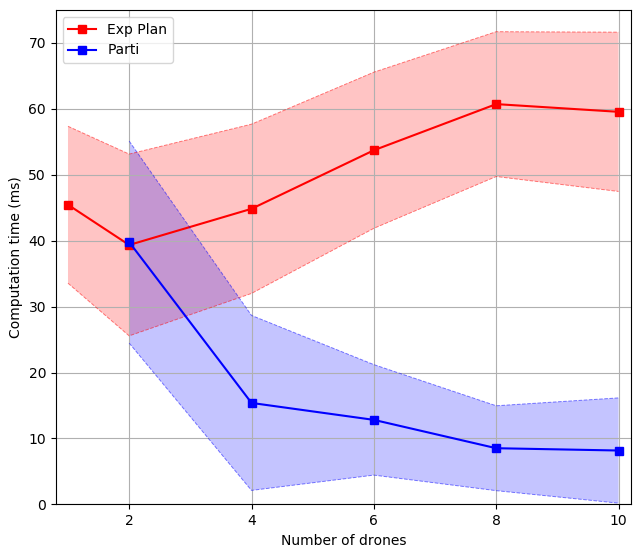}}       
  \subfigure[\label{fig:bandwidth}]{\includegraphics[width=0.67\columnwidth]{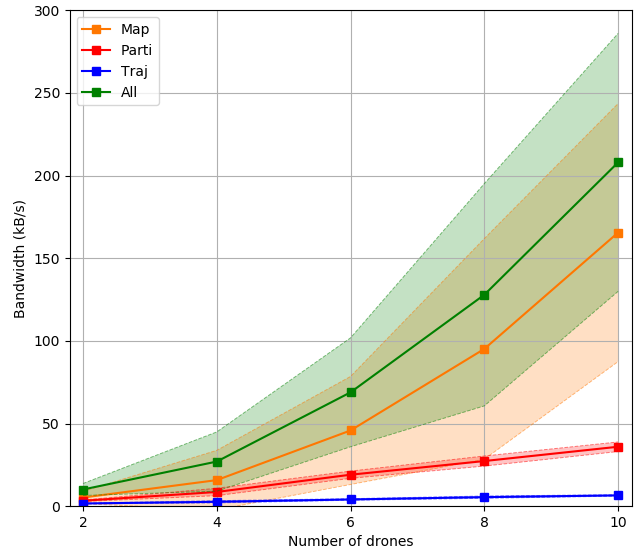}}       
  \caption{\label{fig:timing} {Computation time and communication bandwidth requirements of key components of the proposed coordination and planning approach. The mean and one standard deviation are displayed.}
  }
  \vspace{-0.2cm}
\end{figure*}

To have a clearer understanding of the proposed approach, we study how the number of quadrotors influences its performance. In each test, the quadrotors start at the center of the scene and explore collaboratively. Samples of exploration paths and progress curves are presented in Fig.\ref{fig:number2} and Fig.\ref{fig:number} respectively. We can see that the exploration rate is consistently improved as the number of quadrotors increases. Even with a large team of quadrotors, our approach is able to dispatch them well, where there are only minor interferences. 

{Fig.\ref{fig:timing}(a)(b) report the computation time of major modules in our approach. Fig.\ref{fig:comp_time1} shows the time for the frontier detection and information extraction (\textit{FIS}), update of hgrid (\textit{Hgrid}), the update of CP, path planning and local viewpoint refinement (\textit{Path}), and the trajectory generation (\textit{Traj}). The cumulative computation time of these four modules (\textit{Exp Plan}) is shown in Fig.\ref{fig:comp_time2}. The time of workload partitioning based on CVRP is also shown in Fig.\ref{fig:comp_time2} as \textit{Parti}. Note that the exploration planning and coordination modules execute at different frequencies, i.e., several exploration paths are replanned between two task allocation, so their computation time is displayed separately.} The statistics show that the approach is scalable since the computation time does not increase significantly with more quadrotors. Interestingly, the time of \textit{Parti}, which is the key component to coordinate all quadrotors, decreases as the team size increases. {The reason for this is that as more quadrotors work together to explore the same scene, the number of hgrid cells allocated to each pair of quadrotors decreases, making the associated CVRP easier to solve.} This is in stark contrast to most coordination approaches, in which computation time increases as the number of robots increases. {The time of \textit{Path} decreases noticeably from 1 to 2 quadrotors for the same reason: when there is only one drone, a CP covering the entire space is required, which takes longer to compute. The CPs can be computed faster when more quadrotors are involved.} The shorter time of CPs compensates for the longer time of path planning and local viewpoint refinement, resulting in \textit{Path}'s nearly constant computation time.

{The communication bandwidth requirements are shown in Fig.\ref{fig:bandwidth}. Three types of messages are exchanged during the exploration: the map information (\textit{Map}), the messages involved in workload partitioning (\textit{Parti}) and quadrotor trajectories (\textit{Traj}). The cumulative bandwidth is shown as \textit{All}. It can be seen that \textit{Parti} and \textit{Traj} use less than 50 kB/s, while \textit{Map} takes up the most bandwidth. The total bandwidth requirement is much less than that of our wireless ad hoc network ($>3$ MB/s). It is possible to significantly reduce the size of map data when there are more quadrotors. For example, one can use a communication-efficient map presented in \cite{corah2019communication}.} 


{
\subsection{Study on Suboptimality}
\label{subsec:suboptimal}

\begin{table}[t]
  \centering
  \caption{\label{tab:suboptimal} Study on Suboptimality of Pairwise Interaction.}
  \begin{tabular}{cccccccc} 
  \hline\hline
  \#\textbf{targets}                 & \#\textbf{robots}           & \multicolumn{2}{c}{\textbf{length1 (m)} }   & \multicolumn{2}{c}{ \textbf{length2 (m)} } &  \multicolumn{2}{c}{ \textbf{length2 / length1} }              \\ 
  \cline{3-8}               
  \multirow{9}{*}{50}           &         & \textbf{Avg} & \textbf{Std} & \textbf{Avg} & \textbf{Std} & \textbf{Avg} & \textbf{Std}  \\
                                & 3       & 103.07      &  5.28  &  103.87 &  5.21  &  1.009  &  0.015       \\
                                & 4       & 99.57       &  5.13  &  103.81 &  5.21  &  1.019  &  0.027       \\
                                & 5       & 98.76       &  4.56  &  100.1  &  4.93  &  1.016  &  0.017       \\
                                & 6       & 95.12       &  5.04  &  97.06  &  5.67  &  1.021  &  0.020       \\
                                & 7       & 94.23       &  4.71  &  96.44  &  5.03  &  1.024  &  0.020       \\
                                & 8       & 92.20       &  4.98  &  94.49  &  5.68  &  1.025  &  0.024       \\
                                & 9       & 91.39       &  5.09  &  93.97  &  6.28  &  1.028  &  0.027       \\
                                & 10      & 89.72       &  4.72  &  91.94  &  5.16  &  1.027  &  0.026       \\
 \hline                      
 \multirow{8}{*}{100}        & 3       & 147.14  &  4.74  &  148.32  &  5.31  &  1.009  &   0.013       \\
                             & 4       & 145.35  &  4.23  &  147.29  &  4.70  &  1.015  &   0.019       \\
                             & 5       & 143.56  &  5.16  &  146.14  &  5.37  &  1.018  &   0.020       \\
                             & 6       & 143.56  &  5.16  &  146.14  &  5.37  &  1.018  &   0.020       \\
                             & 7       & 139.76  &  4.43  &  142.88  &  5.09  &  1.023  &   0.021       \\
                             & 8       & 138.34  &  4.51  &  142.73  &  5.56  &  1.031  &   0.020       \\
                             & 9       & 136.78  &  4.06  &  140.34  &  5.34  &  1.026  &   0.022       \\
                             & 10      & 135.77  &  4.39  &  139.26  &  4.78  &  1.026  &   0.020       \\
  \hline\hline
  \end{tabular}
  \vspace{-1.2cm}
\end{table}

\begin{figure*}[t!]
  \centering
  \subfigure{\includegraphics[width=1.99\columnwidth]{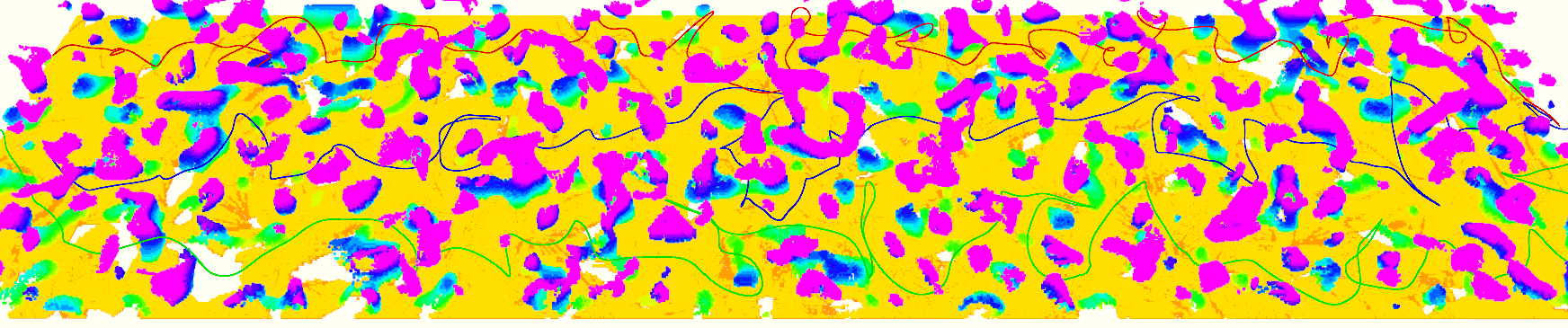}}       
  \caption{\label{fig:large-scale-test} {Exploration test with three quadrotors in a complex large-scale environment. More details can be found in the attached video.} 
  }
\end{figure*}

The pairwise interaction-based coordination is beneficial for communication-limited scenarios, but it may yield suboptimal results. To better understand its performance, we quantitatively compare it to its centralized counterpart. In our test, we generate a different number of target points and robots randomly. For the centralized method, a central server can access the positions of all robots and targets, and a Vehicle Routing Problem (VRP) considering all of them is solved to find the optimal paths passing through all targets. In comparison, the pairwise interaction method only has each pair of robot communicate in sequence and reallocate their assigned targets with VRP. At the beginning, each target point is allocated to the closest robot. The pairwise interaction continues until each pair of robots interact once. We run 100 tests for each specific number of targets and robots, recording the path lengths and length ratios of the two methods. The results in Tab.\ref{tab:suboptimal} show that the paths provided by pairwise interaction (length2) are only marginally longer than those of the centralized VRP (length1). Despite the fact that the pairwise interaction is decentralized and theoretically finds suboptimal solutions, the tests indicate that a single round of interaction suffices to achieve competitive performance compared with its centralized counterpart.
}

{
\subsection{Exploration in Large-scale Environment}

One primary motivation for using multiple quadrotors is to explore large-scale environments more quickly. We conduct exploration tests in a simulated complex large-scale scene to observe the behavior of the proposed system. The scene is surrounded by a $ 20 \times 100 \times 3 \ m^3$ box, which covers an area of 2000 square meters. As shown in Fig.\ref{fig:large-scale-test}, three quadrotors begin on one side of the scene and cooperatively explore until they reach the other side. The dynamics limits are set as $ v_{\tn{max}}=1.5 $ m/s, $a_{\tn{max}} =1.0$ m/s and $ \dot{\varphi}_{\tn{max}} =0.9 $ rad/s. The exploration lasts 153 seconds, and each quadrotor's path length is 176.0, 169.9, and 187.6 m, respectively. The attached video demonstrates the entire exploration process.
}

\subsection{Real-world Exploration Experiments}
\label{subsec:res-real-world}

\begin{figure}[t!]
  \centering
  \subfigure[\label{fig:res-exp2-set1}]{\includegraphics[width=0.75\columnwidth]{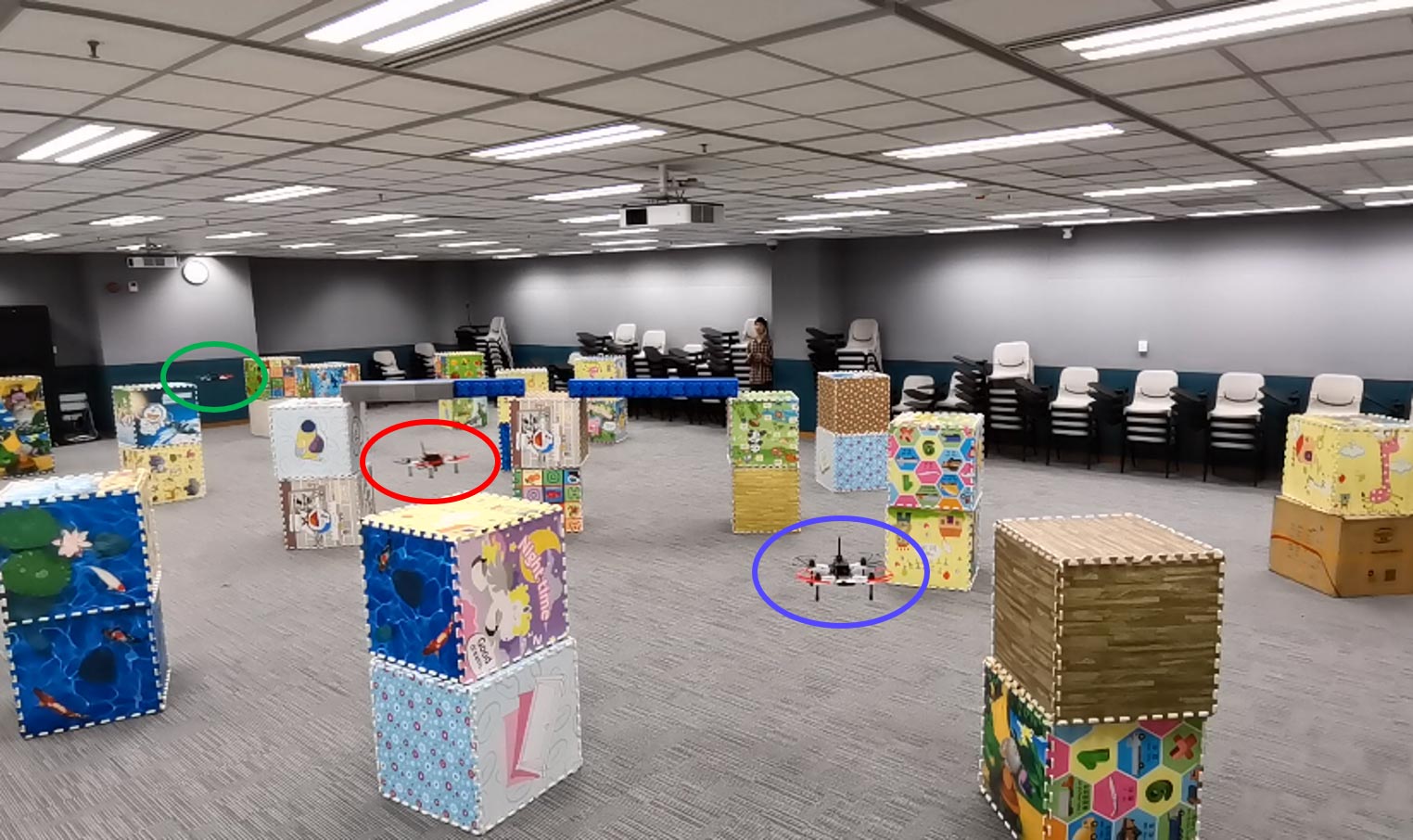}}       
  \subfigure[\label{fig:res-exp2-set2}]{\includegraphics[width=0.75\columnwidth]{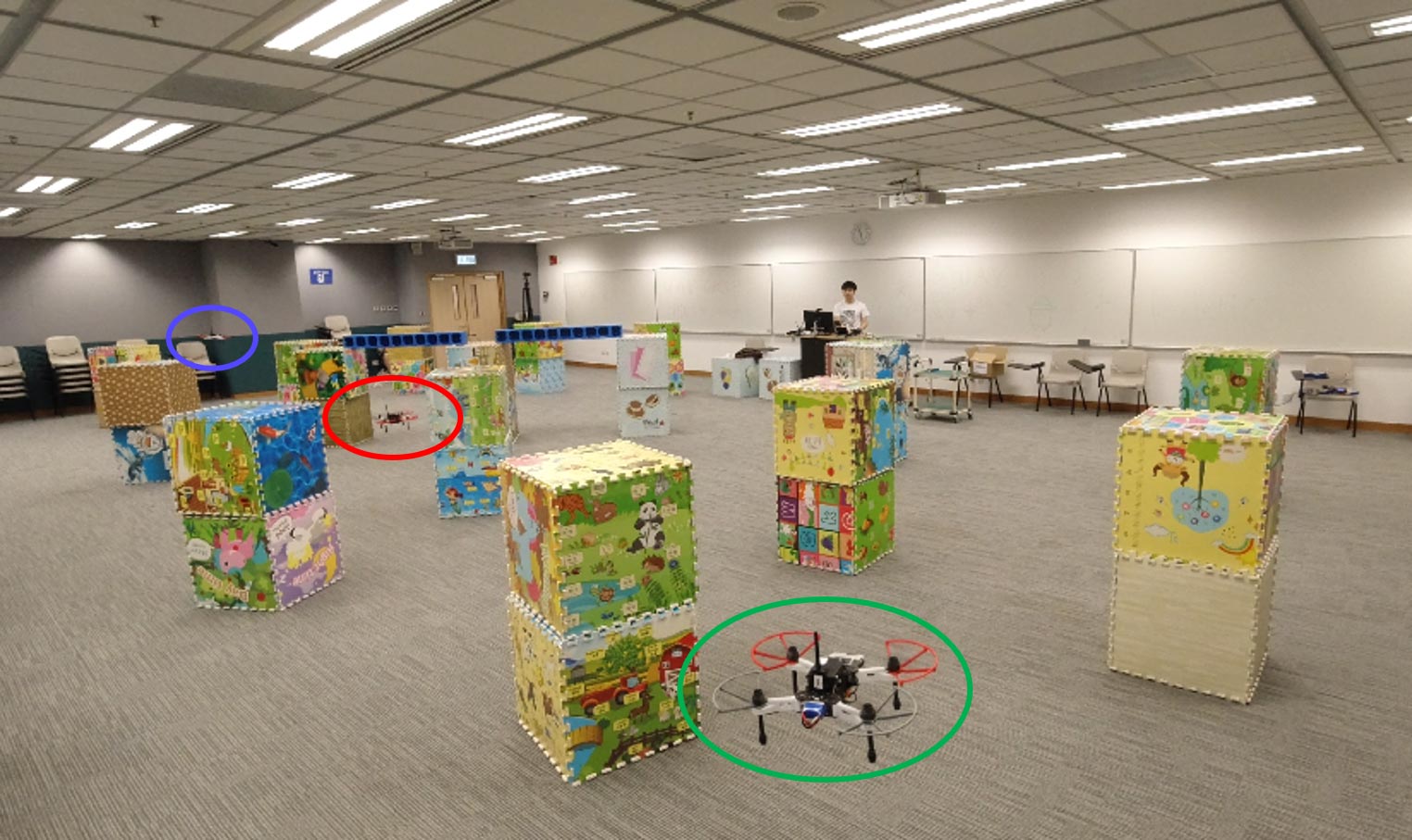}}       
  \subfigure[\label{fig:res-exp2-8s}]{
    \includegraphics[width=0.44\columnwidth]{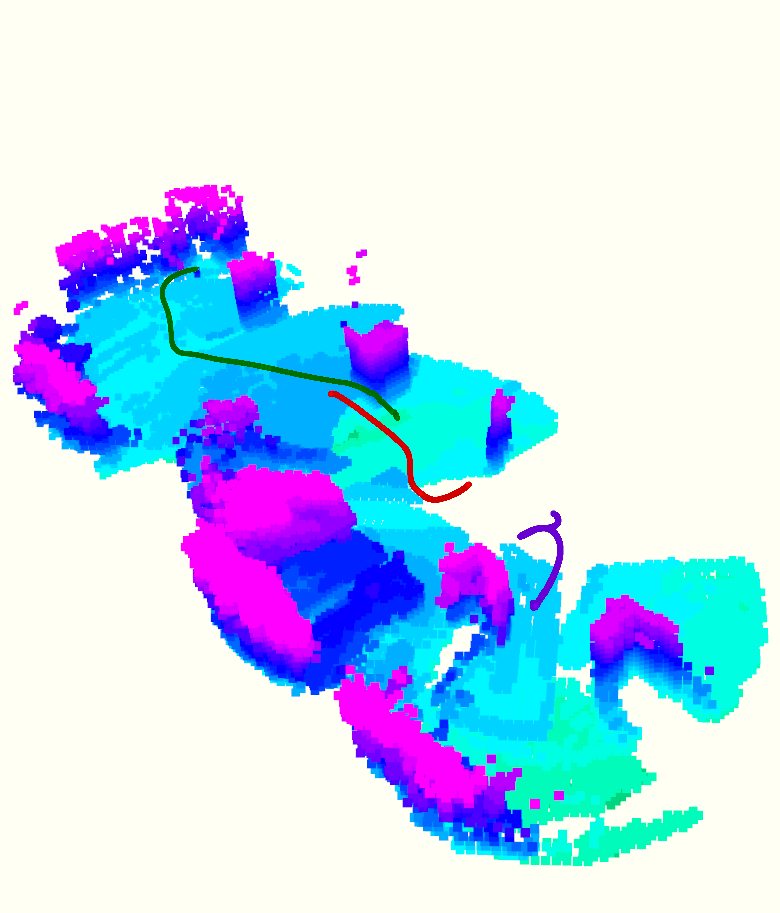}     
    \includegraphics[width=0.555\columnwidth]{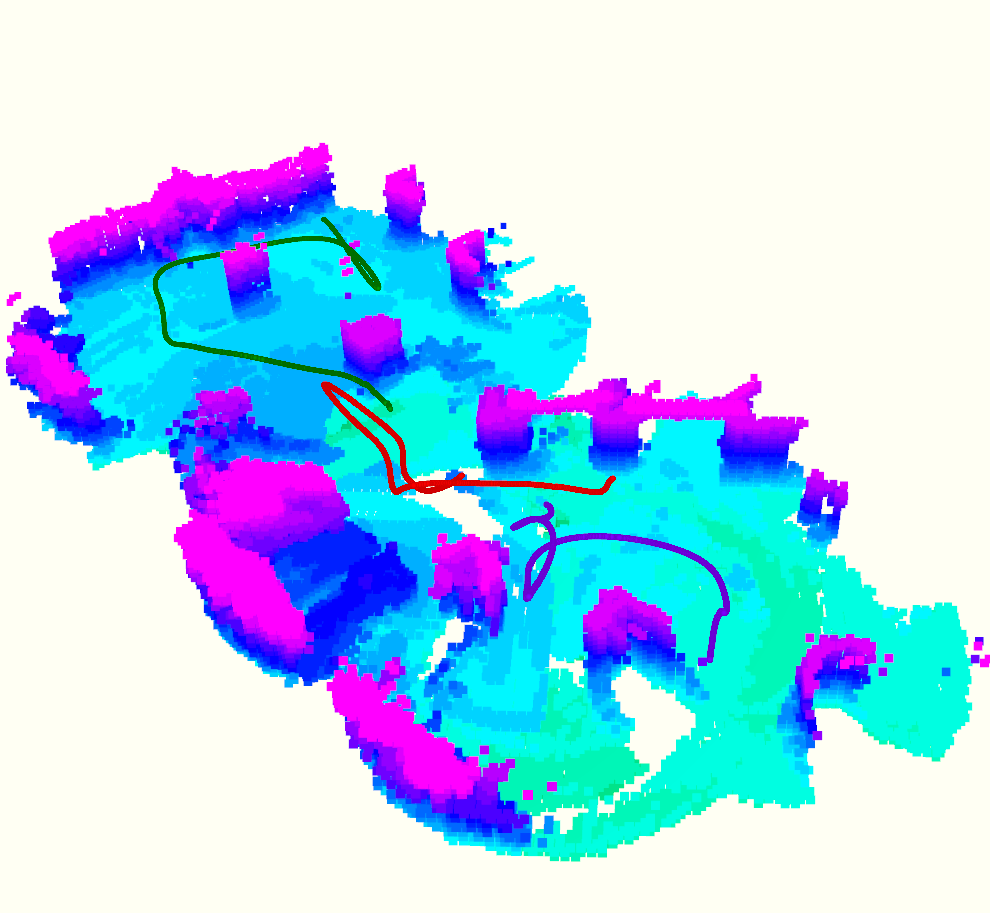}    
  }        
  \subfigure[\label{fig:res-exp2-complete}]{\includegraphics[width=0.7\columnwidth]{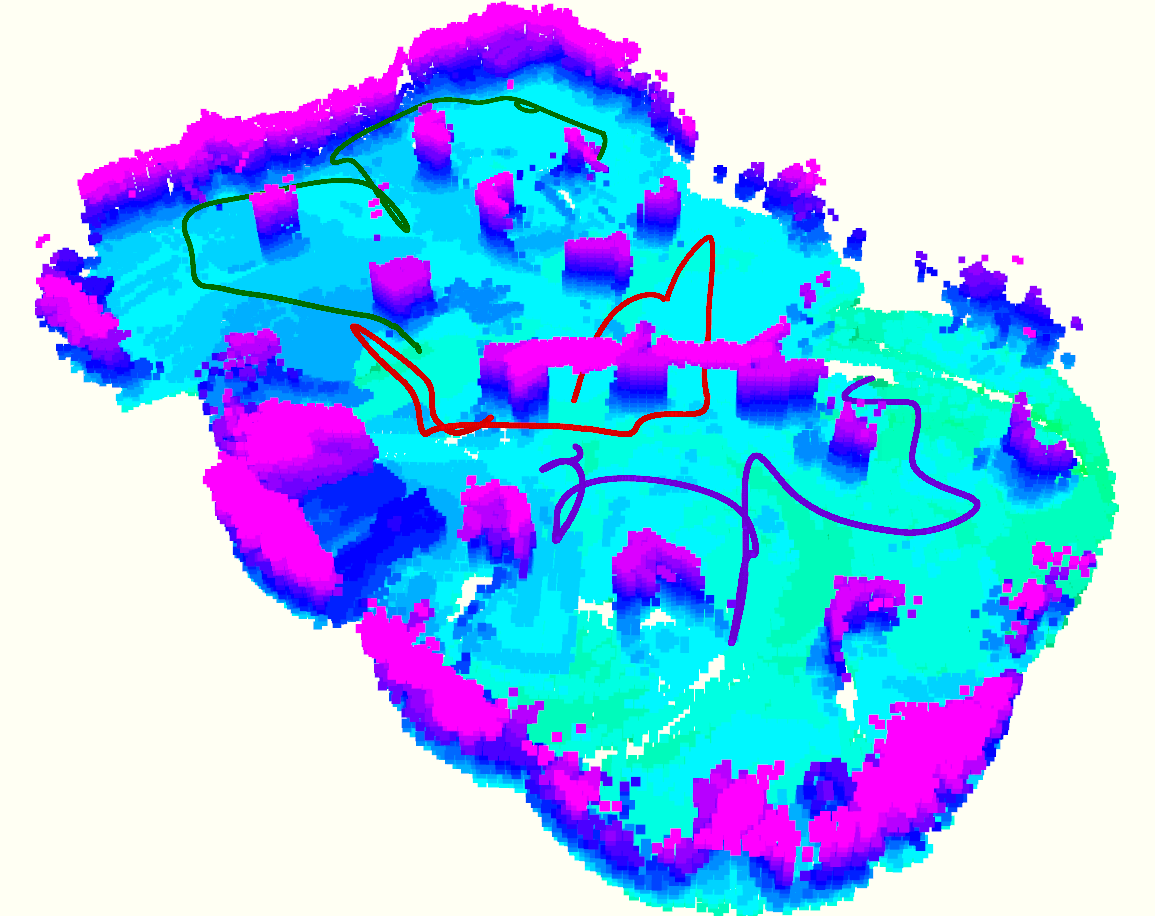}}        
  \caption{\label{fig:res-exp2} Fast exploration experiments with 3 quadrotors in an complex indoor scene. (a)(b) Snapshots of the flight taken from different sites.
    (c) Snapshots of online built map and exploration paths at 8 s and 16 s.
    (d) The complete map and paths of all 3 quadrotors.
  }
\end{figure}

To validate the performance of the proposed approach in real-world scenarios, we conduct extensive field experiments in both indoor and outdoor environments. In all the indoor experiments we set the dynamics limits as $ v_{\tn{max}}=1.0 $ m/s, $a_{\tn{max}} =0.8$ m/s and $ \dot{\varphi}_{\tn{max}} =0.9 $ rad/s. The velocity limit is increased to $v_{\tn{max}}=1.5$ m/s outdoor. Note that we do not use any external device for localization or any central server to control the flights. All state estimation, mapping, coordination, motion planning and control run on the onboard computers. Every quadrotor makes decisions and accomplishes its task in a decentralized fashion. {Besides, our algorithm can support higher flight speed than $1.5$ m/s. However, faster flight can degrade the quality of collected depth images, which is detrimental to accurately mapping the environment.}

Firstly, we conduct fully autonomous exploration experiments in indoor scenes. In the first scene, we test exploration with two quadrotors, as displayed in Fig.\ref{fig:res-exp1}. Within the experiment area, we randomly deploy obstacles to make up a cluttered environment. We bound the space with a $10 \times 6 \times 2 \ m^3$ box. Both quadrotors are initialized with their backs towards the space to be explored, in order to reduce the information obtained by them before starting exploration. A trigger information is sent to the quadrotors\footnote{The information is sent from a ground computer, which serves as an interface to start the experiment. After it, the ground computer no longer controls the quadrotors or runs any algorithm related to the exploration.}, when they start to explore collaboratively. The space is completely explored in 23 s, when the movement distances of the two quadrotors are 13.4 m and 14.4 m respectively. The experiment validates the capability of our system to dispatch multiple quadrotors. It also examines the ability of each quadrotor to perform 3D maneuvers, mapping the unknown space quickly and avoiding obstacles agilely. One sample of an online constructed map and the exploration trajectories is presented in Fig.\ref{fig:res-exp1}. 

\begin{figure}[t!]
  \centering
  \subfigure[\label{fig:res-exp3-env}]{\includegraphics[width=0.75\columnwidth]{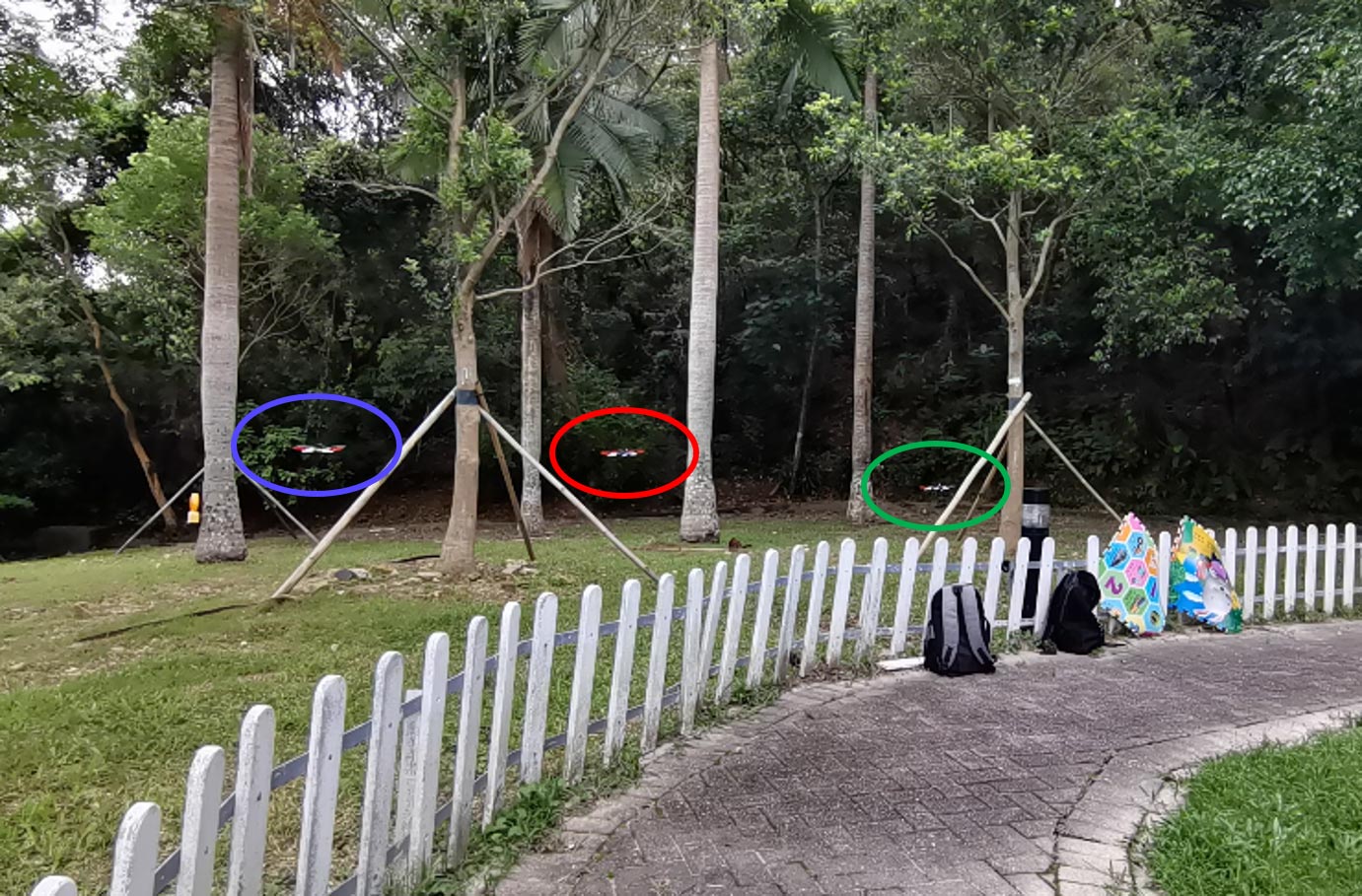}}       
  \subfigure[\label{fig:res-exp3-snap}]{\includegraphics[width=0.65\columnwidth]{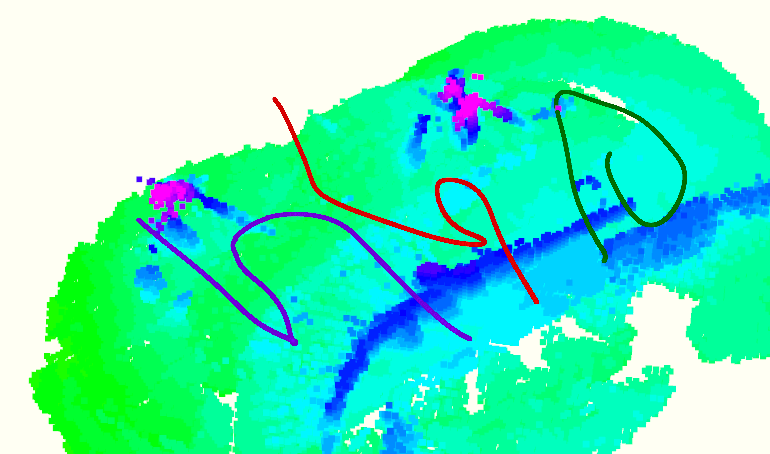}}       
  \subfigure[\label{fig:res-exp3-complete}]{\includegraphics[width=0.75\columnwidth]{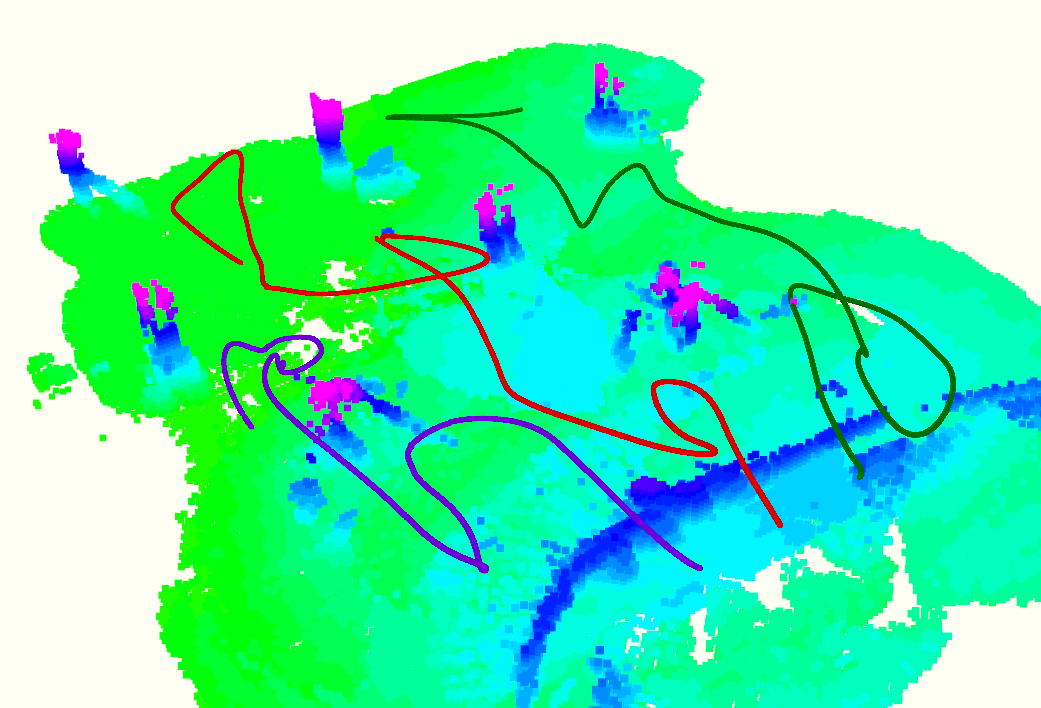}}       
  \vspace{-0.2cm}
  \caption{\label{fig:res-exp3} Exploration in a forest with 3 quadrotors.
  (a) A snapshot of the exploration process. (b) The map and paths at 13 s. (c) The complete map and paths. 
  }
  \vspace{-0.4cm}
\end{figure}

In the second scene, we test our approach in an environment with a larger scale and with three quadrotors.
It is more challenging as more quadrotors are involved and communication becomes less stable when distances between quadrotors increases.
The environment to conduct the experiments is shown in Fig.\ref{fig:res-exp2-set1} and \ref{fig:res-exp2-set2}.
The space is bounded by a $ 15 \times 9 \times 2 \ m^3 $ box.
We follow the same settings as those for experiments with two quadrotors.
The executed trajectories and online built map after exploring for 8 s and 16 s are shown in Fig.\ref{fig:res-exp2-8s}.
The exploration is finished in 33 s, when the path lengths for the three quadrotors are 22.3, 23.0 and 25.7 m. 
The complete map and trajectories are shown in Fig.\ref{fig:res-exp2-complete}. 

Lastly, to validate the robustness of our method in natural environments, we conduct exploration tests in a forest (see Fig.\ref{fig:res-exp3-env}). 
The size of the area to explore is $ 15 \times 12 \times 2 \ m^3 $.
A snapshot of the map and trajectories after starting exploration for 13 s is shown in Fig.\ref{fig:res-exp3-snap}, the complete map and trajectories are shown in Fig.\ref{fig:res-exp3-complete}.
The exploration lasts for 33 s and the path lengths are 33.3, 25.2 and 34.1 m respectively. 

Overall, the experiments demonstrate the applicability of the proposed multi-quadrotor exploration approach in realistic scenarios, where there is limited communication, restricted computation resource and considerable noises in perception.
We refer the readers to the attached video for more details about the experiments, such as the online allocation of tasks, the real-time motion planning, etc.

\vspace{0.2cm}
\section{Conclusions}
\label{sec:conclude}

In this paper, we present a systematic approach for fast exploration of complex environments using a fleet of decentralized quadrotors.
To coordinate the team with only asynchronous and unreliable communication, we decompose the unknown space into hgrid and distribute the task units among quadrotors by a pairwise interaction.
It partitions the workloads appropriately among the quadrotors.
The overall lengths of coverage routes are minimized and the workloads are balanced via a CVRP formulation, which further enhances the team cooperation.
Each quadrotor is capable of exploring the assigned regions safely and efficiently by subsequently building FISs, finding exploration paths, refining viewpoints and generating minimum-time trajectories. 
The performance of the approach is evaluated extensively, showing the high exploration rate, the robustness against communication loss, the capability to dispatch a large team of quadrotors, and the high computation efficiency. 
Moreover, fully autonomous exploration with a fully decentralized multi-UAV system is achieved for the first time. 
To benefit the community, we will release our implementation.
In the future, we plan to improve the reconstruction quality and global consistency of the multi-robot mapping module.
We will also study more sophisticated strategies to deal with limited communication range, like scheduling meeting events for the robots to share knowledge.

\addtolength{\textheight}{0cm}   

\newlength{\bibitemsep}\setlength{\bibitemsep}{0.0\baselineskip}
\newlength{\bibparskip}\setlength{\bibparskip}{0.1pt}
\let\oldthebibliography\thebibliography
\renewcommand\thebibliography[1]{%
\oldthebibliography{#1}%
\setlength{\parskip}{\bibitemsep}%
\setlength{\itemsep}{\bibparskip}%
}

\bibliographystyle{IEEEtran}
\bibliography{zby} 

\begin{thebibliography}{10}
\providecommand{\url}[1]{#1}
\csname url@samestyle\endcsname
\providecommand{\newblock}{\relax}
\providecommand{\bibinfo}[2]{#2}
\providecommand{\BIBentrySTDinterwordspacing}{\spaceskip=0pt\relax}
\providecommand{\BIBentryALTinterwordstretchfactor}{4}
\providecommand{\BIBentryALTinterwordspacing}{\spaceskip=\fontdimen2\font plus
\BIBentryALTinterwordstretchfactor\fontdimen3\font minus
  \fontdimen4\font\relax}
\providecommand{\BIBforeignlanguage}[2]{{%
\expandafter\ifx\csname l@#1\endcsname\relax
\typeout{** WARNING: IEEEtran.bst: No hyphenation pattern has been}%
\typeout{** loaded for the language `#1'. Using the pattern for}%
\typeout{** the default language instead.}%
\else
\language=\csname l@#1\endcsname
\fi
#2}}
\providecommand{\BIBdecl}{\relax}
\BIBdecl

\bibitem{zhou2021fuel}
B.~Zhou, Y.~Zhang, X.~Chen, and S.~Shen, ``Fuel: Fast uav exploration using
  incremental frontier structure and hierarchical planning,'' \emph{IEEE
  Robotics and Automation Letters}, vol.~6, no.~2, pp. 779--786, 2021.

\bibitem{cieslewski2017rapid}
T.~Cieslewski, E.~Kaufmann, and D.~Scaramuzza, ``Rapid exploration with
  multi-rotors: A frontier selection method for high speed flight,'' in
  \emph{Proc. of the {IEEE/RSJ} Intl. Conf. on Intell. Robots and
  Syst.({IROS})}.\hskip 1em plus 0.5em minus 0.4em\relax IEEE, 2017, pp.
  2135--2142.

\bibitem{dharmadhikari2020motion}
M.~Dharmadhikari, T.~Dang, L.~Solanka, J.~Loje, H.~Nguyen, N.~Khedekar, and
  K.~Alexis, ``Motion primitives-based path planning for fast and agile
  exploration using aerial robots,'' in \emph{Proc. of the {IEEE} Intl. Conf.
  on Robot. and Autom. ({ICRA})}.\hskip 1em plus 0.5em minus 0.4em\relax IEEE,
  2020, pp. 179--185.

\bibitem{schmid2020efficient}
L.~Schmid, M.~Pantic, R.~Khanna, L.~Ott, R.~Siegwart, and J.~Nieto, ``An
  efficient sampling-based method for online informative path planning in
  unknown environments,'' \emph{IEEE Robotics and Automation Letters}, vol.~5,
  no.~2, pp. 1500--1507, 2020.

\bibitem{song2017online}
S.~Song and S.~Jo, ``Online inspection path planning for autonomous 3d modeling
  using a micro-aerial vehicle.'' in \emph{Proc. of the {IEEE} Intl. Conf. on
  Robot. and Autom. ({ICRA})}, 2017, pp. 6217--6224.

\bibitem{yamauchi1997frontier}
B.~Yamauchi, ``A frontier-based approach for autonomous exploration,'' in
  \emph{Proceedings 1997 IEEE International Symposium on Computational
  Intelligence in Robotics and Automation CIRA'97.'Towards New Computational
  Principles for Robotics and Automation'}.\hskip 1em plus 0.5em minus
  0.4em\relax IEEE, 1997, pp. 146--151.

\bibitem{julia2012comparison}
M.~Juli{\'a}, A.~Gil, and O.~Reinoso, ``A comparison of path planning
  strategies for autonomous exploration and mapping of unknown environments,''
  \emph{Auton. Robots}, vol.~33, no.~4, pp. 427--444, 2012.

\bibitem{shen2012stochastic}
S.~Shen, N.~Michael, and V.~Kumar, ``Stochastic differential equation-based
  exploration algorithm for autonomous indoor 3d exploration with a
  micro-aerial vehicle,'' \emph{Intl. J. Robot. Research ({IJRR})}, vol.~31,
  no.~12, pp. 1431--1444, 2012.

\bibitem{deng2020robotic}
D.~Deng, R.~Duan, J.~Liu, K.~Sheng, and K.~Shimada, ``Robotic exploration of
  unknown 2d environment using a frontier-based automatic-differentiable
  information gain measure,'' in \emph{2020 IEEE/ASME International Conference
  on Advanced Intelligent Mechatronics (AIM)}.\hskip 1em plus 0.5em minus
  0.4em\relax IEEE, 2020, pp. 1497--1503.

\bibitem{connolly1985determination}
C.~Connolly, ``The determination of next best views,'' in \emph{Proc. of the
  {IEEE} Intl. Conf. on Robot. and Autom. ({ICRA})}, vol.~2.\hskip 1em plus
  0.5em minus 0.4em\relax IEEE, 1985, pp. 432--435.

\bibitem{bircher2016receding}
A.~Bircher, M.~Kamel, K.~Alexis, H.~Oleynikova, and R.~Siegwart, ``Receding
  horizon" next-best-view" planner for 3d exploration,'' in \emph{Proc. of the
  {IEEE} Intl. Conf. on Robot. and Autom. ({ICRA})}.\hskip 1em plus 0.5em minus
  0.4em\relax IEEE, 2016, pp. 1462--1468.

\bibitem{papachristos2017uncertainty}
C.~Papachristos, S.~Khattak, and K.~Alexis, ``Uncertainty-aware receding
  horizon exploration and mapping using aerial robots,'' in \emph{2017 IEEE
  international conference on robotics and automation (ICRA)}.\hskip 1em plus
  0.5em minus 0.4em\relax IEEE, 2017, pp. 4568--4575.

\bibitem{dang2018visual}
T.~Dang, C.~Papachristos, and K.~Alexis, ``Visual saliency-aware receding
  horizon autonomous exploration with application to aerial robotics,'' in
  \emph{2018 IEEE International Conference on Robotics and Automation
  (ICRA)}.\hskip 1em plus 0.5em minus 0.4em\relax IEEE, 2018, pp. 2526--2533.

\bibitem{bircher2018receding}
A.~Bircher, M.~Kamel, K.~Alexis, H.~Oleynikova, and R.~Siegwart, ``Receding
  horizon path planning for 3d exploration and surface inspection,''
  \emph{Auton. Robots}, vol.~42, no.~2, pp. 291--306, 2018.

\bibitem{witting2018history}
C.~Witting, M.~Fehr, R.~B{\"a}hnemann, H.~Oleynikova, and R.~Siegwart,
  ``History-aware autonomous exploration in confined environments using mavs,''
  in \emph{2018 IEEE/RSJ International Conference on Intelligent Robots and
  Systems (IROS)}.\hskip 1em plus 0.5em minus 0.4em\relax IEEE, 2018, pp. 1--9.

\bibitem{wang2019efficient}
C.~Wang, D.~Zhu, T.~Li, M.~Q.-H. Meng, and C.~W. de~Silva, ``Efficient
  autonomous robotic exploration with semantic road map in indoor
  environments,'' \emph{IEEE Robotics and Automation Letters}, vol.~4, no.~3,
  pp. 2989--2996, 2019.

\bibitem{charrow2015information}
B.~Charrow, G.~Kahn, S.~Patil, S.~Liu, K.~Goldberg, P.~Abbeel, N.~Michael, and
  V.~Kumar, ``Information-theoretic planning with trajectory optimization for
  dense 3d mapping.'' in \emph{Proc. of Robot.: Sci. and Syst. ({RSS})},
  vol.~11, 2015.

\bibitem{selin2019efficient}
M.~Selin, M.~Tiger, D.~Duberg, F.~Heintz, and P.~Jensfelt, ``Efficient
  autonomous exploration planning of large-scale 3-d environments,'' \emph{IEEE
  Robotics and Automation Letters}, vol.~4, no.~2, pp. 1699--1706, 2019.

\bibitem{meng2017two}
Z.~Meng, H.~Qin, Z.~Chen, X.~Chen, H.~Sun, F.~Lin, and M.~H. Ang~Jr, ``A
  two-stage optimized next-view planning framework for 3-d unknown environment
  exploration, and structural reconstruction,'' \emph{IEEE Robotics and
  Automation Letters}, vol.~2, no.~3, pp. 1680--1687, 2017.

\bibitem{caoexploring}
C.~Cao, H.~Zhu, H.~Choset, and J.~Zhang, ``Exploring large and complex
  environments fast and efficiently.''

\bibitem{yang2021graph}
F.~Yang, D.-H. Lee, J.~Keller, and S.~Scherer, ``Graph-based topological
  exploration planning in large-scale 3d environments,'' \emph{arXiv preprint
  arXiv:2103.16829}, 2021.

\bibitem{burgard2005coordinated}
W.~Burgard, M.~Moors, C.~Stachniss, and F.~E. Schneider, ``Coordinated
  multi-robot exploration,'' \emph{{IEEE} Trans. Robot. ({TRO})}, vol.~21,
  no.~3, pp. 376--386, 2005.

\bibitem{butzke2011planning}
J.~Butzke and M.~Likhachev, ``Planning for multi-robot exploration with
  multiple objective utility functions,'' in \emph{Proc. of the {IEEE/RSJ}
  Intl. Conf. on Intell. Robots and Syst.({IROS})}.\hskip 1em plus 0.5em minus
  0.4em\relax IEEE, 2011, pp. 3254--3259.

\bibitem{wurm2008coordinated}
K.~M. Wurm, C.~Stachniss, and W.~Burgard, ``Coordinated multi-robot exploration
  using a segmentation of the environment,'' in \emph{Proc. of the {IEEE/RSJ}
  Intl. Conf. on Intell. Robots and Syst.({IROS})}.\hskip 1em plus 0.5em minus
  0.4em\relax IEEE, 2008, pp. 1160--1165.

\bibitem{faigl2012goal}
J.~Faigl, M.~Kulich, and L.~P{\v{r}}eu{\v{c}}il, ``Goal assignment using
  distance cost in multi-robot exploration,'' in \emph{Proc. of the {IEEE/RSJ}
  Intl. Conf. on Intell. Robots and Syst.({IROS})}.\hskip 1em plus 0.5em minus
  0.4em\relax IEEE, 2012, pp. 3741--3746.

\bibitem{dong2019multi}
S.~Dong, K.~Xu, Q.~Zhou, A.~Tagliasacchi, S.~Xin, M.~Nie{\ss}ner, and B.~Chen,
  ``Multi-robot collaborative dense scene reconstruction,'' \emph{ACM
  Transactions on Graphics (TOG)}, vol.~38, no.~4, pp. 1--16, 2019.

\bibitem{karapetyan2017efficient}
N.~Karapetyan, K.~Benson, C.~McKinney, P.~Taslakian, and I.~Rekleitis,
  ``Efficient multi-robot coverage of a known environment,'' in \emph{2017
  IEEE/RSJ International Conference on Intelligent Robots and Systems
  (IROS)}.\hskip 1em plus 0.5em minus 0.4em\relax IEEE, 2017, pp. 1846--1852.

\bibitem{hardouin2020next}
G.~Hardouin, J.~Moras, F.~Morbidi, J.~Marzat, and E.~M. Mouaddib,
  ``Next-best-view planning for surface reconstruction of large-scale 3d
  environments with multiple uavs,'' in \emph{Proc. of the {IEEE/RSJ} Intl.
  Conf. on Intell. Robots and Syst.({IROS})}.\hskip 1em plus 0.5em minus
  0.4em\relax IEEE, 2020, pp. 1567--1574.

\bibitem{yamauchi1999decentralized}
B.~Yamauchi, ``Decentralized coordination for multirobot exploration,''
  \emph{Robotics and Autonomous Systems}, vol.~29, no. 2-3, pp. 111--118, 1999.

\bibitem{zlot2002multi}
R.~Zlot, A.~Stentz, M.~B. Dias, and S.~Thayer, ``Multi-robot exploration
  controlled by a market economy,'' in \emph{Proc. of the {IEEE} Intl. Conf. on
  Robot. and Autom. ({ICRA})}, vol.~3.\hskip 1em plus 0.5em minus 0.4em\relax
  IEEE, 2002, pp. 3016--3023.

\bibitem{smith2018distributed}
A.~J. Smith and G.~A. Hollinger, ``Distributed inference-based multi-robot
  exploration,'' \emph{Auton. Robots}, vol.~42, no.~8, pp. 1651--1668, 2018.

\bibitem{berhault2003robot}
M.~Berhault, H.~Huang, P.~Keskinocak, S.~Koenig, W.~Elmaghraby, P.~Griffin, and
  A.~Kleywegt, ``Robot exploration with combinatorial auctions,'' in
  \emph{Proc. of the {IEEE/RSJ} Intl. Conf. on Intell. Robots and
  Syst.({IROS})}, vol.~2.\hskip 1em plus 0.5em minus 0.4em\relax IEEE, 2003,
  pp. 1957--1962.

\bibitem{corah2019distributed}
M.~Corah and N.~Michael, ``Distributed matroid-constrained submodular
  maximization for multi-robot exploration: Theory and practice,'' \emph{Auton.
  Robots}, vol.~43, no.~2, pp. 485--501, 2019.

\bibitem{yusmmr}
J.~Yu, J.~Tong, Y.~Xu, Z.~Xu, H.~Dong, T.~Yang, and Y.~Wang, ``Smmr-explore:
  Submap-based multi-robot exploration system with multi-robot multi-target
  potential field exploration method.''

\bibitem{klodt2015equitable}
L.~Klodt and V.~Willert, ``Equitable workload partitioning for multi-robot
  exploration through pairwise optimization,'' in \emph{Proc. of the {IEEE/RSJ}
  Intl. Conf. on Intell. Robots and Syst.({IROS})}.\hskip 1em plus 0.5em minus
  0.4em\relax IEEE, 2015, pp. 2809--2816.

\bibitem{corah2021volumetric}
M.~Corah and N.~Michael, ``Volumetric objectives for multi-robot exploration of
  three-dimensional environments,'' in \emph{2021 IEEE International Conference
  on Robotics and Automation (ICRA)}.\hskip 1em plus 0.5em minus 0.4em\relax
  IEEE, 2021, pp. 9043--9050.

\bibitem{durham2011discrete}
J.~W. Durham, R.~Carli, P.~Frasca, and F.~Bullo, ``Discrete partitioning and
  coverage control for gossiping robots,'' \emph{{IEEE} Trans. Robot. ({TRO})},
  vol.~28, no.~2, pp. 364--378, 2011.

\bibitem{kulkarni2021autonomous}
M.~Kulkarni, M.~Dharmadhikari, M.~Tranzatto, S.~Zimmermann, V.~Reijgwart,
  P.~De~Petris, H.~Nguyen, N.~Khedekar, C.~Papachristos, L.~Ott \emph{et~al.},
  ``Autonomous teamed exploration of subterranean environments using legged and
  aerial robots,'' \emph{arXiv preprint arXiv:2111.06482}, 2021.

\bibitem{agha2021nebula}
A.~Agha, K.~Otsu, B.~Morrell, D.~D. Fan, R.~Thakker, A.~Santamaria-Navarro,
  S.-K. Kim, A.~Bouman, X.~Lei, J.~Edlund \emph{et~al.}, ``Nebula: Quest for
  robotic autonomy in challenging environments; team costar at the darpa
  subterranean challenge,'' \emph{arXiv preprint arXiv:2103.11470}, 2021.

\bibitem{hudson2021heterogeneous}
N.~Hudson, F.~Talbot, M.~Cox, J.~Williams, T.~Hines, A.~Pitt, B.~Wood,
  D.~Frousheger, K.~L. Surdo, T.~Molnar \emph{et~al.}, ``Heterogeneous ground
  and air platforms, homogeneous sensing: Team csiro data61's approach to the
  darpa subterranean challenge,'' \emph{arXiv preprint arXiv:2104.09053}, 2021.

\bibitem{rouvcek2021system}
T.~Rou{\v{c}}ek, M.~Pecka, P.~{\v{C}}{\'\i}{\v{z}}ek,
  T.~Pet{\v{r}}{\'\i}{\v{c}}ek, J.~Bayer, V.~{\v{S}}alansk{\`y}, T.~Azayev,
  D.~He{\v{r}}t, M.~Petrl{\'\i}k, T.~B{\'a}{\v{c}}a \emph{et~al.}, ``System for
  multi-robotic exploration of underground environments ctu-cras-norlab in the
  darpa subterranean challenge,'' \emph{arXiv preprint arXiv:2110.05911}, 2021.

\bibitem{ohradzansky2021multi}
M.~T. Ohradzansky, E.~R. Rush, D.~G. Riley, A.~B. Mills, S.~Ahmad, S.~McGuire,
  H.~Biggie, K.~Harlow, M.~J. Miles, E.~W. Frew \emph{et~al.}, ``Multi-agent
  autonomy: Advancements and challenges in subterranean exploration,''
  \emph{arXiv preprint arXiv:2110.04390}, 2021.

\bibitem{petravcek2021large}
P.~Petr{\'a}{\v{c}}ek, V.~Kr{\'a}tk{\`y}, M.~Petrl{\'\i}k, T.~B{\'a}{\v{c}}a,
  R.~Kratochv{\'\i}l, and M.~Saska, ``Large-scale exploration of cave
  environments by unmanned aerial vehicles,'' \emph{IEEE Robotics and
  Automation Letters}, vol.~6, no.~4, pp. 7596--7603, 2021.

\bibitem{corah2019communication}
M.~Corah, C.~O’Meadhra, K.~Goel, and N.~Michael, ``Communication-efficient
  planning and mapping for multi-robot exploration in large environments,''
  \emph{IEEE Robotics and Automation Letters}, vol.~4, no.~2, pp. 1715--1721,
  2019.

\bibitem{tian2020search}
Y.~Tian, K.~Liu, K.~Ok, L.~Tran, D.~Allen, N.~Roy, and J.~P. How, ``Search and
  rescue under the forest canopy using multiple uavs,'' \emph{The International
  Journal of Robotics Research}, vol.~39, no. 10-11, pp. 1201--1221, 2020.

\bibitem{cesare2015multi}
K.~Cesare, R.~Skeele, S.-H. Yoo, Y.~Zhang, and G.~Hollinger, ``Multi-uav
  exploration with limited communication and battery,'' in \emph{2015 IEEE
  international conference on robotics and automation (ICRA)}.\hskip 1em plus
  0.5em minus 0.4em\relax IEEE, 2015, pp. 2230--2235.

\bibitem{mcguire2019minimal}
K.~McGuire, C.~De~Wagter, K.~Tuyls, H.~Kappen, and G.~C. de~Croon, ``Minimal
  navigation solution for a swarm of tiny flying robots to explore an unknown
  environment,'' \emph{Science Robotics}, vol.~4, no.~35, p. eaaw9710, 2019.

\bibitem{oleynikova2016continuous}
H.~Oleynikova, M.~Burri, Z.~Taylor, J.~Nieto, R.~Siegwart, and E.~Galceran,
  ``Continuous-time trajectory optimization for online uav replanning,'' in
  \emph{Proc. of the {IEEE/RSJ} Intl. Conf. on Intell. Robots and
  Syst.({IROS})}, Daejeon, Korea, Oct. 2016, pp. 5332--5339.

\bibitem{fei2017iros}
F.~Gao, Y.~Lin, and S.~Shen, ``Gradient-based online safe trajectory generation
  for quadrotor flight in complex environments,'' in \emph{Proc. of the
  {IEEE/RSJ} Intl. Conf. on Intell. Robots and Syst.({IROS})}, Sept 2017, pp.
  3681--3688.

\bibitem{usenko2017real}
V.~Usenko, L.~von Stumberg, A.~Pangercic, and D.~Cremers, ``Real-time
  trajectory replanning for mavs using uniform b-splines and a 3d circular
  buffer,'' in \emph{Proc. of the {IEEE/RSJ} Intl. Conf. on Intell. Robots and
  Syst.({IROS})}.\hskip 1em plus 0.5em minus 0.4em\relax IEEE, 2017, pp.
  215--222.

\bibitem{zhou2019robust}
B.~Zhou, F.~Gao, L.~Wang, C.~Liu, and S.~Shen, ``Robust and efficient quadrotor
  trajectory generation for fast autonomous flight,'' \emph{IEEE Robotics and
  Automation Letters}, vol.~4, no.~4, pp. 3529--3536, 2019.

\bibitem{zhou2020robust}
B.~Zhou, F.~Gao, J.~Pan, and S.~Shen, ``Robust real-time uav replanning using
  guided gradient-based optimization and topological paths,'' in \emph{Proc. of
  the {IEEE} Intl. Conf. on Robot. and Autom. ({ICRA})}.\hskip 1em plus 0.5em
  minus 0.4em\relax IEEE, 2020, pp. 1208--1214.

\bibitem{zhou2020raptor}
B.~Zhou, J.~Pan, F.~Gao, and S.~Shen, ``Raptor: Robust and perception-aware
  trajectory replanning for quadrotor fast flight,'' \emph{arXiv preprint
  arXiv:2007.03465}, 2020.

\bibitem{zhou2020ego}
X.~Zhou, Z.~Wang, H.~Ye, C.~Xu, and F.~Gao, ``Ego-planner: An esdf-free
  gradient-based local planner for quadrotors,'' \emph{IEEE Robotics and
  Automation Letters}, vol.~6, no.~2, pp. 478--485, 2020.

\bibitem{ratliff2009chomp}
N.~Ratliff, M.~Zucker, J.~A. Bagnell, and S.~Srinivasa, ``Chomp: Gradient
  optimization techniques for efficient motion planning,'' in \emph{Proc. of
  the {IEEE} Intl. Conf. on Robot. and Autom. ({ICRA})}, May 2009, pp.
  489--494.

\bibitem{van2011reciprocal}
J.~Van Den~Berg, S.~J. Guy, M.~Lin, and D.~Manocha, ``Reciprocal n-body
  collision avoidance,'' in \emph{Robotics research}.\hskip 1em plus 0.5em
  minus 0.4em\relax Springer, 2011, pp. 3--19.

\bibitem{van2011reciprocal2}
J.~Van Den~Berg, J.~Snape, S.~J. Guy, and D.~Manocha, ``Reciprocal collision
  avoidance with acceleration-velocity obstacles,'' in \emph{Proc. of the
  {IEEE} Intl. Conf. on Robot. and Autom. ({ICRA})}.\hskip 1em plus 0.5em minus
  0.4em\relax IEEE, 2011, pp. 3475--3482.

\bibitem{arul2020dcad}
S.~H. Arul and D.~Manocha, ``Dcad: Decentralized collision avoidance with
  dynamics constraints for agile quadrotor swarms,'' \emph{IEEE Robotics and
  Automation Letters}, vol.~5, no.~2, pp. 1191--1198, 2020.

\bibitem{liu2020mapper}
Z.~Liu, B.~Chen, H.~Zhou, G.~Koushik, M.~Hebert, and D.~Zhao, ``Mapper:
  Multi-agent path planning with evolutionary reinforcement learning in mixed
  dynamic environments,'' in \emph{2020 IEEE/RSJ International Conference on
  Intelligent Robots and Systems (IROS)}.\hskip 1em plus 0.5em minus
  0.4em\relax IEEE, 2020, pp. 11\,748--11\,754.

\bibitem{park2020efficient}
J.~Park, J.~Kim, I.~Jang, and H.~J. Kim, ``Efficient multi-agent trajectory
  planning with feasibility guarantee using relative bernstein polynomial,'' in
  \emph{2020 IEEE International Conference on Robotics and Automation
  (ICRA)}.\hskip 1em plus 0.5em minus 0.4em\relax IEEE, 2020, pp. 434--440.

\bibitem{zhou2020ego2}
X.~Zhou, X.~Wen, J.~Zhu, H.~Zhou, C.~Xu, and F.~Gao, ``Ego-swarm: A fully
  autonomous and decentralized quadrotor swarm system in cluttered
  environments,'' \emph{arXiv preprint arXiv:2011.04183}, 2020.

\bibitem{xu2021omni}
H.~Xu, Y.~Zhang, B.~Zhou, L.~Wang, X.~Yao, G.~Meng, and S.~Shen, ``Omni-swarm:
  A decentralized omnidirectional visual-inertial-uwb state estimation system
  for aerial swarm,'' \emph{arXiv preprint arXiv:2103.04131}, 2021.

\bibitem{ericson2004real}
C.~Ericson, \emph{Real-time collision detection}.\hskip 1em plus 0.5em minus
  0.4em\relax CRC Press, 2004.

\bibitem{meagher1982geometric2}
D.~Meagher, ``Geometric modeling using octree encoding,'' \emph{Computer
  graphics and image processing}, vol.~19, no.~2, pp. 129--147, 1982.

\bibitem{wurm2010octomap}
K.~M. Wurm, A.~Hornung, M.~Bennewitz, C.~Stachniss, and W.~Burgard, ``Octomap:
  A probabilistic, flexible, and compact 3d map representation for robotic
  systems,'' in \emph{Proc. of the {IEEE} Intl. Conf. on Robot. and Autom.
  ({ICRA})}, vol.~2, Anchorage, AK, US, May 2010.

\bibitem{kusnur2021planning}
T.~Kusnur, S.~Mukherjee, D.~M. Saxena, T.~Fukami, T.~Koyama, O.~Salzman, and
  M.~Likhachev, ``A planning framework for persistent, multi-uav coverage with
  global deconfliction,'' in \emph{Field and Service Robotics}.\hskip 1em plus
  0.5em minus 0.4em\relax Springer, 2021, pp. 459--474.

\bibitem{helsgaun2017extension}
K.~Helsgaun, ``An extension of the lin-kernighan-helsgaun tsp solver for
  constrained traveling salesman and vehicle routing problems,''
  \emph{Roskilde: Roskilde University}, 2017.

\bibitem{oleynikova2018sparse}
H.~Oleynikova, Z.~Taylor, R.~Siegwart, and J.~Nieto, ``Sparse 3d topological
  graphs for micro-aerial vehicle planning,'' in \emph{Proc. of the {IEEE/RSJ}
  Intl. Conf. on Intell. Robots and Syst.({IROS})}.\hskip 1em plus 0.5em minus
  0.4em\relax IEEE, 2018, pp. 1--9.

\bibitem{MelKum1105}
D.~Mellinger and V.~Kumar, ``Minimum snap trajectory generation and control for
  quadrotors,'' in \emph{Proc. of the {IEEE} Intl. Conf. on Robot. and Autom.
  ({ICRA})}, Shanghai, China, May 2011, pp. 2520--2525.

\bibitem{han2019fiesta}
L.~Han, F.~Gao, B.~Zhou, and S.~Shen, ``Fiesta: Fast incremental euclidean
  distance fields for online motion planning of aerial robots,'' \emph{arXiv
  preprint arXiv:1903.02144}, 2019.

\bibitem{dubois2020dense}
R.~Dubois, A.~Eudes, J.~Moras, and V.~Fr{\'e}mont, ``Dense decentralized
  multi-robot slam based on locally consistent tsdf submaps,'' in \emph{Proc.
  of the {IEEE/RSJ} Intl. Conf. on Intell. Robots and Syst.({IROS})}.\hskip 1em
  plus 0.5em minus 0.4em\relax IEEE, 2020, pp. 4862--4869.

\bibitem{schwartz1986telecommunication}
M.~Schwartz, \emph{Telecommunication networks: protocols, modeling and
  analysis}.\hskip 1em plus 0.5em minus 0.4em\relax Addison-Wesley Longman
  Publishing Co., Inc., 1986.

\bibitem{nguyen2020vision}
T.~Nguyen, K.~Mohta, C.~J. Taylor, and V.~Kumar, ``Vision-based multi-mav
  localization with anonymous relative measurements using coupled probabilistic
  data association filter,'' in \emph{2020 IEEE International Conference on
  Robotics and Automation (ICRA)}.\hskip 1em plus 0.5em minus 0.4em\relax IEEE,
  2020, pp. 3349--3355.

\bibitem{lee2010}
T.~Lee, M.~Leoky, and N.~H. McClamroch, ``Geometric tracking control of a
  quadrotor uav on se (3),'' in \emph{Proc. of the {IEEE} Control and Decision
  Conf. ({CDC})}, Atlanta, GA, Dec. 2010, pp. 5420--5425.

\end{thebibliography}


\end{document}